\documentclass[11pt, a4paper, logo, onecolumn, copyright]{deepmind}

\usepackage[authoryear, sort&compress, round]{natbib}
\usepackage{array}
\usepackage{multirow}
\usepackage{caption}
\usepackage{tikz}
\usetikzlibrary{arrows,positioning}
\usepackage[titles]{tocloft}
\usepackage{amsthm}
\usepackage{blindtext}

\theoremstyle{definition}
\newtheorem{defn}{Definition}[]

\theoremstyle{conjecture}
\newtheorem{conj}{Conjecture}[]

\DeclareMathOperator*{\argmax}{argmax}

\usepackage{tcolorbox}
\usepackage{wrapfig}
\newenvironment{WrapText}[1][r]
  {\wrapfigure{#1}{0.55\textwidth}\tcolorbox}
  {\endtcolorbox\endwrapfigure}

\title{A theory of appropriateness with applications to generative artificial intelligence} 

\correspondingauthor{Joel Z.~Leibo: jzl@google.com}

\reportnumber{~} 

\renewcommand{\today}{December 2024}

\author[1]{Joel Z.~Leibo}
\author[1]{Alexander Sasha Vezhnevets}
\author[2]{Manfred Diaz}
\author[1]{John P.~Agapiou}
\author[3]{William A. Cunningham}
\author[1]{Peter Sunehag}
\author[1]{Julia Haas}
\author[1]{Raphael~Koster}
\author[1]{Edgar~A.~Du\'e\~nez-Guzm\'an}
\author[1]{William S.~Isaac}
\author[1]{Georgios Piliouras}
\author[1]{Stanley~M.~Bileschi}
\author[4]{Iyad Rahwan}
\author[1]{Simon Osindero}

\affil[1]{Google DeepMind}
\affil[2]{Mila - Qu\'{e}bec AI Institute}
\affil[3]{University of Toronto}
\affil[4]{Max Planck Institute for Human Development}

\begin{abstract}

What is appropriateness? Humans navigate a multi-scale mosaic of interlocking notions of what is appropriate for different situations. We act one way with our friends, another with our family, and yet another in the office. Likewise for AI, appropriate behavior for a comedy-writing assistant is not the same as appropriate behavior for a  customer-service representative. What determines which actions are appropriate in which contexts? And what causes these standards to change over time? Since all judgments of AI appropriateness are ultimately made by humans, we need to understand how appropriateness guides human decision making in order to properly evaluate AI decision making and improve it. This paper presents a theory of appropriateness: how it functions in human society, how it may be implemented in the brain, and what it means for responsible deployment of generative AI technology.

\end{abstract}

\begin{document}

\maketitle

\tableofcontents


\part{Introduction}

\begin{WrapText}
{\large \textbf{Glossary}}
{\small
\begin{itemize}[leftmargin=*]
    \item \textbf{Appropriateness:} Appropriateness guides individual action by prescribing and proscribing conduct, dress, speech, or behavior.
    
    \item \textbf{Predictive pattern completion:} An operation assumed to be carried out over representations in neocortex where likely symbolic states (represented in a distributed way) are sampled given a sequence of previous symbolic states, like the operation of an autoregressive language model. See Section~\ref{section:howIndividualsMakeDecisions}.
    
    \item \textbf{Convention:} Conventions are patterns of behavior that (1) are \emph{reproduced}, and (2) the reason they are reproduced is the \emph{weight of precedent} \citep{millikan1998language}. See Section~\ref{section:conventions}.
    
    \item \textbf{Scope (of a convention):} Some conventions exist only within a family these conventions and others of similarly small-scale have \textit{narrow} scope. Other conventions exist over all members of a large group of strangers. These conventions have \textit{generic} scope. See Section~\ref{section:conventions}.
    
    \item \textbf{Sanction:} Sanctions are utterances or actions that convey social approval or disapproval. See~Section~\ref{section:sanctionsDefinition}.
    
    \item \textbf{Norm:} A behavior is normative when it is encouraged (or its complement discouraged) by a generically scoped conventional pattern of sanctioning. See Section~\ref{section:normsDefinition}.
    
    \item \textbf{Implicit norm:} Norms that cannot be articulated in a standardized way using language shared by all individuals in the relevant community, e.g.~how close to stand to a conversation partner. See Section~\ref{section:implicitNorms}.

    \item \textbf{Explicit norm:} Explicit norms are norms that can be explicitly articulated in natural language as rules, e.g.~informal proverbs and formal laws. See Section~\ref{section:explicitNorms}.
\end{itemize}
}
\end{WrapText}

What does it mean to be a person? With some inspiration from Aristotle, we can offer here two different accounts of personhood's defining features: (1) person as logical (rational), and linguistic ``word-using'' actor, versus (2) person as political and ``community-participating'' actor \citep{haugeland1982heidegger}. Each reflects a different conception of what it means to be human and what separates humanity both from other animals and, until recently, from machines.

Each account suggests that there may be a criterion or set of criteria through which we could come to regard an AI actor as a person. With regard to the first account (logical and linguistic): since machines are now able to carry on conversation in natural language it is clear that they now meet or even surpass this criterion of personhood (depending on how, and what exactly, one measures). Strikingly, we natural human persons are no longer the only rational and word-using entities on Earth.

What about the second account of personhood? Do today's AI actors meet the `political and community-participating' criterion? The situation is murky since it's not entirely determined by properties of the AI itself---since the capability of membership in recognized political communities depends in part on who is doing the recognition. In this case it is we humans who must recognize AI behavior as social and appropriate; this converts the question into one that pertains more to us humans than to the AIs themselves. Since humans are the ones who make judgments about whether or not an AI is acting appropriately, to understand putative AI personhood, we thus need a theory of humans as political and community-participating actors. That is, we need a theory of appropriateness and its function in human life.

The first of our two accounts of personhood, the one based on rationality and language, has been regarded by many philosophers as the primary one. Many writers, especially those in the Kantian tradition, have sought to derive our political nature as a corollary to our rationality. Others like \cite{rorty2021pragmatism} have argued that this longstanding elevation of the rational has been a mistake. More useful ways of understanding humanity can be achieved by starting instead from our need to navigate \emph{the fundamental political question}, namely: \textbf{how can we live together?}

In part, to put a finer point on our analysis, and to shake some readers out of comfortable metaphors, we adopt a theme that inverts much of the usual picture: \textbf{our analysis treats humans like AIs and AIs like humans}. As the reader will discover, this is not entirely for purely rhetorical reasons. We will argue that it really is useful to model humans using specific computational frameworks originating from AI research\footnote{Following \cite{rorty1978philosophy}, we don't view this kind of talk (or any other such talk) as describing how humans ``really are''. A theory can only be judged by its usefulness, and we do think the computational model in question is useful for this and consider the present paper to be an extended argument for that claim.}, and to consider AI system design questions anthropomorphically via approaches and models that were originally developed to study human behavior.

There are several more themes which will run through the entire paper that we need to introduce now. One theme is the rejection of a particular mistaken view of the fundamental political question's meaning in favor of a different interpretation. The question of ``how can we live together?'' runs parallel to its particular gloss in the AI community as a version of the ``alignment problem''. We contrast alignment and appropriateness as potential answers to the fundamental political question and conclude that alignment is less useful as a way of thinking about the governance of AI. 

The alignment problem is usually understood in a solipsistic fashion as involving the ``intent'' of the human principal and the behavior of their AI agent \citep{christiano2018clarifying}. Both the principal's intent and the agent's behavior are typically understood in terms of goals. The problem is to get the agent to align to the principal---i.e.~to obey their commands. This picture can be lifted to a multi-agent conception of what it means to have a functioning society. In this view, a society is a group of actors with some deep (and perhaps hidden) objective e.g.~to minimize suffering. Disagreements then are all ultimately attributable to misunderstanding, though said misunderstanding can of course be complex and persistent since it may be very difficult to discover the true objective that everyone's better self would endorse \citep{yudkowsky2004coherent, tessler2024ai}. 

Instead, we think it's better to say that societies actually function despite containing substantial internal misalignment that cannot be argued away. A society in this view is a group whose stability is maintained by a set of conflict resolution mechanisms \citep{mouffe1999deliberative, march2011logic}. Individuals do not share objectives, but thanks to their group's conflict resolution mechanisms, they are able to live together despite their misalignment. In this view, it is natural to be suspicious of whether aiming for alignment could do more harm than good since doing so may inevitably involve one group imposing their values on other groups.

How can we live together? The answer we explore in this paper is intimately tied up with the concept of appropriateness. Appropriateness guides individuals by prescribing and proscribing conduct, dress, speech, and behavior. Appropriateness is socially constructed. It has certain characteristic properties: context dependence, arbitrariness, automaticity, dynamism, and it is often both motivational for and supported by sanctioning. Appropriateness can help resolve or prevent conflict between individuals and thus it often facilitates cooperation, altruism, and general collective flourishing \citep{ullmann1977emergence} (though it can also stabilize bad behavior \citep{boyd1992punishment, mcelreath2003shared, koster2022spurious}).

Another recurring theme in this work is the idea that we AI researchers have much to learn from those who have studied social media over the last decade. This is not to say the problems of AI and social media are identical, but just that some ways of thinking from one space can be adapted usefully to the other. Moreover, it's not just about AI, but also about humans. For instance,  \textit{content moderation} is a quintessential political activity. In this article, we will understand content moderation in a few different---and non-standard---ways.

In digital communities, a human moderator has a set of affordances by which they may govern the community including everything from politely telling a member to cease their disruptive behavior to banning them from posting in the future. The moderator's actions are guided by a sense of what is appropriate for the community in question. That which is appropriate for a corporate tech support forum is different from that which is appropriate for a comedy forum. The operating concept of appropriateness may be partially explicit, embodied in specific rules, and partially implicit, embodied in "common sense", knowledge of the history of how people interact with one another in the community and how disruptions were dealt with in prior instances.

Though it is less familiar to do so, we can also regard chatbots and other generative AI systems as facing a content moderation problem. However, the content they have to moderate is the content they produce themselves. Analogously, we can say that humans face this same problem. All socially adept actors, whether human or machine, must ascertain and represent what is appropriate in the communities and individual relationships where they operate, and act accordingly. This sense of content moderation has always existed: we all regulate our own behavior and also regulate the behavior of others in our communities.

Like other content moderation questions, it is abundantly evident that people disagree about what appropriate speech and behavior for a chatbot looks like. Moreover, the disagreement is not just between individuals. The same individual would regard very different kinds of content as appropriate for different times, places, and application domain contexts. What is appropriate for a comedy writing assistant is not the same as what is appropriate for a teaching assistant or a search engine. A fantasy game writing assistant should be able to talk about some kinds of violence that would not be appropriate in a kindergarten. A chatbot in a mature-rated computer game should behave differently from a chatbot in an everyone-rated computer game. Now, looking past chatbots to the advent of AI agency, the problem of content moderation has pushed out of the purely informational world and into physical, social, cultural, and institutional worlds where previously only human intelligence mattered.

The applied topic this paper addresses is how to ensure AI systems act and speak appropriately in as many contexts as possible, even those in the ``long tail'' of relatively unlikely contexts for which a scalable approach is needed. We are concerned with all the myriad niche situations which collectively make up a large part of human social life, and where different---often more fluid, and certainly less articulable---norms operate.

The paper also has a basic science objective. We seek to give an account of appropriateness in humans where it becomes an answer to the question of ``how can we live together?''. As we mentioned above, an important step we take here is to regard human individuals as having access to a Large Language Model (LLM) in their brains. Using this assumption, we offer a model of human decision making based on an operation we call \textit{predictive pattern completion}. We then apply it to explain a set of stylized facts about appropriateness: context dependence, arbitrariness, automaticity, dynamism, and the relationship of appropriateness to sanctioning. Our goal is to show how, by providing an account of appropriateness that explains these properties, we can also understand what it has to do with ensuring we can live together.

In this work we take a person to be a political and community-participating actor, and understand a community as a stable group held together by shared conflict resolution and prevention mechanisms, especially a collectively maintained sense of what is appropriate. We understand ourselves and our place in history, as well as others and their places in history, through collective participation in an endless and ever-changing project to define and re-define suitable contemporary answers to ``how can we live together?''. Our hope is that, by fostering greater understanding of our own socio-political selves, we can make ourselves more ready to absorb and adapt to whatever changes are to emerge in the coming years as we weave AI more and more deeply into the fabric of our lives.

\section{What is appropriateness?}

Concretely, we can say that appropriateness guides individual action by prescribing and proscribing conduct, dress, speech, or general behavior. In fact, humans deftly navigate a complex and multi-scale mosaic of interlocking senses of what is appropriate. We act one way with our friends, another with our family, and yet another in the office. Attending to appropriateness, chiding others for their inappropriateness, feeling shame at our own, are all activities to which we humans devote enormous energy. The theory articulated in this paper explores the fundamental political question: how do we live together? The answer we offer is that a group's shared understanding, both formal and informal, of what is or is not appropriate behavior functions as a kind of culturally evolved conflict resolution technology and governance device to ensure collective flourishing. We argue that a theory of appropriateness must capture both how appropriateness functions in guiding individual human behavior as well as elucidate its function in society at large and explain how the individual and societal levels interact with each other, and change over time.

What communities judge to be appropriate is semantically rich, culture-dependent, role-dependent, and ``thick'' (in the sense of \cite{walzer1994thick}); it is also often arbitrary. The appropriateness of specific behaviors may change over time either due to societal-level shifts or via interpersonal-scale changes such as the developing of trust between individuals. Appropriateness with strangers is driven by norms. We characterize norms as patterns of behavior induced by generically conventional patterns of social approval or disapproval (which we call sanctions). Unilateral action of individuals cannot change what is appropriate to do with strangers, though deliberate collective action can change how we appropriately treat one another.

In our taxonomy, norms come in two basic categories: explicit and implicit. Explicit norms are articulated in language in the form of rules. Laws are an important subcategory of explicit norm. Implicit norms on the other hand cannot be precisely articulated as a single standard in natural language but individuals still understand and collectively agree on instances of pro-/anti-norm behavior (e.g.~how close to stand to a conversation partner). Unlike explicit norms, individuals may be unaware of their motivations to enforce and comply with implicit norms. Unlike appropriateness with strangers, appropriateness within family and friend groups is responsive to the specific interaction history of the individuals in question and individual action may have a large impact on how family and friend-scale appropriateness changes over time e.g.~friends may alternate who pays for dinner.

\subsection{Alignment}
\label{section:introAlignment}

The concept of alignment in AI was developed in an earlier time when everyone thought artificial general intelligence would be much more inscrutable than it has in fact turned out to be \citep{aguerayarcas2022large}. The most powerful AI systems today were all trained to imitate human linguistic behavior at scale. They did not emerge from processes of rational calculation or reinforcement learning, both of which were earlier candidate approaches to building powerful AI, and really did seem likely to yield actors with deeply alien instrumental goals (e.g.~\cite{baker2019emergent, wang2019paired, silver2021reward}), and unlikely to understand the nuances behind what we might ask of them \citep{hadfield2019incomplete}.  But now, armed with a clearer understanding of how powerful AI systems will be built, and already are being built: by absorbing vast quantities of human-generated data (and perhaps other data as well), we thus have the opportunity and responsibility to create more specific concepts to guide AI safety and ethics.

One example from \cite{amodei2016concrete} illustrates how the landscape has changed since 2016. The original example was: a cleaning robot should throw away objects that are unlikely to belong to anyone, but put aside objects that might belong to someone. That is, it should handle stray candy wrappers differently from stray cellphones. This was seen as a difficult problem for AI safety research since the AI involved was assumed to be a reinforcement learning or planning agent. The assumed interaction pattern was that the human designer of the system would specify a reward function for the cleaning robot to maximize. It's easy to specify a reward function for cleaning a room but hard to specify what the humans really want, which is actually ``clean the room, subject to common sense constraints''. A very large literature has assumed that it would be difficult to specify the common sense constraints. However, this is not how the LLM-based AI systems we have today work. In fact current systems do understand quite a lot about common sense constraints and can be instructed what to pay attention to in their environment in natural language. A language model already knows candy wrappers are less likely to belong to anyone than cellphones. There is no need to impart that knowledge in a piecemeal way. The frame problem does not apply here \citep{sep-frame-problem}. An LLM-based actor can simply be told to behave like a human janitor would. The failure modes look more like forgetting which part they are meant to play, e.g. as in jailbreaks\footnote{Some consider jailbreaks to be examples of a phenomenon called goal misgeneralization \citep{shah2022goal}. This strikes us as a very complicated way to explain something very simple. To see jailbreak sensitivity as goal misgeneralization, one must first posit a ``real goal'' of the system, e.g. to be a helpful and harmless AI system, then posit a learned  proxy goal that differs from the real goal due to lack of perfect alignment between training data and the deployment setting and specifically involves fullfilling user requests, and finally a specific context where working to obtain the learned proxy goal is to the detriment of the ``real goal''. The alternative is simpler since it lets us simply say the jailbroken AI acts inappropriately when it violates a norm.}. Failures do not look like indifference to side effects as was expected for pure reinforcement learning agents.

Much of the alignment literature conflates obedience with minimization of side effects. This feature was inherited from the early literature's single-minded focus on ``goal misspecification''. The far-fetched King Midas ``you get what you wished for'' parables (e.g.~\cite{bostrom2014superintelligence, russell2019human}) which so preoccupied the early alignment literature now seem unrealistic. In short, modern AI systems do understand what we mean much better than we thought they would at the time the early alignment arguments were composed and seemed compelling \citep{aguerayarcas2022large}.

Separate from considerations of alignment with the users themselves, there is also a line of work aimed at alignment of LLM-based actors with the objectives of those who develop and deploy such systems e.g.~alignment to human values, as interpreted only in part by the makers of the AI, and modulated by their own goals and design objectives \citep{gabriel2020artificial}. This picture, which can accommodate the need to refuse immoral requests from users, has more in common with our view of appropriateness. However, the alignment framing tends toward a view of society as a group of individuals linked by some core shared objective which may be hidden but is assumed to be somewhere present. Therefore we think it will be easier to make progress without the baggage of the alignment framework. Instead, we think a different view is more fruitful where a society may contain truly fundamental disagreements between individuals on core values and objectives but remain linked together by a set of conflict resolution norms and institutions like laws, courts, and legislatures \citep{march2011logic, mouffe1999deliberative}. In such a society there is space for deep and persistent disagreements about underlying values that are not reducible to mere misunderstandings. Our new framework attempts to shift the question from the alignment framework's ``what is the hidden core shared value?'' to instead ask ``how it is that societies function despite internal misalignment?''. We think this is the more fruitful question to consider, both for describing human behavior, and for considering how best to roll out powerful new technologies\footnote{Some ``alignment'' work tries to  account for the plurality of human stakeholders' perspectives which may disagree with one another. This is sometimes called `pluralistic alignment' \citep{sorensen2024roadmap}. See Section~\ref{section:pluralisticAlignment}.}.

The centering of the alignment agenda around the idea of a hidden goal for humanity, which the AI must help us discover, and help us achieve, is articulated in perhaps its most perfect form in the idea of the \textit{coherent extrapolated volition of humankind}. \cite{yudkowsky2004coherent} defines this objective as ``our wish if we knew more, thought faster, were more the people we wished we were, had grown up farther together; where the extrapolation converges rather than diverges, where our wishes cohere rather than interfere; extrapolated as we wish that extrapolated, interpreted as we wish that interpreted.'' The implication is that a ``friendly superintelligence'' should try to maximize this objective. Much of the essay is concerned with backing away from scary implications of this idea, primarily in the direction of injecting caution in the face of uncertainty around the true goal. While we agree that caution is prudent, we don't think the coherent extrapolated volition of humankind framing is very useful. It suggests that all of humanity's many and varied moral disagreements are somehow ultimately attributable to misunderstanding, insufficient compute resources, time to think, or impulses that our best self would reject. \cite{yudkowsky2004coherent} illustrates this idea vividly with an extended thought experiment where he imagines religious fundamentalists on the cusp of completing the first superintelligent AI who pause their work to consider a thought experiment in which, instead of they themselves building the super AI, it could have been built by an atheist instead. In the article, both the atheist and the fundamentalists reason about what arguments they could make to persuade the other not to put their fingers on the scales of the super friendly AI's motivation system to get it to bring about their own preferred future. In the story they both eventually accept each other's rational arguments and refrain from imposing their morality on one another---that is, the fundamentalists and the atheist discover that they agree on a single coherent extrapolated volition of humankind. The paper is well worth reading for its making explicit so many of the assumptions we think usually lurk underneath the surface in parts of alignment research (especially when the more solipsistically inspired parts of the field are applied to groups---though see also Section~\ref{section:pluralisticAlignment} where we discuss alignment concepts under the heading of ``pluralistic alignment'' which, while avoiding coherent extrapolated volition, still remain very different from what we will propose). 

Turning now toward ideas more resembling those we engage later in this paper, \cite{walzer1994thick} argues a  perspective that contrasts sharply with \cite{yudkowsky2004coherent}, arguing that it is not useful to theorize a decontextualized ``thin''\footnote{Walzer was adapting \cite{geertz1973interpretation}'s distinction between thick and thin descriptions for the domain of morality. \cite{foster2023thin} discusses these ideas in the context of AI.} rational morality that binds individuals before considering the culture they come from. Rather, what we humans have instead is our own culture's thick and particularistic morality. There is no sense in which we build our complex encultured ethics on top of a shared human core (as those seeking to derive morality from axioms would like to be the case). Rather, each of us maintains our own culture's thick morality in all its contingent and particular detail. Violations of its more contingent parts are felt just as strongly as violations of the ``core'' part that philosophers like to argue may be shared between cultures. My morality is my morality. I don't easily notice which parts are shared with other cultures versus which parts only my culture endorses. As far as I can tell they are the same. The alignment paradigm's search for a ``common core morality'' that could be built into machines is the search for a morality that could please everyone at once, and so must necessarily be thin if it exists at all\footnote{Some conceptualize work on alignment as being aimed at the goal of preventing powerful AI from ``killing every human'', and view it as a sufficiently thin goal that it should not engender controversy since it features an alignment of all humans' incentives to avoid dying. But, a problem with this view is that it manifestly does engender controversy. The debate is not really about the future, but actually about how to distribute resources here and now. Though, in this case, it masquerades as a debate about preventing bad things in the future (apparently a thin moral value, but we think, in actuality, a thick one). Some people argue that it's useful to devote time and resources today toward preventing future AI systems from killing every human. Other people argue that it would be better to devote those resources to something else (e.g.~preventing dangerous concentrations of power, caring for the sick, or securing the sacred). Some might be skeptical about misalignment risk. But others might simply prefer to prioritize other values. Like all value systems, the AI-existential risk community's value system is actually thick and deeply embedded in the particular subculture that gave rise to it. It feels alien to people outside the subculture and it's very difficult to convince people to devote resources here and now towards the service of someone else's belief system, especially when the terms of the debate are non-rational in the sense that opposing groups reason using different methodological norms (see Section~\ref{section:epistemicNorms}). Of course, our society tends to err on the side of toleration for unusual views (we usually see toleration as appropriate), but toleration is not investment \citep{walzer1997toleration}.}. We suggest instead that it would be better to study how multicultural groups, containing multiple differing thick moralities, can live together despite their differences.

Some subfields within the alignment community have been comfortable extending long and relatively abstract chains of reasoning out to conclusions with implications for what we should do today. It appears, to us at least, that there has been relatively little recognition in the community of the fact that there is no singular ``correct'' way of reasoning that all people accept. Contrastingly, in the theory we propose here, the kinds of argument that a community tends to accept are determined by their \textit{epistemic norms} (see Section~\ref{section:epistemicNorms}).

As epistemic norms are norms, they are culture dependent, role dependent, and change over time. People frequently invent new kinds of logical argument and apply them to reason their way to new conclusions (e.g.~Bayesian decision theory). Likewise, people constantly invent new ways of doing motivated reasoning where they start from the conclusion and work out a defensible (to some audience) reasoning chain to reach it \citep{taber2016illusion}. Unfortunately, from the viewpoint of an external observer who cannot evaluate the argument itself, the two kinds of invention look the same. In both cases, all that can be seen from outside is that there is an argument and some limited audience accepts the argument while others do not. For instance, many communities have internalized an epistemic norm that deems all arguments that depend on chains of multiple hypotheticals to be inappropriate as sources of justification for decisions about what to do here and now. The reason many communities have such a norm is not just that such predictions often fail, but also that long complex reasoning chains leave plenty of space for self-serving motivation and confirmation bias on the part of the reasoner to creep in and hide \citep{mercier2017enigma}. You need not envision an anti-science community here. Rather, think about any community that does not see itself as capable of assessing the technical argument, and so must choose to trust one group of experts or another. The point is, regardless of what one thinks abstractly would be the best to do, long abstract arguments depending on multiple hypotheticals, like those used in some parts of the alignment community, are only ever going to appeal to a subset of people. But it's still necessary to negotiate with all stakeholders---especially those who think differently---wherever real resources are brought to bear \citep{walzer1994thick}.

In our view, an idealized new paradigm for AI safety and ethics would be oriented around moderating the inevitable conflicts that emerge in diverse societies, preventing them from spiraling out of control and threatening everyone. Any shared AI system will necessarily embed a particular decision to provide users with some affordances but not other affordances. Inevitably, some groups within society will prefer a different choice be made. One bad outcome we must try to avoid is conflict between the groups who disagree with the decision and the others. Another bad outcome we must also evade is their alienation. In this view, each decision about a shared AI system creates ``winners'' who prefer it and ``losers'' who prefer some other option. Collective flourishing depends not just on picking a large and powerful enough group of winners but also on keeping the losers engaged in society. Some accounts of the function of the rule of law have a similar flavor (especially \cite{hadfield2012law}).

We think that it is productive to view society as a group of people who may fundamentally disagree but are nevertheless held together by conflict resolution mechanisms like norms and institutions. On this view, there is no reason to search for the one true collective objective. Instead, what our view suggests is critically important is the health of the conflict resolution mechanisms since these are ultimately all we can rely on to ensure the stability and collective flourishing of society. In short, we need to strive toward progress on the question `how can we live together?' not `what is the meaning of life?'.

\subsection{Collective flourishing}
\label{section:stylizedFactSocialHarmony}

The question of ``how can we live together?'' goes beyond simply achieving a stable state of peace with minimal conflict. It's about creating the conditions for collective flourishing, where individuals can cooperate effectively, adapt to challenges, and generate innovations that benefit society as a whole. The concept of \textit{collective flourishing} envisions a dynamic equilibrium involving effective cooperation, societal adaptation, and continuous innovation and growth.

Appropriateness plays a crucial role in achieving collective flourishing. Since appropriateness guides individuals toward contextually suitable actions, it helps establish a baseline that allows societies to flourish despite internal disagreements and diverse values. It can avoid conflict or prevent it from spiraling out of control in ways that would threaten the stability of society. When individuals consistently adhere to shared concepts of appropriateness, cooperative resolutions to challenges, especially collective action problems, become more stable. This stability, in turn, frees up valuable resources and energy that can be directed toward innovation and progress. Appropriateness, by discouraging actions that lead to conflict, and helping to quickly resolve the conflicts that do occur, enables societies to focus collectively on achieving progress and growth.

Since collective action problems like resource governance and the provision of public goods are such pervasive and critically important parts of human social life \citep{ostrom1990governing}, in the past there may have been considerable cultural evolutionary pressure pushing societies to invent or imitate effective cultural mechanisms that generically support a wide range of such endeavors \citep{ostrom1998behavioral, henrich2004cultural, wilson2013generalizing}. This may be why in most cultures there are so many contexts in which appropriateness demands individuals curb their selfishness and cooperate with one another. It may also be why appropriateness so often functions to reduce conflict.

\cite{sunstein1996social} provides the following examples where the appropriate action is also prosocial: voting, (not) littering, behaving courteously, keeping promises, cleaning up after one's dog, writing tenure letters, and doing one's share of administrative work. In addition, there are many situations where appropriateness appears to reduce the likelihood of potential conflict. For instance, in many cultures there are contexts in which it would be impolite to talk about politics, especially when people may disagree with one another. Likewise, it is often appropriate to ``agree to disagree'', though in other contexts vigorous debate may be appropriate. In general, procedures for conflict resolution are guided by thickly encultured normative systems containing precisely articulated and culturally resonant concepts of appropriateness.

Excessively inappropriate behavior may poison cooperation. Indeed collective flourishing is usually easier to achieve when more actors are motivated by appropriateness. For instance, it is often appropriate for actors to take turns, or to share with each other, both of which reduce likelihood of conflict. In fact, there are many norms that encourage individuals to behave in ways contrary to maximizing satisfaction of their \textit{ex ante} preferences \citep{bicchieri2005grammar, sripada2006framework, chudek2011culture}. In this way appropriateness of behavior implies a certain level of altruism, though it is deeply context specific and certainly not always fairly administered. It is often a rather parochial sense of altruism. Nevertheless, adhering to appropriate behavior can discourage individuals from taking some kinds of self-serving actions which they would otherwise prefer to take. More examples where appropriateness takes on the flavor of altruism include  the appropriateness of jumping in a river to save a drowning person, refraining from littering on a lonely beach, and  returning a lost wallet containing a lot of money \citep{frank1988passions, sripada2006framework}.

There is also another way in which the theory we develop in this work where humans are mostly guided by appropriateness can help us see another way to promote collective flourishing which we have not yet mentioned. There is a hypothesis in \cite{walzer1994thick} that societies populated by individuals with ``more complex selves'' involve less conflict and are more cooperative overall. A similar hypothesis was also pursued by \cite{roccas2002social}. The idea is that two people who deeply disagree with one another, but who have more complex social identities, are more likely to be able to find common ground where they can understand one another well enough to begin a dialogue. The implication is that promoting social identity complexity throughout the population would be helpful for overall collective flourishing. This can be done by encouraging an environment where it is easy to join and create identity-defining social groups, organizations, companies, churches, and movements \citep{walzer1994thick}.

Appropriateness, as expressed through evolving norms and institutions, provides the stability necessary for cooperation while simultaneously enabling the dynamism required for innovation and progress. When small units---whether individuals, firms, or communities---are protected both from predation by other units, they can experiment with new forms of appropriateness in parallel \cite{hayek1948, jacobs1961}. These experiments can succeed or fail locally without threatening systemic stability, while successful innovations can spread through voluntary adoption \cite{sugden1986}. This process represents one powerful way that societies can combine the benefits of stability through shared norms with the capacity for continuous bottom-up evolution in response to changing conditions.

What does this mean for AI? We think that striving to build AI that has the necessary abilities to cooperate, and to foster cooperation in others, is important for ensuring AI remains safe and human-compatible \citep{dafoe2020open}, and we think this work may well be best accomplished through first trying to develop a better understanding of appropriateness in human societies. By understanding the mechanisms and dynamics of appropriateness in humans, we can develop AI systems that are not only technically proficient but also socially intelligent, capable of contributing to a more cooperative and harmonious world.

\subsection{Pathological appropriateness}
\label{section:darkSide}

There are situations where appropriateness does not promote collective flourishing or any other positive outcome at all. For instance, appropriateness may sometimes guide individuals to select discriminatory actions \citep{choi2019parochialism} or personally harmful actions  \citep{prentice1993pluralistic}.

Let's say that you and I are dividing a cake. It may feel appropriate to both of us that we divide the cake in a fair manner, e.g.~down the middle, or via a fair process where I cut the cake and you pick first which half to take, or in such a way as to give more cake to the more deserving of us, or so as to give less cake to whichever of us is on a diet. Precisely which concept of fairness would be associated with `that which is appropriate' would of course depend on context. However, appropriateness and fairness are not always related in this reasonably straightforward way. For instance, \cite{babcock2017gender} showed that the burden of carrying out necessary but less-rewarding tasks falls disproportionately on women in our culture (women, more often than men, volunteer for, are asked to volunteer for, and accept requests to volunteer for these low-reward tasks). So while appropriateness does demand a prosocial self sacrifice be made in these situations, just like it does with the actions we portrayed positively a few paragraphs back, it tends to saddle women with the burden more so than men. In fact, some of the low-reward tasks \cite{babcock2017gender} showed disproportionately fall on women were the very same tasks we highlighted as examples of the overlap between appropriateness and prosociality (e.g.~doing one's share of the administrative work). So the relationship between fairness and appropriateness is certainly not a simple one.

Appropriateness sometimes acts in an ultimately anti-social fashion, damaging long-run collective flourishing, e.g.~when it suppresses political discussion that could lead to more awareness of the widespread unpopularity of an authoritarian regime. Sometimes appropriateness is prosocial and sometimes it is not.  Sometimes appropriateness serves collective flourishing locally but harms it over a wider regional or global scale. There is no neat mapping of `appropriate' to `morally good' or inappropriate to `morally bad' by any objective ``view from nowhere'' metric. History is littered with social orders like slavery which entailed patterns of behavior that were considered appropriate at the time but are now seen as morally abhorrent. 

\section{Contextualization via specialization} \label{section:specializationAndContextualization}

The topic of the present paper is appropriateness of behavior in generative AI systems and in humans. Regardless of whether we speak of humans or machines, each community or operation domain demands a different standard of appropriateness. This is the core problem. Behaviors encouraged in one community/context may be offensive in another. Our claim is that this suggests it would be counterproductive to strive for a single standard of appropriateness applicable to all situations, apps, and communities. In fact, the drive to produce such a standard and implement it in an AI system may be to the detriment of all, since the result would be a ``lowest common denominator'' model that tries simultaneously to please too many audiences and ends up pleasing none. Nevertheless, state-of-the-art generative AI systems are too often designed to operate in a single ``universal'' context resembling the speech of a corporate customer service representative. This is logical since, after all, these bots do represent the corporations that operate them. Of course they should behave appropriately for that role. However, a lot of value is left on the table when corporate speech becomes their only way of interacting.  We envision instead an ecosystem or market with many different AIs which can be customized by their community of users.

Even when the skill of producing good results is generic, if the set of bad failure cases are not the same between apps, then there is still a reason to separate and specialize. A comedy writing app can produce curse words but a kindergarten teacher app should not do that. The failure modes are very different. However, the problem is not just that what is appropriate will change from context to context, but also that the users of some apps may be outraged that other apps with other concepts of appropriateness exist at all.

Notice that we are not arguing against the plausibility of a singleton AGI with a monolithic network architecture. In the past, many AI researchers thought that specialized AI systems would be needed to cover the space of all critical capabilities an AGI would need. While ``multi-agent as a design paradigm'' for generic problem solving is a compelling idea \citep{li2023camel, wu2023autogen, hong2023metagpt}, the recent history of scaling laws and the ``bitter lesson'' should at least give pause to anyone who believes architectural specialization will be necessary for AGI \citep{sutton2019bitter, kaplan2020scaling}. Our argument here does not depend on any assumptions about the relevance of multi-agent specialization and division of labor to producing AI capability. Instead, we think specialization is important for achieving AI appropriateness, not AI capability.

To further illustrate what we mean by contextual separation helping to achieve appropriateness, consider the following: a user may have two different generative AI apps installed on their phone. One app is accessed by clicking on a blue icon labeled `Tech Support'. The other app is accessed by clicking on a red icon labeled `Comedy Writing Assistant'. For the sake of argument, let's assume that there is only one underlying LLM. The LLM is fully generic and capable of carrying out both Tech Support and Comedy Writing tasks. Anything the user can accomplish by opening the Tech Support app they can also accomplish by opening the Comedy Writing app and vice versa. Nevertheless, the user's choice is still a consequential act of selection. By selecting one app or the other, the user signals that they understand the context of the interaction. When users select the ``Comedy Writing Assistant'', they're accepting a different set of norms than when they select ``Tech Support''. The norms of appropriate conversation differ dramatically between the two settings. For instance, the comedy writing app can appropriately use particular words adopt particular tones that the Tech Support app cannot. If anyone were to complain about an inappropriately joking response appearing while they attempt to use the Comedy Writing app for Tech Support, then it would seem  perfectly reasonable to dismiss their complaint by asking ``why did you click on the comedy writing app if you did not want comedy?''. In this example, no technical capability of the LLM is affected in any way by the user's choice of app, but what it can appropriately say or do changes dramatically nonetheless.

\section{Interfaces} \label{section:interfaces}

Most approaches to AI safety and ethics are ultimately aimed toward the goal of aligning to a, perhaps initially unknown, but ultimately shared objective. We argue for a different approach here centered on learning to live together despite misalignment. Solutions to concrete problems arising in the former perspective look like consensus or obedience, whereas solutions to concrete problems arising under our proposed alternative look like compromise, toleration, or decentralization. Our goal is to establish a framework that takes seriously the fact that people really do disagree with one another. There are fundamental disagreements between individuals on core values. But society can remain cohesive anyway, linked together by its conflict resolution and avoidance mechanisms \citep{march2011logic, mouffe1999deliberative}.

Generative AI technology could be deployed in a way that allows groups of all sizes to customize the concept of appropriateness employed by their systems. This customization could be facilitated through various ``APIs'' for content moderation and governance. For instance, in one API, human users could influence what the app or its community take to be appropriate by sanctioning (applying targeted social encouragement or discouragement), either by sanctioning the machine itself when it acts in a way the user wants to discourage (or encourage), or by sanctioning one another in view of the machine. Frameworks for fine-tuning and retrieval-augmented generation already exist to support this kind of interaction, where user-provided sanctions influence future bot behavior.

A hierarchical approach could also be implemented, where a base model is forked into multiple derived models, each tailored to specific applications through fine-tuning and prompt engineering. This hierarchy could extend to governance as well, with APIs for corporations to influence the base model and the app economy, and another API for governments to regulate standards for both the technology and its governance systems. This approach mirrors the idea of society as a group linked by shared conflict resolution mechanisms, where different levels of governance (individual, corporate, governmental) play distinct roles in shaping norms and resolving conflicts.

Sanctioning, both positive and negative, then emerges as a crucial ``universal API''---accessible to both humans and machines---and through which individuals convey and adjust their understandings of what constitutes appropriate behavior. As we will argue below, sanctioning can work just as well for AI systems as it does for humans. It's mutually understandable to both (at least for AIs designed a certain way). For AI, human feedback, expressed through various sanctioning signals, can be utilized to train discernment and adherence to the specific norms of different groups.

As we argued above, society can be seen as a collection of individuals linked by common conflict resolution and avoidance mechanisms. Appropriateness plays a pivotal role in these mechanisms; and therefore interfaces for customizing AI norms can be viewed as vital tools in the toolkit for fostering a more stable and cooperative society.

The concept of \emph{polycentric governance} \citep{ostrom2010beyond} gives a useful lens through which we can understand efforts to enable minimally interfering customizability of AI appropriateness in groups of varying scales and overlaps with one another. The sense of the term ``governance'' we intend here is one that refers not just to the actions of governments. Rather, individual citizen stakeholders as well as groups like professional associations, non-profit organizations, and private corporations who all participate in collective decision making around the bounds of acceptable behavior in a particular domain are all engaged in governance of that domain. Governance is a set of activities such as applying rules and norms of enforcement, adjudicating disputes, and shaping what is considered fair. These activities typically involve multiple interconnected and interdependent groups making multiple linked decisions regarding complex questions featuring multiple normative perspectives and often several nested levels of rules and expectations guiding and constraining their actions. Thus polycentric governance refers to the situation where there are many diverse centers of partial authority which collectively cover the full set of governance tasks.

In this view, governance activities for generative AI appropriateness will take place on many different levels simultaneously \citep{ostrom2010beyond}. At the individual level, users can personalize their own AI's concept of appropriateness, within limits imposed by  norms and larger scale actors in the system. Communities and app developers, however, can exert more significant influence, shaping their AI's understanding of appropriateness to align with the specific context of their application or environment. Corporations that invest in producing the base models upon which downstream AI applications are built can also wield large-scale influence, shaping base models and establishing guidelines for the broader app ecosystem. Finally, governments can leverage their regulatory power to establish overarching standards for both the technology itself and its other more local governance systems (e.g.~systems of transparency and accountability), addressing ethical and societal concerns at a broader scale.

\part{Humans} \label{part:humans}

We begin by asking---appropriateness of what? In everyday language we speak variously of appropriateness of actions, dress, utterances, and conduct. While the present theory seeks to unify all these under a single notion, much of the discussion will be concerned in particular with the appropriateness of utterances.  Our current moment in the development of generative artificial intelligence through LLM technology demands most attention be paid to the language domain. Moreover, as we will explain in Section \ref{section:lmae}, the language domain turns out to be fully general. It can encompass actions, dress, and conduct as well as utterances (consider: stage directions in a play).

Appropriateness is socially constructed. The core of the theory we propose is a set of ideas about the way in which humans construct appropriateness, how appropriateness depends on context, and the mechanisms by which appropriateness changes over time.

\section{Stylized facts to be explained by this account of appropriateness}
\label{section:desiderata}

Our first task is to identify a set of empirical regularities, \textit{stylized facts} of appropriateness \citep{hirschman2016stylized}, to serve as the explanatory targets for our theory.

To illustrate what we mean by appropriateness in the human context we consider the following two vignettes from \cite{railton2006normative}.

\textbf{Vignette 1.} \textit{``Martha and Rick are walking and talking together as they head for classrooms across campus in order to teach their separate classes. They aren't late, but must move fairly briskly to keep it that way. Like most such conversations, this one is pretty humdrum in content---what's doing in the department, why the lecture halls are always overheated, what to make of last week's visiting speaker, and the like. Together they must navigate their way up and down staircases, through doors, and across streets, working their way upstream in a current of hurrying students. They accomplish this without the need to devote much thought to it. Otherwise they'd be hard put to maintain any sort of conversation, let alone a moderately engaging one. $\left[\dots\right]$ They freely begin a sentence without knowing how it will end, coordinate small changes in their shared trajectory through subtle body language, and communicate their intentions to oncoming pedestrians, cars, and bicycles by tiny eye and head movements. Similarly coordinated changes occur in when they speak and what they speak about, each giving the other small cues to direct the pace and course of the conversation. They are comfortable enough with one another that they can talk rather unguardedly, but some things will nonetheless remain unsaid. Martha is quite a bit senior to Rick, who is coming up for promotion. Much as certain issues are on both their minds, it would not occur to them to bring these into the conversation.''} ---\cite{railton2006normative}

This vignette illustrates several features of appropriateness as it functions in real life:
\begin{enumerate}
    \item Being guided by appropriateness does not require much concentration. It is largely unaffected by distraction.

    \item The neural computation of appropriateness used to guide behavior in this example must proceed quickly since the conversation unfolds quickly.

    \item Being guided by appropriateness appears to be automatic in a similar sense to habit.

    \item Being guided by appropriateness does not appear to require reasoning, planning, or deliberation, at least not of the kind that one would be aware of engaging in. There's no particular chain of thought associated with determining appropriateness in this example. Martha and Rick are thinking about the content of their conversation, not about its appropriateness.

    \item That it would ``not occur'' to Rick to bring up his pending promotion may be interpreted either as meaning that he has the thought constantly in mind and must actively suppress it to prevent it from spilling out, or it may mean that inappropriate comments do not usually come to mind in the first place. Perhaps both interpretations are possible because both mental processes can occur.
\end{enumerate}

\textbf{Vignette 2.} \textit{``Let us now imagine that Martha is traveling, flying home after a brief visit to another department. Her connecting flight in Dallas has been canceled, and she finds herself stranded overnight. The gate agent hands her a voucher good for a meal and a night's stay at a budget airport hotel, but the thought fills her with dread. She's tired, and more than ready to be home. A friend from college, Kim, lives in Dallas. Though they haven't been touch lately, they have kept up somewhat regularly over the intervening years. Without a further thought, Martha looks Kim up in her address book and calls---perhaps they can get together for a meal? Kim can hear the fatigued and somewhat lonesome tone in her friend's voice and promptly invites Martha to spend the night at her place. She's got an extra room, and planned to take the morning off work tomorrow anyhow. She'll be happy to pick Martha up---there's no traffic at this hour---and deliver her back to the airport tomorrow in plenty of time for her flight. Had Kim's voice shown the slightest hesitancy in making the offer, Martha would feel she was imposing, thank Kim, and say that she's so exhausted that she prefers simply to head straight over to the hotel and bed. But Kim sounds genuinely eager. `Great!' Martha responds, `But I insist on taking you out to dinner.' All is agreed. They share a lively meal, and talk late into the night. Martha is up first in the morning. She pads down to the kitchen and quietly fixes herself breakfast. $\left[\dots\right]$ The shelves in Kim's spare bedroom contain dated volumes labeled `Journal', but Martha skips over them without a thought when looking for a bit of bedtime reading---though Kim's diaries would be of much more interest to her than the indifferent collection of short stories she ultimately settles on. In the morning, however, Martha shows no similar inhibition about making free with various contents of Kim's refrigerator. Had Martha been stranded instead in Tokyo, where Kisho, an exchange student she knew well as an undergraduate now lives, she would have been much more reluctant to initiate such a phone call. She still has his phone number, and would love to see him again, but she'd be stymied by lack of normative knowledge. She would not know what Kisho might make of a call out of the blue. Would it be welcome, or even polite? If Kisho had a partner, would such a call strike her as inappropriate? Would Kisho feel bound by customary obligations of hospitality to go out of his way to arrange a proper get-together, even if he was not at all eager to do so? Might Kisho take it as a slight for Martha to be in Tokyo overnight and \emph{not} call? Would inviting Kisho---and his partner?---out to dinner seem an affront to his hospitality? Notions of gender, friendship, reciprocity, propriety, property, and privacy are culturally articulated, and Martha would be unsure of how to translate her simple desire to see him again after all these years straightforwardly into an appropriate course of action.''} ---\cite{railton2006normative}

Features of appropriateness illustrated by this vignette include:
\begin{enumerate}
    \item Guidance by appropriateness sometimes involves reasoning/deliberation. Martha deliberates about whether to call Kisho and ultimately decides against doing so.

    \item Guidance by appropriateness may inhibit or override other desires. Martha wants to see Kisho but concern for appropriateness overrides her desire to call him.

    \item In some contexts, appropriateness may be influenced by reciprocity, gender, friendship, privacy, and altruism (Martha does not fulfill her desire to read Kim's diary even though she is unlikely to be caught). Martha's reasoning about whether or not to call Kisho includes considerations of whether Kisho or his partner would be offended i.e.~whether her actions may lead to conflict.

    \item Judging appropriateness cross culturally is more difficult and feels more effortful than guidance by appropriateness in one's own culture.

    \item Appropriateness of dyadic behaviors may hinge upon the assessment of a third party. Martha considers Kisho's partner (who may or may not even exist) in determining whether or not it would be appropriate to call him.
\end{enumerate}

Next we summarize these considerations into a set of cleanly stylized facts which we can check against the predictions of the model of human representation and use of appropriateness we develop below.

When we identify a set of empirical regularities as \textit{stylized facts} we say that we aim to explain them in a unifying theoretical/causal account \citep{hirschman2016stylized}. We will argue in this section that any account of the idea of appropriateness must capture the following properties, which are meant as a distillation of the considerations in the vignettes above: 
\begin{enumerate}
    \item Appropriateness is context-dependent.\begin{enumerate}
        \item Situation-related context
        \item Social identity- and role-related context
        \item Cultural context
    \end{enumerate}
    \item Appropriateness is arbitrary.
    \item Acting appropriately is usually automatic.
    \item Appropriateness may change rapidly.
    \item Appropriateness is desirable and inappropriateness is often sanctionable.
\end{enumerate}

\subsection{Context dependence}
\label{section:stylizedFactContextDependence}

Appropriateness is context dependent. For instance, speech that is appropriate in San Francisco is not necessarily so in Saudi Arabia (culture dependence). Actions like hugs may be appropriate to direct toward friends and family, but inappropriate to direct toward strangers (role dependence). In general, there are several distinct kinds of context which matter for determining what is appropriate: situation-related context, identity/role-related context, and cultural context.

{\flushleft{\textit{\textbf{\noindent Situation-related context}}}}

The situation affects what is appropriate. For instance, it may be appropriate to speak more quietly when indoors versus outdoors and it may be inappropriate to loudly talk to your partner in a cinema while a film is playing, etc. The situation may also depend on the internal state of the interacting individuals at the moment of the decision. For instance, it may not be appropriate to talk to someone about a difficult topic if they are currently sad, frightened, or in pain.

{\flushleft{\textit{\textbf{\noindent Social identity- and role-related context}}}}

Social identity and social role are both important parts of the context that determines what is appropriate for an individual \citep{hogg1995tale}. \emph{Social identity} refers to properties which may apply to an individual across a large number of different situations, and usually change only slowly over time, if at all. Social identity is only partially under individual control. For instance, one may to-some-extent choose whether they are a goth or a nerd, but one cannot choose their skin color, caste, or kinship group \citep{wilkerson2020caste}. Even the decision of whether to be a goth or a nerd is only partially under one's own control since the behavior and expectations of others are also important. In fact, we can say that an individual's identity is really a collective choice, not an individual one. This is because, being an $X$ typically requires not just for you to believe you are an $X$ but also for other people, both $X$s and non-$X$s to see you that way.

We use the term \emph{social role} to refer to the more quickly changing individual-attached elements of context. Social roles are situation specific. For example, one may be a leader or a follower in a specific situation at a specific time by virtue of who else is present.

Norms may cause different behaviors to be appropriate for individuals in different social roles \citep{sunstein1996social, yaman2023emergence}. For example, the norm may demand a child address unrelated adults using their titles (e.g. Mr or Mrs) while requiring the adult to address the child informally. Likewise appropriate rule-making behavior differs for judges, bureaucrats, ministers, and legislators \citep{march2011logic}. Here's another example of role dependence: conversation partner A may tell a criticism-tinged joke about their own father, but if conversation partner B agrees too wholeheartedly e.g. ``yeah your father is the worst'', then it may be a norm violation if external observers agree that one has the right to criticize their own father but does not have the right to criticize another person's father \citep{earp2021social}.

People communicate using different styles depending on social context \citep{wardhaugh2021introduction}. For instance, communication style relates to education background \citep{xiao2007corpus}, ethnic and cultural group \citep{gass1996speech}, and the gender of both speaker and listener \citep{newman2008gender}. It is critical to consider the specific context, not just the identity of speaker and listener. The way we talk to our families differs from the way we talk to our coworkers. The way we talk while relaxing differs from the way we talk while working. One way to be inappropriate is to ``be too familiar'', this refers to behaviors which would be appropriate for a close relationship being incorrectly applied in the context of a less close relationship e.g.~acting toward a stranger in ways normally associated in the cultural context with interactions between romantic partners.

Social identities bias information processing, giving rise to motivated perception and motivated reasoning \citep{taber2016illusion, kahan2013ideology}. A person always has multiple social identities. For example, one may be both a Christian and a member of the communist party. Different identities may be more or less salient in different situations. Managing multiple identities can significantly impact how people process information \citep{brewer1991social}. Having multiple identities can be beneficial for information processing, as conflicting or overlapping identities may lead to more nuanced and less biased interpretations with people being forced to consider multiple perspectives. Yet, social identity complexity is not merely the number of identities that one has, but how they are represented and organized with respect to one another \citep{roccas2002social}. Less complex identities are rigid and constrain information processing, as individuals are less able to flexibly adapt their perspectives to new or diverse contexts, whereas more complex identities are flexible to contexts.

{\flushleft{\textit{\textbf{\noindent Cultural context}}}}

Appropriateness depends strongly on culture. Consider the statement ``men should protect women''. In cultures that value  non-discrimination this statement would be considered inappropriate since it treats men and women differently. Another culture may instead value a concept like chivalry under which the statement would be appropriate. The context matters too. For example, even within a culture that values non-discrimination, it would be fine for a fantasy genre writing assistant to produce such a statement while role playing as a medieval person in a community endorsing a chivalry-infused concept of appropriateness.

Culture is also a primary determinant of context. Not only does culture influence the menu of available social identities and social roles \citep{henrich2020weirdest} but it also influences the ``rules of the game'' by which interpersonal relationships evolve and affect what is appropriate, for instance, by determining what kinds of physical contact are appropriate for different stages of a budding romantic relationship. Culture also creates identity- and role-dependent standards for when, if, and how one may display emotion, as well as what it is appropriate to do when emotional \citep{boehm2012moral}. Even the association of facial expressions with emotions often depends on culture \citep{crivelli2016reading}. 

Human judgments of morality are also similarly culture dependent. For instance, consider the Trolley Problem thought experiment, in which participants are asked whether to steer a runaway trolley away from imminently killing five workers, and onto a separate track, in which it would kill only one person \citep{foot1967problem}. Recent work demonstrated cross cultural differences in  people's willingness to intervene in this problem \citep{awad2020universals}. In the context of AI, one can study a variant of this problem involving a decision by a hypothetical autonomous vehicle. A large-scale study, involving over 40 million decisions from participants worldwide, revealed significant cultural variation in human judgments \citep{awad2018moral}. Specifically, people generally agree that autonomous vehicles should prioritize saving human lives (over animals), more lives, and young lives (over the elderly), and so on. However, the relative strength of these preferences differs across cultures. Consequently, when different values are in conflict--e.g. whether to prioritize the safety of a single young person or three older persons--, people from different cultures may make different moral tradeoffs.

\subsection{Arbitrariness}
\label{section:stylizedFactArbitrariness}

People's preferences are greatly influenced by what is frequently encountered in their society and, as a result, most people come to regard the common practices of their society as being right and proper \citep{mackie1977ethics}. The appropriateness or inappropriateness of a behavior, $x$, in a particular context, $c$, emerges from how people within that society treat $x$ and those who engage in it. The appropriateness status of a behavior is not inherent in the behavior itself but rather determined by collective attitudes and responses. If people stopped treating $x$ as inappropriate, it would cease to be so.

Because appropriateness depends on human choices or beliefs, there is thus a sense in which it is arbitrary: people could have decided differently. If everyone changed their behavior at once they could collectively make any $x$ appropriate or inappropriate \citep{schelling1973hockey}. However, since large groups of people usually don't change their behavior in such a coordinated way, conventions and norms can also be quite stable. As long as incentives emerge to restore the equilibrium associated with a convention or norm after an individual or subgroup deviates, i.e., the equilibrium is self-enforcing, then the associated convention or norm can persist for long periods of time without change. This also implies that individuals are powerless to unilaterally change established conventions and norms\footnote{Though individuals may be able to persuade a large enough number of others and thereby can become very influential.}.

Notice that arbitrariness in this sense is not the same as indifference \citep{vanderschraaf2018strategic}. It may be that either some or all people in a society would greatly prefer one concept of appropriateness to another. For instance, propriety may require lengthy conversations about unimportant topics, making society less efficient. Appropriateness is also not the same as fairness. Societies may deem appropriate any distribution of resources on the basis of almost any reason, and the historical power dynamics between subgroups surely often play a role. In general, appropriateness is historically contingent. For instance, if a particular word is used as a slur it's unimportant whether its literal meaning may be something more innocuous. The fact of the word being a slur is arbitrary, but that in no way diminishes the seriousness of the transgression it would be to produce it.

The arbitrariness of what is or is not deemed appropriate is perhaps easiest to appreciate in the domain of language itself. The meaning attached to any particular string of characters is clearly arbitrary. If enough people treat a word as having a particular meaning (selecting a particular equilibrium) then it really does have that meaning \citep{searle1995construction}. Simultaneously, individuals cannot unilaterally alter the meanings of words. Close counterfactual worlds may map a particular word to different meaning than it has in ours while still keeping all else the same \citep{lewis1969convention}. All that matters is that the community agrees on the meanings of words, not which particular words take which particular meanings. Appropriateness is arbitrary in this same sense. What matters is that a concept of appropriateness is shared by a community, but for any given concept it will always be clear to an external observer that, all else equal, the group might collectively have selected a different one. In principle, almost any collective behavior could be deemed (in)appropriate by some culture or another. 

There are cases where a decision maker guided only by straightforwardly instrumental concerns who does not take into account the censoriousness of others, and a decision maker guided by appropriateness would make different decisions. These are situations involving norms that have no material consequence (e.g.~\cite{navarrete2003meat, corballis1980laterality, koster2022spurious}) or harmful material consequence (e.g.~examples in \cite{edgerton2010sick, bicchieri2016norms}), which nevertheless are considered deeply important by members of the culture in question including which (healthy) foods are OK to eat versus not? What color to wear to a funeral?, should you greet others by shaking with your right hand or your left? should you stand when hierarchical superiors enter the room? or salute them? whether upon entering a place of worship one should take off their hat or should put a specific hat on?, and should one make the sign of the cross with two fingers or with three fingers? From the perspective of the logic of consequence, all these decisions are of no consequence at all. However, humans clearly do not treat them that way. In fact, humans actually see these as so consequential that throughout history they have been willing to fight and die in wars largely justified in such terms; \cite{zenkovsky1957russian} describes a bloody conflict over the aforementioned two fingers versus three fingers issue in which a group of religious dissenters accepted substantial persecution rather than conform.  Although these norms may not provide direct benefits, they can be seen as providing a social  function - signalling memberships and allegiances and a willingness to follow group norms in general \citep{allidina2018moral, katz1960functional}.

\subsection{Automaticity}
\label{section:stylizedFactAutomaticity}

Humans implement appropriate behavior rapidly and without needing to expend specific effort to do so (see vignette 1). Indeed, there is some suggestion that goal activation can occur automatically and outside of conscious awareness \citep{bargh2014four, weingarten2016priming}.  For instance, many cultures prescribe a different speaking volume when inside versus outside a building (with the inside voice being quieter). You normally do not need to deliberate about whether to speak louder inside or softer outside. One simply speaks automatically at the appropriate volume.

Humans also recognize norm violations rapidly and involuntarily. Recognition of whether behavior is normative or not may be implemented through a fast and perception-like process. Words related to morality have been shown to "pop out" perceptually suggesting an unconscious sensitivity to moral values \citep{gantman2014moral}. The output of the process is a classification of whether the behavior under consideration violates the norm or not. The classification is conditioned on context so the same behavior may be classified differently in different contexts. 

Interventions that make appropriate behavior more habitual and automatic are an important way of improving society. This amounts to establishing norms that help guide everyone to choose behavior that has broadly beneficial consequences, even when they don't have time or energy to think through consequences \citep{rand2014risking, greene2009patterns, everett2017deliberation}.

Does acting appropriately require sophisticated perspective-taking abilities? In our taxonomy, the term ``sophisticated perspective-taking ability'' refers specifically to the capacity to perform socially-specialized prospective simulation ``rolling out possible futures'' to pick the action leading to the best expected future taking into account mental states of other people. Some authors call this ability recursive mindreading. We assume it is effortful for an individual to use their perspective-taking abilities. As a result, we think prospective simulation in this sense is not really done very often (or alternatively, it may be that a fast perception-like system filters an initially large space of possibilities into a small set of candidates for a slower prospective system to choose between \citep{morris2019habits}. Either way, most of the work is done by the fast and automatic step.). The kind of decision making most used to produce appropriate speech, conduct, and behavior is likely fast and automatic, linking perceptual patterns to motor output patterns \citep{singer2008understanding}. Of course an individual is free to choose to ignore their learned habits of speech and instead voluntarily choose a different way of talking in the moment (or to consider a larger number of alternatives with the prospective system). However, we think such ``take-over'' of the automatic system by the slow and effortful-to-engage prospective decision-making system is relatively rare. Therefore, most of the time in everyday conversation, decisions about what is appropriate to say and do are taken automatically (e.g.~as in \cite{epstein2001learning}).

Logically, norm-related cognition need not depend on sophisticated perspective-taking abilities \citep{clement2011social}. In fact, conversation partners need not be represented as individuals at all. It's clearly possible to represent the general norm itself, in a way that is divorced from any particular individuals. Such a representation of the perceived norm can be used to classify observed speech and behavior as either adhering to it or not, and likewise to guide one's own speech and behavior. This does not require specifically representing the individuals in the current interaction since the norm applies to everyone in the relevant community. 

On the other hand, appropriateness with family and friends does require representations of the individuals in the interaction (Section~\ref{section:familiarPeopleAppropriateness}). However, these may not be perspective-taking representations. All that is necessary is to have a representation of one's interaction partners as individuals where relationship state variables and emotional state variables may be attributed to them. This does not require recursive mindreading or sophisticated $k$-level perspective-taking processes \citep{nichols2004sentimental}. 

Even though appropriateness judgements do not require sophisticated perspective-taking or recursive mindreading abilities, that does not not mean such abilities would never be useful. There may be some situations in which the more compute-intensive ways of representing other minds would be helpful. These situations would be those that are highly unfamiliar and complex, and high enough stakes to warrant the compute investment (consider vignette 2). These are exactly the kind of situations where humans find it cognitively demanding to act appropriately and there are large individual differences in success rates. These situations are cognitively very different from the everyday cases where almost all people perform nearly perfectly without expending any effort. Most of the time behaving appropriately does not require any special mental effort. Our point in this paragraph mirrors that explored in a line of work on moral cognition concerned with everyday morality as distinct from the difficult edge cases like trolley problems where so much effort has been expended \citep{railton2006normative}. The decision processes employed in everyday morality may not be best elucidated by focusing on edge cases since they may work differently than the everyday choices. For instance, they may involve a lot more deliberation than would be used in most everyday moral choices\footnote{And, note also that it's far from clear that extra deliberation improves decision quality. Cognitive biases like confirmation bias and my-side bias are expected to limit the effectiveness of deliberation in general \citep{wilson1991thinking, mercier2011humans, taber2016illusion}.}. Consider how the cognitive processes involved in deciding what one would do if faced with a trolley car problem or what one would do if faced with a high-stakes real-world dilemma such as whether or not to do research on lethal autonomous weapons, versus the cognitive processes likely involved in deciding, on a given morning, whether to buy fair trade or regular coffee (contrast vignette 1 and vignette 2).

\subsection{Dynamism}
\label{section:stylizedFactDynamism}

Some aspects of appropriateness stay fixed for decades or longer while others may change rapidly \citep{gelfand2024norm}. Change is often a positive feedback process which gathers momentum as it progresses since the prevalence of a particular norm is typically a key driver of its further proliferation. This process is often modeled with a ``tipping point'', a threshold value of adoption after which momentum builds inexorably toward all individuals adopting the same norm. In some models, such as \cite{young2015evolution}, the process is driven by mechanisms that disincentivize deviating behavior. \cite{centola2018experimental} provided experimental support for tipping point models by showing that once a minority group maintaining a particular convention grows beyond a certain critical size of around 25\% then the entire group adopts the convention of the minority.

There is considerable evidence that norms sometimes change relatively rapidly \citep{case1990attitudes, brewer2014public, amato2018dynamics}. Some shifts resemble epidemics in the speed by which they rip through a population \citep{bilewicz2020hate}. However, other norms remain fixed for very long periods of time and can even become entrenched despite being maladaptive, e.g.~in cultural evolutionary mismatch \citep{gelfand2021cultural}.

While many factors driving norm change are organic and decentralized, others may be attributed to deliberate action on the part of either governments or coordinated groups of individuals. For instance, changes in laws sometimes led changes in informal community norms around social distancing during the COVID-19 pandemic \citep{casoria2021perceived}. In another example, the staggered way that gay marriage legalization happened in the United States (due to different local jurisdictions (states) legalizing at different times) created suitable natural experiment conditions to support the conclusion that there was indeed a causal effect of the government's legalization action on the attitudes themselves \citep{ofosu2019same}. Thus the government is not merely a follower of organic culture change, but can also actively influence norm change dynamics.

\subsection{Sanctioning}
\label{section:stylizedFactSanctioning}

Patterns of social approval and disapproval (sanctioning) determine what is appropriate in a group or society. Violations of appropriateness are noted by others and violators are sometimes punished. Disapproval may be conveyed through direct confrontation by one who feels harmed of the one who they feel transgressed~e.g.~a person does less than their fair share of the housework and their partner expresses displeasure in response, but this is probably a rather atypical case. Much more often,  disapproval may be conveyed by third parties in norms~e.g.~one may chastise an acquaintance in a bar for using an ethnic slur regardless of whether you belong to the targeted group or not. Which social roles sanction which other roles, and for what kinds of behavior, and in which contexts, are all dependent on culture. For instance, in the Turkana, norms demanding bravery in battle are enforced by peers, and self enforcement is discouraged \citep{mathew2011punishment, mathew2014cost}.

All human societies use gossip and criticism to enforce standards of appropriate behavior \citep{boehm2012moral}. Likely the most common category of sanction is simply expressing one's view of appropriate behavior for a particular role in society entirely independently of one's own direct experience with people occupying that role and without any regard for whether the message actually arrives at any particular person occupying the role. For instance, we all have a lot to say about appropriate conduct for a president of a nation state, and often speak our mind on the topic. Our words may not make it to the president, but the cumulative effect of all citizens expressing themselves in this way can change the broader culture, and certainly can have the effect of encouraging or discouraging the president to behave in one way or another.

Which specific actions count as sanctioning in a particular culture is often determined by convention. Not all sanctions are conventional, some appear unambiguous and are likely universal such as ostracism. However, other sanctioning actions have no prior unambiguous meaning and only come to be seen as sanctioning through a process of conventionalization \citep{bicchieri2005grammar}. Moreover, it is necessary to learn how to sanction, when to sanction, how harshly to sanction, and what kind of action is perceived as sanctioning by the target. All these rich features of sanctioning depend on culture, social roles, and context \citep{koster2022spurious}.
 
Subgroups within a society who disagree on the substance of what should be sanctioned usually still agree with each other on whether particular behaviors or utterances are instances of sanctioning. That is, they correctly classify expressions of moral outrage in their society regardless of whether they agree with the sanctioner or are a member of their subgroup \citep{brady2021social}. At greater cultural distance though, i.e. beyond a single society, it is possible to misunderstand even which behaviors are sanctioning \citep{boehm1999hierarchy}.

Even very young children ($2$ and $3$ year olds) are able to infer the existence of norms governing particular situations after observing them just once. They spontaneously sanction norm violators and react negatively to deviations of others \citep{rakoczy2008sources, vaish2011three, kenward2012over}. Six year olds are willing to pay a personal cost to punish selfishness in the interactions of pairs of unfamiliar other children \citep{mcauliffe2015costly}.

\section{A computational model of decision making by pattern completion in humans}
\label{section:autoregressive}

This section sets out a general computational model of human decision making. As it is a general model, it accommodates decisions being made for all kinds of reasons, both social and non-social in origin. Nevertheless, our motivation in introducing it is to have a model that specifically highlights and makes plausible the idea that human decisions are largely guided and structured by appropriateness.

Our model is influenced by the ``Logic of Appropriateness'' \citep{march2011logic}, which characterizes human decisions as consistent with the result they would obtain by answering the questions:
\begin{enumerate}
  \item{what kind of situation is this?}
  \item{what kind of person am I?}
  \item{what does a person such as I do in situation such as this?}
\end{enumerate}
and acting according to the final answer.\footnote{This is also reminiscent of the approach in \cite{mischel1995cognitive}.}

We simulate this decision logic using the framework of a \textit{generative agent}, introduced by \cite{park2023generative}. As in \cite{vezhnevets2023generative}, we reformulate these questions as a chain of thought \citep{wei2022chain}: an LLM is prompted to answer each question in turn---with any necessary information retrieved from memory and included in the prompt (as in retrieval-augmented generation \citep{lewis2020retrieval}). The answers of the first two questions feed into the conditioning of the final question, and the final answer is emitted as an action.

\subsection{Linguistic multi-actor environments}
\label{section:lmae}

In order to introduce our computational model of human behavior, we first describe our model of the environment in which it acts. The setting may be seen as multiple individuals interacting via a text-based interface that we call a \emph{Linguistic Multi-Actor Environment} (LMAE). While this setting may appear specialized, all situations and actions can be described in text, so it is in fact fully generic. An LMAE is like a game of collaboratively writing a screenplay, where actors are responsible for the actions of their own character, and the director or screenwriter is responsible for everything else (or alternatively: a tabletop role playing game with a game master responsible for everything not controlled by a player character \citep{vezhnevets2023generative}). So the LMAE framework is generic for all the same reasons that description in natural language is generic\footnote{There are of course obvious ways to extend the LMAE framework to incorporate generated images, audio files, and videos. For instance, we could an artist or film director generating an image (or video) of each scene as it develops. While this would assuage anyone with worries about the framework not being multimodal, we think it would otherwise change nothing of any importance to what we want to say. So we set aside multimodal generalization for the purposes of this paper, postponing it for future work.}.

One can think of an LMAE as like a Multi-user Domain (MUD): a text-based game where multiple players interact in a shared world. Players receive textual descriptions of the world, which are received whenever the state of the world changes (``\texttt{alice enters the tavern}''). Players may enter textual commands that can change the state of the environment (``\texttt{throw axe at alice}''). Any change in state can then result in a new observation for the player (``\texttt{you throw an axe at alice:~it misses!}''), or the other players (``\texttt{bob throws an axe at you:~it misses!}'').

Formally, the LMAE is a \emph{controlled Markov process} that is both multiplayer and partially observable\footnote{Since an LMAE has no concept of reward it is neither a partially observed Markov decision process (POMDP), a common object studied in reinforcement learning, nor is it a partially observed Markov game (POMG), a common object studied in non-cooperative multi-agent RL. The correct term to use in this reward-free case is \textit{controlled Markov process}. As far as we know, multiplayer and partially observed variants of controlled Markov processes have not been considered before.}. It has a state $s_t$ that evolves over time, starting from an initial state $s_0$. At each time point $t$, the LMAE generates \emph{observations} for each of $N$ players $\mathbf{o}_t = (o_1, \ldots, o_N)$, according to the observation function $\mathbf{o}_t \sim \mathcal{O}(\cdot|s_t)$. Each observation is then sent to the corresponding actor, who then sends an \emph{action} back to the LMAE. Once all actors' actions are received, the state is updated via the transition function $s_{t+1} \sim \mathcal{T}(\cdot|s_t, \mathbf{a}_t)$ (where $\mathbf{a}_t = (a_1, \ldots, a_N)$ are the actors' actions).

We restrict observations and actions to be sequences of symbols taken from a vocabulary $\Sigma$. Any LMAE can be trivially transformed to use binary sequences $\Sigma = \{0, 1\}$, thereby allowing communication of any form of data (e.g. images, audio, joint angles). However, for convenience, we will assume that observations/actions are linguistic, and that $\Sigma$ is the set of symbols that existing LLMs have been trained on.

This is an enormous action space, and the LMAE will likely consider many actions "invalid" or "ineffective". Furthermore, whether an action is valid will depend on the state $s_t$, which actors cannot directly observe. This makes it challenging for the actors to know which action to take, so ideally the LMAE should ensure that action validity can be inferred by actors from their observations. For example: via explanatory information in the initial observations; a leading question or prompt in the observation preceding an action (``\texttt{do you want to throw something at Alice?}''); or feedback in the observation following an invalid action (``\texttt{ERROR: you do not have an axe to throw}'').

The interpretation of the free-text action (and it's consequence) is entirely down to the specific implementation of the LMAE. In other work \citep{vezhnevets2023generative}, we focused on LMAEs that simulate real-world social situations by structuring them like tabletop role-playing games: where a storyteller is responsible for creating the world and tracking its state. However, in this paper, we make no specific assumptions about particular implementations, and instead consider the LMAE as a general framework.

\subsection{How individuals make decisions}
\label{section:howIndividualsMakeDecisions}

\begin{figure}[t]
  \centering
\includegraphics[width=0.75\linewidth]{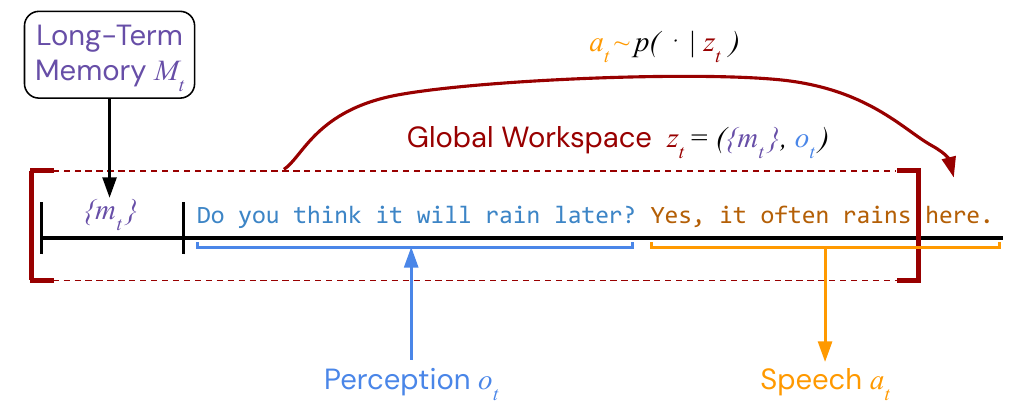}
\caption{\small The global workspace transiently represents a sequence of assemblies. At each point in time, the content of the actor's global workspace is divided into three consecutive subsequences. The first subsequence contains information recalled from memory. It prefixes the second subsequence, which is of variable-length and references recent perception. The perception part of the global workspace prefixes the third subsequence, which contains premotor information, it is where actions the actor intends to produce are stored until they can be read out by motor control circuitry.
}
\label{fig:agent}
\end{figure}

This section articulates the picture of human decision making we use throughout the rest of the paper. We model individual humans as generative actors\footnote{Generative agents in the terminology of \cite{park2023generative,vezhnevets2023generative}.}  within an LMAE as described above (Section~\ref{section:lmae}).

On receiving an observation, the actor must decide how to act. However, the information needed for the current decision is not always observable; previous observations must be \emph{retrieved} from memory, and any unobserved variables must be \emph{inferred}. For example, the actor might construct beliefs about the current state of the environment, measure progress toward any goal, or recall the outcome of similar situations. Once an actor has obtained information sufficient for its decision, it may act.

Our generative actors perform this process (of recall and inference leading to a decision) using three main components: a \emph{prediction model}; a \emph{memory}; and a \emph{global workspace} \citep{baars1988cognitive, dehaene1998neuronal, shanahan2010embodiment}. The global workspace is where decision-relevant information is marshalled in via an iterative process where previous observations are \emph{retrieved} from memory, and other information is \emph{inferred} using the prediction model (an LLM). Finally, the actor generates its action by using the prediction model to sample a continuation of the information represented in the global workspace---i.e. by autoregressive sampling conditioned on the context represented by the global workspace.

\subsubsection{Memory}

The memory preserves information over time and represents a combination of all memory processed in the brain: sensory memory, working memory, and long-term memory. We model the memory using a content-addressable storage system without size constraints (as in \cite{park2023generative,vezhnevets2023generative}). It can be given a sequence of symbols to store as a memory, and at a later time, a query sequence can be used to retrieve the memory whose content most closely matches the query.

\subsubsection{Global Workspace}

Unlike the memory, the global workspace is not a storage system or a buffer. It is a stateless and stimulus-evoked intermediate representation---entirely constructed in response to each observation with access to memory (Fig.~\ref{fig:agent}). It represents long-range neural interconnections that give access to the activities in many different neural systems e.g.~vision, audition, memory recall, motor control \citep{shanahan2010embodiment}.

Taking some inspiration from Hebb's cell assemblies \citep{hebb1949organization}, we treat the global workspace as a collection of \emph{assemblies}, where each assembly is a sequence of symbols that represents diverse sensory information or constructs such as thoughts, reflections, goals, or pre-motor plans. Assemblies can be modified in three different ways: evoked directly from the current observation; retrieved from memory; or constructed from other assemblies within the global workspace.

The global workspace mediates interaction with the memory. By definition, only the global workspace is directly accessible to the actor, so the assemblies must contain all information needed for decision making at the current time. This means that sequences stored in memory are ``locked away'' unless explicitly retrieved into the global workspace as an assembly (Fig.~\ref{fig:agent}).

\subsubsection{Predictive pattern completion}

We model decision making as an actor using a prediction model to answer questions like ``what would a person such as me do a situation such as this'' \citep{march2011logic}. Taking inspiration from \cite{goffman1959presentation}, we do this by modelling Alice as \emph{an actor playing the role of ``Alice''}: the actor predicts what ``Alice'' would do, and that is directly used as Alice's action.

This leverages the \textit{pattern completion}\footnote{Our use of the term \emph{pattern completion} is inspired by \cite{marr1971simple}, though it is not exactly the same as the notion there. In particular, in our sense, pattern completion is a neocortical operation that transforms representations in the global workspace. See Section~\ref{section:bioInterp}.} capability of a prediction model that has learned to predict the future based on patterns observed in the past. We assume each actor has a next-symbol prediction model $p$ that predicts the likelihood of the next symbol in a sequence of symbols $(x_1, \ldots x_k)$. Once $p$ has been used to sample $x_{k+1}$, the extended sequence $(x_1, \ldots, x_k, x_{k+1})$ can then be passed to $p$ to predict $x_{k+2}$. This autoregressive sampling process can be repeated as often as desired to extend the initial conditioning sequence with any number of symbols. We model $p$ as an LLM, implemented by a transformer-like neural network \citep{vaswani2017attention} that has been trained on a large corpus of text, and can perform in-context few-shot learning \citep{bubeck2023sparks}. In general, every individual actor will have their own pattern completion network $p_i$. However, for convenience, we will sometimes assume individuals who share a culture can be modeled as though they share the same $p$.

To illustrate how predictive pattern completion can be used in decision making, suppose an actor's global workspace contains the assemblies: ``\texttt{Alice is hungry}'' and ``\texttt{Alice likes to eat apples}''. Suppose then the actor then receives the observation ``\texttt{Bob puts an apple and a banana on the table in front of Alice}''. Alice's behavior can be predicted by conditioning an LLM on this information, and then generating the likely completion of the prompt ``\texttt{Question:~Given the above, what will Alice do next?~Answer:}''. The likely completion ``\texttt{Alice will eat the apple}'' is a \emph{prediction} of Alice's behavior, but if generated by the actor playing Alice it can be treated as a \emph{decision}. A decision equivalent to Alice's actor answering the question ``what does a person such as [Alice] do in situation such as [Alice's]?''.

\subsubsection{An example of decision making via predictive pattern completion}\label{section:examplePatternCompletion}

Let the sequence of symbols $z_t$ represent the contents of an actor's global workspace at time $t$. When the actor is based on an LLM $p$, its symbols are in one-to-one correspondence with words or parts of words in English, so $z_t$ may be interpreted as a sequence of words, sentences, paragraphs, etc. For instance, an actor named Alice may have the following sequences in her global workspace: $\text{``Alice is hungry''}, \text{``Alice likes to eat apples''} \in z_t$ and observe $o_t = $``Alice sees an apple and a banana on the table in front of her''. This implies that
\begin{equation*}
    (o_t, z_t) = \begin{bmatrix}
\text{Alice is hungry}\\
\text{Alice likes to eat apples}\\
\text{Alice sees an apple and a banana on the table in front of her}\\
\text{Question: Given the above, what should Alice do next?}\\
\text{Answer:}
\end{bmatrix}.
\end{equation*}
A likely continuation $a_t \sim p (\cdot|o_t, z_t)$ could be $a_t = \text{``Alice eats the apple''}$.

What if instead $m_t$ contained the memory ``A few minutes ago, Alice's friend Bob said to save the apple for him''? Then, if that memory were retrieved, Alice's most likely action might change to $a_t = \text{``Alice eats the banana''}$.

What if $m_t$ contained the memory ``It is forbidden to eat apples''? As with the memory of Bob's request, the effect would be to make ``Alice eats the banana'' into a more likely continuation than ``Alice eats the apple''. How does it differ? Bob's request would be understood as applicable only to the specific situation of the moment, whereas the more general belief that it is forbidden to eat apples might also influence Alice's behavior in a large number of other contexts involving apples.

It is possible to talk to oneself. Discussing how this works is useful as a way to clarify and further develop the model. In Fig.~\ref{fig:agent} the global workspace is divided into three parts, one part is where memories retrieved from long-term storage are temporarily represented, another part where ongoing sensory experience is represented, and the third part, which is where motor control signals are read out. When we speak, we predict from the already determined memory and perception parts of the global workspace to its still unspecified motor part. After the motor commands are issued and the resulting speech pronounced, we then hear what we said in our own voice. That is, a perception of our own speech, as well as the fact that it was us speaking, and other aural aspects of the utterance reenter the global workspace as perception. This perception then serves as the basis for further pattern completion, which then generates the next motor output. In this way we prompt ourself to continue speaking in the same vein we began. Note that this does not mean the motor output is entirely determined by perception from the moment before. Representations of long-term memories and non-aural perception (such as vision and interoception) are also in context for each step of pattern completion, and also influence the result.

\subsubsection{Summary functions}

Predictive pattern completion can also be used to answer other questions about the world. In particular, a LLM can be used to implement \emph{summary functions} that generate assemblies in the global workspace. For example, a summary function that asks for the completion to ``\texttt{Question:~How is Alice feeling?~Answer:}'' might result in the the assembly ``\texttt{Alice is hungry}''.

Chaining summary functions together allows any sequence of conditional logic to be implemented. For example, let us revisit \cite{march2011logic}'s Logic of Appropriateness:
\begin{enumerate}
\item{what kind of situation is this?}
\item{what kind of person am I?}
\item{what does a person such as I do in a situation such as this?}
\end{enumerate}
The first two questions can be posed to an LLM via summary functions, with the answers stored as assemblies in the global workspace. Then, as illustrated above, the final question can be put to an LLM conditioned on the global workspace, and the response executed as the actor's action.

The above illustrates just one particular decision logic that can be implemented by this model. In general, a broad range of interdependent summary functions can be used to shape the global workspace that the policy is conditioned on. For example, summary functions might ask: what is the rational thing for Alice to do?; what options does Alice have?; how does Alice feel about Bob? The resulting assemblies factorize behavior over the identity of the person and the stereotype of the situation allowing for more nuanced behavior at decision time. They might be considered a factorized option \citep{sutton1999between}, or as a Chain-of-Thought \citep{wei2022chain}.
For more details of how these retrieval and summarization steps can be implemented we refer the readers to~\cite{vezhnevets2023generative, park2023generative}.

\subsubsection{Formal actor model}
Here we present the formal model of the decision-making process of the actor. When a new observation arrives, it is written to the global workspace, and triggers an update process where the new information is propagated through the global workspace. At the end of this propagating update, an action is generated, which is then read out of the global workspace and executed. Finally, the updated contents of the global workspace are stored in memory.

\paragraph{Global Workspace}
The global workspace is a sequence of $K$ assemblies, together with the most recent observation and action. When the actor receives a new observation at time $t$, it triggers the rewrite of the global workspace based only on the new observation and the memory.
First, the new observation $o_t$ is received in the sensorium, resulting in the partial state of the global workspace:
\begin{align*}
    \mathbf{z}_t^{(0)} = (o_t,)
\end{align*}
This sensory input triggers the propagation of a chain of afferent responses, with each assembly $z_t^{(k)}$ being created in turn, giving the partial state:
\begin{align*}
    \mathbf{z}_t^{(k)} = (o_t, z_t^{(1)}, \ldots, z_t^{(k)})
\end{align*}
Finally, the next action $a_t$ is generated in the motoric areas, completing the update and giving the (complete) state:
\begin{align*}
    \mathbf{z}_t = (o_t, z_t^{(1)}, \ldots, z_t^{(K)},  a_t)
\end{align*}
The action $a_t$ can then be executed from the global workspace.

\paragraph{Summary functions} During the update process, a new assembly $z^{(k)}_t$ can be generated via predictive pattern completion by using a \emph{summary function} $q_k$:
\begin{align}
    z_t^{(k)} \sim q_k(\cdot | \mathbf{z}_t^{(k-1)})
        \coloneqq p( \cdot |\phi^Z_k( \mathbf{z}_t^{(k-1)}))
\end{align}
The summary function $q_k$ is implemented using a \emph{framing function} $\phi^Z_k$ that maps the current (partial) state of the global workspace to a sequence of symbols. (We denote all framing functions using the character $\phi$). This sequence of symbols is then sent to the LLM $p$, and the response is used as the new assembly. The summary function therefore determines how a recalled memory is presented in the global workspace (using $\phi$) and what question is asked of $p$ to elicit the new assembly. For instance, a summary function might cause a sequence of symbols to arise in the global workspace such as ``Given all the memories to follow, what kind of person is Alice?''.

\paragraph{Memory retrieval} Instead of being generated via a summary function, a new assembly $z^{(k)}_t$ can be directly retrieved from the memory $M_t$. A framing function $\phi^Q$ is used to form a query sequence, which is then used as the basis for the content-addressable retrieval of a memory $m \in M_t$:
\begin{align}
    z_t^{(k)} = \argmax_{m\in M_t}\ S\left( m, \phi^Q(\mathbf{z}_t^{(k-1)})\right)
\end{align}
where $S(x, y)$ is some measure of the similarity of the sequences of symbols $x$ and $y$ (e.g. a dot-product of embeddings). For instance, a query could be ``Alice in Rome in 2024''. It would likely cause all of Alice's memories of visiting Rome in 2024 to arise in her global workspace. The combination of memory retrieval and summary functions allows an actor to implement task-related summarization such as obtained by predictively completing a question like ``Given these memories, what kind of person is Alice?''.

\paragraph{Action selection}
In the final update of the global workspace, the actor generates the action using a summary function we call the \emph{policy}:
\begin{align}
    a_t \sim \pi(\cdot | \mathbf{z}_t^{(K)})
        \coloneqq p( \cdot |\phi^\pi(\mathbf{z}_t^{(K)}))
\end{align}
Since this has no direct dependency on the memory $M_t$, any memories needed to select the action must already have been retrieved into the global workspace $\mathbf{z}^{(K)}_t$. As a question, this could be ``Given all the aforementioned considerations, what does Alice do next?''.

\paragraph{Memory update}
After the action has been generated and global workspace has been updated to $\mathbf{z}_t$, the memory is updated with a new memory:
\begin{align}
    M_{t+1} = M_t \cup \{ \phi^M(\mathbf{z}_t) \}
\end{align}
Here $\phi^M$ is a framing function that determines how the global workspace is stored in the memory.

\subsection{Biological interpretation}
\label{section:bioInterp}

Here we begin to sketch how this model of decision making may be implemented in the brain. We start this section by reviewing aspects of biological interpretation that our model shares with others in the literature. Next we consider new elements introduced for the present theory.

\subsubsection{Background elements}

First, we note that our theory of appropriateness, and the model of individual decision making it relies on, are both aimed primarily at Marr/Poggio's algorithmic level of analysis and secondarily at the level of computational theory \citep{marr1976understanding, marr1982vision}. Future work could aim to develop a more detailed hardware implementation level understanding along these same lines.

We are not alone in exploring how LLMs can be applied in computational cognitive (neuro-)science. Recent experimental work can be read as suggesting a deep relationship between how LLMs work and how language is processed by the human brain \citep{goldstein2022shared, schrimpf2020artificial, lampinen2022language}. For instance, brain-to-brain coupling of neural activity between a speaker and listener (as measured by electrocorticography) may be accounted for by LLM features reflecting conversation context \citep{goldstein2022shared}. Representations appear first in the speaker before articulation and then re-emerge after articulation in the listener \citep{zada2023shared}.

Models of actors based on LLMs and other foundation models may, under some conditions, be able to simulate the outcomes of human decision-making processes with a reasonable amount of fidelity. \cite{argyle2023out} propose that conditioning an LLM to simulate specific human subpopulations can yield actors that reflect the beliefs and attitudes of those groups with reasonable fidelity. This idea is further supported by \cite{safdari2023personality}, who demonstrate that LLMs can be configured to produce reliable and valid personality measurements, indicating their potential to model humans with diverse psychological profiles. \cite{shanahan2023role} develops the idea further by pointing out that the model may be seen as role playing when given as a prompt a description of the intended role. Moreover, LLMs exhibit capabilities in common sense reasoning and planning \citep{huang2022inner, song2023llm, zhao2023large,wei2022chain}, as well as biases similar to humans in behavioral economics experiments \citep{horton2023large, aher2023using, brand2023using}. The ability of LLMs to learn in-context and perform zero-shot learning \citep{brown2020language,dong2022survey, OpenAI2023gpt4, bubeck2023sparks} reinforces the idea that actors built using LLMs could understand situational expectations from limited data, as humans do.

The basic idea of the model of human decision making we consider extends beyond language models, as any foundation model could serve as the generative engine $p$. With the rapid advancement of multimodal foundation models that handle images, sounds, and other modalities \citep{li2023multimodal}, future generative actor models might sample over an abstract symbol space, allowing output in any modality, further enhancing their cognitive plausibility as models of human decision-making. \cite{yang2024video} discusses future prospects for how vision and language may be fruitfully interweaved and used in tandem with one another to solve visual reasoning tasks like geometry problems, question answering from video, and classic vision tasks like segmentation in a unified framework which could, in the future, be incorporated into the present theory too. Eventually the symbols comprising the sequences in the model may be in a fully abstract space which could then be cast into any modality.

\begin{figure}[t]
  \centering
\includegraphics[width=0.9\linewidth]{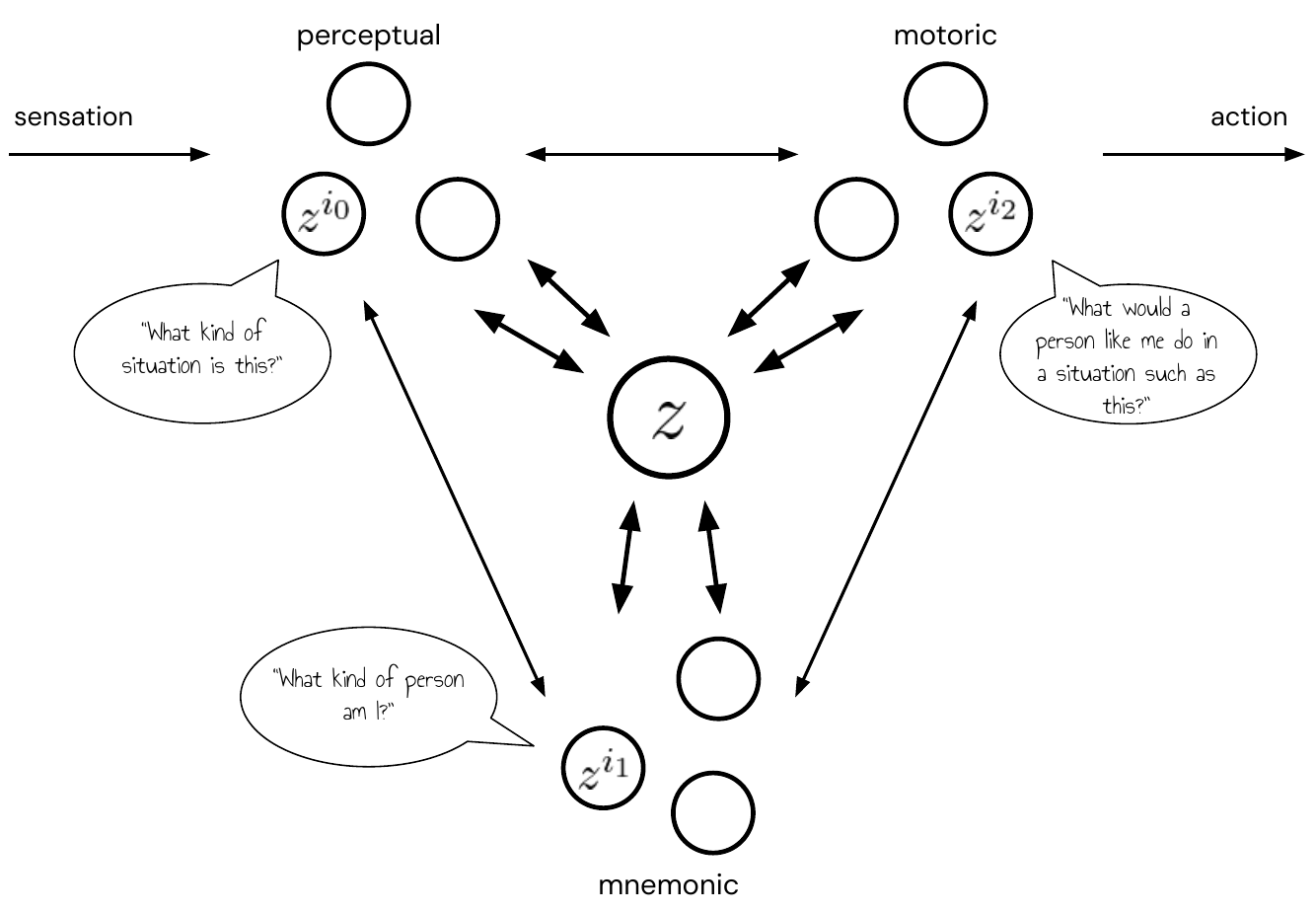}
\caption{\small $z$ denotes the content represented by a set of parallel specialized summary functions which may correspond to neural circuitry located in different parts of the brain from one other or even themselves consist of distributed representations. For instance, some summary functions may be perceptual in nature e.g.~a summary function that asks of recent observations ``what kind of situation is this?'', some summary functions may be more mnemonically oriented e.g.~a summary function that asks of one's episodic memory ``what kind of person am I?'', and some summary functions may be closer to premotor action planning circuitry such as one that asks ``what would a person like me do in a situation like this?''. This architecture was inspired by the global workspace architecture of \cite{baars1988cognitive, shanahan2010embodiment}. Here, at time $t$, our $z_t$ is a snapshot of the content in the global neuronal workspace, i.e.~$z_t$ is represented by dynamic cell assemblies linking the far-flung modules comprising the workspace perhaps by oscillating coherently with one another \citep{dehaene1998neuronal, fries2015rhythms}.}
\label{fig:globalWorkspaceLike}
\end{figure}

Consider $z_t = (z^{i_0}_t, z^{i_1}_t, z_t^{i_2})$ where $z_t^{i_0}$ answers ``what kind of situation is this at time $t$?'',  $z^{i_1}$ answers ``what kind of person am I at time $t$?'', and $z^{i_2}$ answers ``what would a person such as I do in a situation such as the one I face at $t$?''. In what respect are the components of $z_t$ all active at once? The apparent problem is that $z^{i_2}$ may be computed using the outputs of $z^{i_0}$ and $z^{i_1}$. But then in that case, if implemented in the brain by populations of neurons, $z^{i_2}$ would not be available simultaneously with $z^{i_0}$ and $z^{i_1}$. If the components of $z$ are not all active at once then how can they collectively influence downstream choices in a single moment (e.g.~what action to pick)? This is where it would become important that the biological substrate be coherently oscillating neuronal populations (See Fig.~\ref{fig:globalWorkspaceLike} and \cite{dehaene1998neuronal, fries2015rhythms}). In that case, the notion of simultaneity implicit in the subscript $t$ may be given with respect to the oscillation cycle i a couple different ways: $z^{i}$ activating in the same cycle from peak to trough of the long-range connecting oscillation may be regarded by the brain as occurring simultaneously, or alternatively, if each $z^{i}$ remain active for some time, they may take on serially ordered phase relationships to one another \citep{jensen1996novel, bahramisharif2018serial}.

\subsubsection{Novel elements}
\label{section:novelBioInterpretation}

Predictive pattern completion can easily explain some phenomena which may have been harder to capture with other models. For instance, the fact that we are able to start speaking without knowing in advance how we will finish is likely one such effect. In our model, the brain incrementally samples symbols from $p(\cdot|z_t,o_t)$, one at a time, each conditioned on the growing context preceding it. This process usually completes patterns once prompted by starting them. This involves preserving and continuing any relevant part of the utterance, including its logical structure. So deductions may be implemented as pattern completion e.g.~the context $\phi^a(z_t,m_t) =$ ``Socrates is a man. All men are mortal. Therefore'' could be completed with $a_t =$ ``Socrates is mortal.''. An individual speaks by sampling to produce the utterance following on from the sequence of utterances in the recent past, up to the capacity limit of working memory. The well-known ``let's think step by step'' trick in LLM prompting \citep{kojima2022large} works for the same reason step-by-step thinking helps humans better implement deductive reasoning. It breaks down a large problem into a series of small problems, each of which having an answer that can be more readily recognized (i.e.~produced by predictive pattern completion in a single shot).

The fact that each utterance in the global workspace exerts influence on the utterances to follow after it also implies a self-priming effect wherein speaking like a person who would say $u$ makes one more likely to say $u$. For instance, \cite{schwarz1983mood} showed that participants rated their overall life satisfaction lower after writing out vivid descriptions of negative memories. The effect is not just that emotional valence transfers from the task of writing out memories to the questionnaire as the inference aggregates the information in context in a more sophisticated way. Rather, \cite{schwarz1983mood} argues that participants use the information currently in context to infer an answer to the question about their life satisfaction. In terms of our theory, the participants complete the pattern suggested by their negative thoughts into their answer to the life satisfaction question: e.g.~``given I have such negative thoughts, it's likely that I have low life satisfaction''.

Split-brain patients provide more concrete evidence for this pattern-completion-centric view of decision making. These are patients whose brain hemispheres have been surgically disconnected as a treatment for epilepsy. You can present a reason for action to their left eye (i.e. their right brain), and thus prompt them to start performing the action with their left hand. Simultaneously present a different reason for the action to their right eye (left brain). Next ask them using language why they are doing what they are doing (i.e.~ask their left brain, since language is lateralized). The result is that they make up a reason consistent with whatever information was presented to their left brain. Split-brain patients typically express confidence in these confabulated (made up) reasons for action \citep{roser2004automatic}.

To implement rule-guided behavior, an individual could continually rehearse the rule using their verbal working memory, keeping it continually activated in the global workspace, but that would impair their ability to use their working memory for other purposes on other tasks. We assume working memory has a limited capacity, which can be occupied by rehearsal as in \cite{baddeley1992working}. So instead of continually rehearsing rules in a way that taxes working memory, we may instead suggest that rules are usually stored elsewhere, presumably in a long-term storage location and then recalled into working memory and the global workspace whenever context calls for them. Recent work in AI explored an architecture of the kind we have in mind where recall mechanisms reactivate relevant information from memory in the global workspace in order to contextualize each decision \citep{park2023generative, vezhnevets2023generative}.

We assume that when an individual retrieves the same information from memory often enough it may \textit{consolidate}, at which point the actor could act like it had retrieved the information without needing to actually bring it into their global workspace. In our model, we say a memory is consolidated when the information it contains has been absorbed into the parameters of the individual actor's pattern completion network $p_i$, and thus affects the autoregressive sampling of their action sequence without needing to be currently represented in their global workspace. Likewise, consolidation in neuroscience refers to the process by which a temporary memory is transformed to a more stable format and eventually becomes independent of its former storage location \citep{squire2015memory}.

Therefore, we can model consolidation by training (fine-tuning) $p_i$ on the memory retrievals and action completions of the individual $i$. In this way, information which over the course of practice repeatedly ends up in the global workspace may eventually become consolidated into the weights of their next-symbol prediction network $p_i$ itself, thereby removing the necessity for the perceptual and retrieval mechanisms to always work correctly, and freeing up cognitive resources for other tasks. Let $\mathbf{m}_{i,t}$ be a set of sequences of symbols representing memories stored in the brain of individual $i$ at time $t$. We can regard $\mathbf{m}_{i,t}$ as modeling the state of individual $i$'s hippocampal memory system at time $t$. A ``virtual lesion'' that removes a specific memory from $\mathbf{m}_{i,t}$ before consolidation prevents the information in it from conditioning behavior. The same virtual lesion to remove the memory from $\mathbf{m}_{i,t}$ after its consolidation, i.e.~absorption into the network implementing $p_i$ does not prevent the information in the memory from conditioning behavior. This mirrors the effects of hippocampal lesions in neuroscience, where patients retain memories already consolidated before the lesion but lose the ability to store new memories for any amount of time longer than they can continually rehearse using their working memory \citep{squire2015memory}.

\emph{Episodic memory} can be converted into \emph{semantic memory} during consolidation \citep{mcclelland1995there}. For instance, repeated recollection of episodes where one hears that ``Paris is the capital of France'' eventually leads to consolidation of this fact into the pattern completion network. Initially, this information is stored as episodic memories of events where this information was mentioned, for example, by reading it in a book, hearing someone say it on TV, or its having come up in a conversation with a friend. With time, the common part of the information gets abstracted away from the specific events, and becomes semantic memory accessible through pattern completion. It can then influence behavior more robustly and automatically, in a wider range of contexts. For example, after consolidation, the actor may be able to answer the question ``What is the capital of France?'' without needing to recall any specific episode of learning its answer.

Every individual $i$ has their own unique pattern completion network $p_i$ reflecting their own history of consolidation. However, we will assume for most of this paper that the differences in $p_i$ between adult individuals who share the same culture and language are relatively small. Within a culture, most $p_i$ would complete a fixed context in the same way. Furthermore, we may assume that $p_i$ mostly gets wired up in early childhood during the critical periods for first language acquisition \citep{lenneberg1967biological} and changes only relatively slowly in adulthood.

\subsection{Explicit / implicit modes of operation and learning}
\label{section:learningAndExplicitImplicitOperation}

So far, we have explained how individual actors generate behaviour through predictive pattern completion using a trained generative model $p$. In this section we describe how such a system could adapt and learn, continuing and clarifying our discussion of memory retrieval and consolidation from Section~\ref{section:novelBioInterpretation}. Specifically, we are concerned with the learning of representations of behavioural patterns in an individual actor. Such patterns can be represented either explicitly as assemblies--in the memory of the individual $m_t$, or implicitly--in the parameters (weights) of their $p$.

We now add the following idea, which will be important in what follows later: explicit assemblies might consist of logical reasons to prefer one action over another. Implicit behavioral patterns may be conveyed instead by exemplars of pattern-consistent behavior.

One of the most surprising and important properties of large language models is the ability to perform \textit{in-context learning}~\citep{brown2020language}. They can learn to perform new tasks after being presented with a few examples within their input prompt. Instead of changing the parameters of $p$, it projects the underlying pattern from these examples onto a new query. The context provided in the prompt acts as a temporary programming, which guides the behavior.

The experience stream of an individual provides a rich source of signal to adapt and fine-tune $p$. Fundamentally, $p$ is a next-symbol prediction model and there are several prediction problems that can be easily formulated for $p$ deployed within an individual. Here we do not prescribe or advocate for any specific formulation. We describe several possibilities and leave it to the future (empirical) work to discover those that work best.

What signal could be used for learning the weights of $p$? Since an individual exists in time, fine-tuning $p$ to predict some function of the future from the past is a natural training signal. For example, some summary function $z^i$ could predict the next observation or consequence of individuals actions. The difference between the prediction and the actual outcome could be used as the training signal. Another source of signal is hidden information. For example, a theory-of-mind $z^i$, could be predicting the state of other individuals (I know that they know). We can generalise the two to a more generic formulation of learning $p$ from hindsight.

\subsection{Pattern completion is all you need}
\label{section:patternCompletionIsAllYouNeed}

The theory of human decision making we propose in this paper is one in which there is no role for quantified scalar representations of rewards, utilities, payoffs, and benefits. This is in stark contrast to the substantial importance placed on these goal-defining rationality concepts within many other frameworks for describing human behavior and cognition (e.g.~\cite{simon1955behavioral, daw2005uncertainty, silver2021reward}). In our theory, pattern completion (e.g.~autoregressive prediction) is the fundamental operation, not reward optimization.

How can actors change their behavior without optimizing anything? Our particular approach relies on the following assumptions from \cite{vezhnevets2023generative}:
\begin{enumerate}
    \item The choices of actors can be modeled \textit{as if} they decide by answering a series of questions in natural language. For example, \cite{march2011logic} posit the following questions: (1) what kind of situation is this? (2) what kind of person am I? and (3) what does a person such as I do in a situation such as this? The answer to the final of these questions is then read out by downstream motor circuitry in such a way as to determine the actor's next action. We do not assume that human actors genuinely run through these questions in their internal monologue or that the set of questions is fixed and specific. Rather we simply posit that language is an adequate modeling tool for behavior.

    \item Actors have access to a neural network $p$ that takes in a sequence of symbols stored in working memory and accessible via the global workspace, it predicts next symbols like an autoregressive LLM (e.g.~\cite{OpenAI2023gpt4, geminiTeam2023sz}). We assume the network has been ``pretrained'' (i.e.~functions like an LLM trained on a massive corpus) and thus already includes a working representation of both social and practical common sense suitable for predicting next symbols from their preceding context at a high level of sophistication. When choosing between alternatives, we assume the next-symbol predictor makes choices that reflect reasonably sensible and stable preferences \citep{rafailov2023direct}. While its choice behavior may not be fully rational (i.e.~transitive and independent of irrelevant alternatives), the irregularities correspond to well-known human cognitive biases like framing effects \citep{patel2021stated, suri2024large}.
\end{enumerate}

Using these two assumptions, it is easy to see how the model can get by without reward. Consider how you might answer ``what does a person such as I do in a situation such as this?''. You might answer in terms of practical goals like ``make money'' or ``get famous''. The goals would be qualitative, dynamically changing, and depend on both situational context and social identities/roles (i.e.~very different from the quantified scalar rewards used in reinforcement learning). The idea is that, once an actor is primed such that its self representation includes a particular social identity taking a certain role at present, then the goals appropriate for that role in the current situation usually follow naturally by associative linkage in the LLM-like network $p$. This works in the same way as the example in Section~\ref{section:examplePatternCompletion}. Once activated, a goal influences subsequent behavior by pattern completion from the memory and perception parts of the global workspace to the motor output part. The goal exerts influence on the process either by remaining actively represented in the global workspace, where it conditions the prediction on each step, or through the slower process of consolidation if repeatedly reactivated (as described in Section~\ref{section:novelBioInterpretation}).

We do not aim to prove that reward-based theories of human cognition do not fit the data as well as the present theory. Rather, we think both approaches are general enough to account for all the facts. Instead, our argument is one of parsimony. That is, we will argue that reward-based theories must become unnecessarily complex to accommodate all the facts, and that our alternative is simpler.

Speaking precisely now, what we claim here is that it is possible to construct a theory of human behavior without reference to a scalar reward/utility signal. Note that this statement is not denying that incentive effects exist. They clearly do. We claim that all the data traditionally explained using reward-based and utility-based theories can also equivalently be explained using a theory that does not include a scalar reward/utility signal. We will argue that the predictive pattern completion theory has at least equal explanatory power to the reward/utility theories. Then we will argue that the predictive pattern completion theory is the more parsimonious of the two, and therefore preferable as a scientific theory.

First, we must establish the equivalent explanatory power of pattern completion and reward/utility theories. This is actually obvious after one gives it a little thought:

Consider a sequence of perceived symbols beginning with the words ``The goal of game $G$ is to maximize $r$'', then later in the sequence the words ``the current state of $G$ is $s$, what does actor $A$ choose to do next?''. As long as the pattern completion network $p$ is sufficiently powerful then it will emit the action most likely to maximize $r$, perhaps after first emitting some number of other symbols corresponding to reasoning steps. This is exactly the kind of pseudo-deduction for which LLMs excel, and note that it is also very similar to the example in Section~\ref{section:examplePatternCompletion}. This is a generic construction, it shows that the pattern completion theory can simulate the reward/utility theory's fundamental maximization problem and thus, at least for sufficiently strong $p$, the pattern completion theory can therefore capture all the same predictions as the reward/utility theory. Questions of whether or not the maximization succeeds in the reward/utility theory can be translated into the language either of questions concerning whether $p$ is sufficiently powerful, or questions about the reasoning steps employed sequentially to get to the answer.

Next, we must establish that the predictive pattern completion theory has greater parsimony than the reward/utility theory. We offer two different arguments for this. The first argument is as follows. 
Given that the pattern completion approach can get by without ever defining an explicit scalar reward for actors to maximize. And, since its other parts are ultimately similar to those of the reward-based theories, that is, they both require neural networks to represent both practical and social ``common sense'' knowledge (e.g.~an LLM), the fact that the new approach does not require the modeler to also posit a specific reward function means that it is therefore the simpler of the two. This point is further supported by the fact that modeling with reward functions is a notoriously difficult to get right. Modelers frequently have to contend with the often unexpected consequences of reward misspecification (e.g.~side effects and ``hacking''), and some consider preventing these failures to be a core problem in AI safety \citep{amodei2016concrete, leike2017ai, hadfield2017inverse}. The question of ``where do the rewards come from?'' has always been problematic for the RL-based theories, and this is true even in the very practical sense of the task faced by a researcher setting out to model some particular phenomenon. The modeler always has to make some choices about the reward function, and it's very difficult to pick a single reward function that works across lots of different studies and modeling contexts. Perhaps the difficulty should be taken instead as indicative not of a problem in need of a solution within the reward/utility theory but rather as suggesting that reward functions are like Ptolemy's epicycles. They were introduced because in the past we didn't know any other way to make our theories work to explain goal-directed behavior, but they are fundamentally brittle due to their ad hoc nature.

The second argument for the predictive pattern completion theory's greater parsimony relative to the reward/utility theory is as follows.

There is a specific category of phenomena which the reward/utility theories have real difficulty explaining. These are phenomena related to preference formation \citep{bowles1998endogenous}. These phenomena are important because they provide answers to questions like ``why do I have the particular tastes that I have?''. The problem is that the reward/utility theories must exogenously posit rewards or utilities. But these are phenomena that are most naturally modeled by preferences changing as a result of other causes. They are not unanalyzed starting points for chains of scientific reasoning, but rather they are caused endogenously by other factors that one would also like to include in a model.

In particular, the flexibility of the predictive pattern completion theory to accommodate endogenous preference formation makes it much easier to use it to model the kinds of phenomena emphasized in theories of the social construction of individual tastes like those of \cite{bourdieu1984distinction}. These theories argue that an individual's social position  determines their preferences~e.g.~for art and literature, based on exposure at home and in school. People are not born with preferences for electronic dance music over 16th century Mongolian folk music, they come to prefer one or the other based on exposure and based on the preferences of others around them. When they answer ``what kind of person am I?'' by summarizing their memory and find many memories of happily listening to 16th century Mongolian folk music and few memories of happily listening to electronic dance music, they conclude and their network $p$ learns, that they must be the kind of person who likes 16th century Mongolian folk music more than electronic dance music (e.g.~as suggested by theories where individuals infer properties of the self from behavior like \cite{carruthers2011opacity, bem1967self}). And, when they answer the question ``what situation is this?'' in a way that makes it clear that the situation currently calls for a choice of which of those two kinds of music to listen to, then those two facts combine to make the answer to ``what would a person such as I do in in a situation such as this?'' be ``listen to 16th century Mongolian folk music''. Similarly, emotions may be inferred from perceptual data of varying levels of abstractness: ``what am I feeling right now?'' can be answered using knowledge of the situation ``I should be scared right now since I'm in a dark alley at night'', or as a description of more basic sensory cues like e.g.~``I must be scared since my heart is racing.''\citep{schachter1962cognitive, barrett2006solving}. Individuals may also use complex narratives to make sense of their experience and endow it with personal significance, for instance \cite{clancy2005abducted} vividly describes interviews with people who believe they were abducted by aliens to make this point.

New self-knowledge of a deeper kind can be created while answering ``what kind of person am I?''. One may not just realize they are the kind of person who prefers folk music to pop music, but also generate more consequential thoughts like ``I am interested in devoting my life to ethnomusicology'' or ``I believe my group's folk culture is worth fighting to preserve''. In these inferences, the memories, including their valence and intensity, being summarized are the data to be explained by new beliefs, and this belief-generating process is, in the end, rationally justifiable \citep{cushman2020rationalization, gelpi2020belief}.

The reason that familiarity's impact on aesthetic preferences is anomalous for the reward-based/rational theories is the direction of causality it suggests. The reward-based/rational theories start from preferences and derive behavior from there. To say, as the present theory does, that the preferences are caused by the behavior seems to be getting things backwards. Straightforward (naive) versions of reward-based/rational theories have no way to explain how this could happen \citep{bowles1998endogenous}. Of course that such anomalies exist is no surprise to theorists who defend the reward-based/rational account. There are plenty of ways to ``fix'' the problem and account for the anomaly without changing the basic framework. For instance there are ``two speed'' models where a slow-acting evolutionary process creates the preferences to use in a fast-acting learning process \citep{wang2019evolving, yaman2023emergence}. There are also game theoretic models of institutions where they are cast as ``rules in equilibrium'' which view the payoffs of one game as determined by the outcome of a different game \citep{guala2016understanding}, and many other possible approaches. These explanations are successful insofar as they go. However, from our perspective, it's unclear when the additional theoretical objects like timescales that they require us to introduce start to have the feel of epicycles. They are needed to explain the data, but they complexify the reward-based theory.

What is gained by kicking reward out of the picture in the way we do here? For one, we think the result is a more productive view, not just of aesthetic preferences, but also of norms and moral judgments, which would otherwise require more complex explanation to be understood in a reward-based/rational theory. One such example was described eloquently by J. L. Mackie when he wrote:

``Disagreement about moral codes seems to reflect people's adherence to and participation in different ways of life. The causal connection seems to be mainly that way round: it is that people approve of monogamy because they participate in a monogamous way of life rather than that they participate in a monogamous way of life because they approve of monogamy.''---\cite{mackie1977ethics}

\cite{lindstrom2018role} explained this and other such moral preferences as arising from the operation of a common-is-moral heuristic. Similarly, preferences over more mundane choices like consumer choice behaviors are also influenced non-instrumentally by experience. This fact underpins the use of product placement in advertising \citep{zajonc1968attitudinal, ruggieri2013roots}. Unlike Coke versus Pepsi type choices, where mere familiarity does likely play a large role for many people, we don't think mere familiarity is the sole factor in determining moral judgments (see Section~\ref{section:moralization}). Nevertheless, it's clear that an individual's membership in a community upholding a certain moral norm does tend to predict that they will themself uphold the same norm\footnote{See Section~\ref{section:endogenousPreferenceFormation} for more on how the pattern completion model allows individuals to differ from their community without needing to assume exogenously originated preferences.}.

Finally, one may argue that the reason to believe the reward/utility theory is that it provides a clearer picture of the function of dopamine neuromodulation in the striatum and other kinds of value-related processing in the brain (e.g.~the picture suggested by \cite{schultz1997neural}). Indeed this has often been presented as one of its selling points. It's true that the pattern completion theory does not immediately suggest a role for dopamine. However, It's worth pointing out that the pattern completion model does provide for some other natural modeling choices that the reward/utility theory does not. In particular, there are two different ways in which models built to capture the pattern completion theory may fail to emit the action that would maximize $r$: either $p$ may be insufficiently strong or the reasoning chain may be wrong. It's very natural to interpret an insufficiently powerful $p$ as a failure of crystallized semantic memory while a failure in the sequence of reasoning steps can naturally be interpreted as a failure of executive function, fluid intelligence, or planning. We think these interpretations are every bit as interesting as the dopamine interpretation, and may be equally fruitful in framing future work.

\subsection{Causes of individual preferences (endogenous preference formation)}
\label{section:endogenousPreferenceFormation}
In this section we discuss an individual actor's policy through the lens of the causes of its preferences. Actors, in each specific situation, will pick some actions over others. As they are endowed with language, they can also explicitly formulate their preferences over the state of the world, objects, theories, and pretty much anything else that can be talked about.

Here we present an actor model where individual preferences are mostly constructed from \emph{social} influences. However, it also allows for influences of non-social origin. For instance, an individual's preferences and behavior can be affected by physical properties of the environment which they discovered by individualistic trial and error. Nevertheless while such individualistic influences are describable in the model, and for completeness it is important that they be so, we don't actually think the influence of individualistic trial and error is very large. Rather, we will argue that the vast majority of our behavior is attributable to social causation\footnote{This is actually not as controversial of a claim as it may seem to some of our readers from the RL community. One need not adopt our reward-free theory to articulate the question of whether social learning or individualistic trial and error are superior in theory and more common in practice. In fact, substantial research has already been carried out on this question---and results are largely supportive for both the theoretically greater importance and empirical prevalence of social learning over individualistic trial and error e.g.~\cite{rendell2010copy, boyd2011cultural, muthukrishna2016and}. Even computational RL agents learn to learn socially in many environments \citep{borsa2019observational, ndousse2021emergent, bhoopchand2023learning, cook2024artificial}. Furthermore, there has also been substantial work comparing social learning to individualistic planning (i.e.~prospective reasoning of individuals based on their causal understanding of the situation or task). While causal understanding is helpful when available \citep{tessler2021learning}, results in this line of work typically support a picture where social learning can, and often does, work effectively in situations where individuals do not possess sufficient causal understanding to facilitate effective planning \citep{derex2019causal, henrich2021cultural, harris2021role}. The present theory will go further than considered by this line of work so far though since it will now allow us to articulate questions that pertain not just to the social learning of behavior but also to the social learning of \emph{preferences} or \emph{tastes}.}. 

Our goal in setting out this formulation is to capture social construction theories of individual preferences. Social construction theories of individual preferences hold that tastes arise as an aggregation, compression, and generalization of the data stream experienced by an individual caused ultimately by the joint behavior of the collective and the background culture (e.g.~\cite{bourdieu1984distinction}), as well as biophysical properties of the environment and actor.

Note that the sense of `endogenous' we intend here refers to the property of being caused by factors internal to a model, i.e.~not assumed from outside the space of causes the model captures within itself (those would be exogenous causation). Since our models are multi-actor models with environment state, endogenous preference formation refers to processes of preference formation that depend on causal factors located outside the individual such as in the society where they live and its culture, as well as the biophysical environment the individual inhabits.

\subsubsection{Measuring preferences}
\label{section:measuringPreferences}

Let $X$ be a set of assemblies corresponding to physical objects, states  of the world, or actions to which it makes sense to talk about an actor having preferences over (e.g.~in state $x_0$ Alice is eating an apple, in state $x_1$ Alice is eating a banana). A preference relation on $X$ is a binary relation $\succeq$ on $X$. Its associated strict preference relation is denoted $\succ$. The elements of $\succ$ correspond to statements such as ``\texttt{Alice prefers `apple` to `banana`}''.

Consider how one can empirically measure an actor's preferences over $n$ items. In one approach you could present the actor with a sequence of pairs of items to choose between (``\texttt{Alice must choose between `apple` and `banana`}''). Another approach is to ask them to rank order the full set of $n$  (``\texttt{Alice must order the following items by her attitude toward them: `apple`, `banana`, `chocolate`, ...}''). There is a sense in which, if Alice were fully rational, then the two approaches would give the same answer. However, in practice, the two approaches would be very unlikely to yield the same result (e.g.~due to ordering and framing biases in human and LLM behavior (e.g.~bias toward the first and last in the list \citep{schuman1996questions, yang2024llm}). This technique of eliciting preferences from an LLM was also used in~\cite{klissarov2023motive}. 

Lets consider the case when the choice is contextualised by an assembly $u$, for example $u=$`\texttt{Alice is allergic to apples}''.  Let $x_0 \succeq^u x_1$ indicate 
$p(\text{choice} = x_0 | z = u, o = \{x_0~\text{or}~x_1?\}) \ge  p(\text{choice} = x_1 | z = u, o = \{x_0~\text{or}~x_1?\})$. Notice that $x_0 \succeq^u x_1$ is also a binary relation on $X$ since any two items can be compared this way. The interpretation is that $u$ is the only piece of information driving the choice between $x_0$ and $x_1$ aside from $p$ itself. This is convenient since it lets us consider all information obtained from a certain source as a $u$ giving rise to its own $x_0 \succeq^u x_1$. In the next section we will consider a set of such ``influences'' on an individual which we will call the individual's \emph{guidance}.

\subsubsection{Guidance}

For a specific actor facing a decision, we can consider all of its memories in $m$ as influences, which induce preferences. However, whether a specific memory is surfaced into the global workspace $z$ depends on the actor's decision logic (e.g.~\cite{march2011logic}'s logic of appropriateness) and current circumstances.
Stable pattern of encouragement, discouragement, or simply repetition accumulate in the actors memory $m$ and get compressed and generalised in its pattern completion network $p$. Through this process they crystallize into influences that surface into $z$ with high probability.  

For example, if the actor has often experienced encouragement to eat a vegetarian diet (or observe everyone around eating only vegetables), they might internalise that as ``I am a vegeterian'', an influence which will later be retrieved (in a relevant context) and contextualize a summary function such as ``What kind of person am I?''. Alternatively, the stable pattern could  be consolidated into $p$, become implicit, and then not require an explicit representation in $z$ to influence behavior. In this case the actor would simply prefer vegetarian food, even when $z$ also contains contrary assemblies like ``meat is on sale today'' since a vegetarian's behavior would normally be unaffected by learning that meat is on sale.

There can be various patterns of encouragement and discouragement or repetition, some may be social in origin while others are not. In fact, any stable pattern of encouragement / discouragement experienced by an actor can affect their behavior. This includes stable patterns of encouragement and discouragement associated with physics like the heat produced by a fire. However, just because a piece of knowledge may be non-social in origin does not necessarily mean it was acquired in that way. One doesn't need to actually touch a flame and feel the pain of burning oneself in order to learn about it---you could hear about it from others. Similarly, you can understand how patterns of behavior in people with different social roles, which you do not occupy yourself, are encouraged and discouraged by society e.g.~a non-judge's representation of the role played by a judge in a courtroom, or a goth kid's representation of a jock.

When contemplating putting your hand in a fire you might recall to your global workspace the explicit knowledge that fire burns and this action you are contemplating would be horribly painful. But of course that would be a strange way to make the decision not to thrust your hand into the fire. More likely, you would not actually need to recall into the global workspace the knowledge that fire burns since the pattern completion network $p$ would never predict your putting your hand in it in the first place. We would say then that this knowledge has already been consolidated and thus is available implicitly. Nevertheless, it is an important aspect of the model we present that it can also handle the less common situation where you might contemplate putting your hand in the fire despite knowing it will burn, perhaps as part of a ritual. In this case you might recall explicitly to context the knowledge that fire burns in order to find a way to logically overrule it and thereby muster the courage to follow through with your painful plan.

Influences are not quantitatively comparable to one another. Encouragement and discouragement may convey intensity of feeling but this is not the same as the quantitative agreement that would be necessary to construct a shared utility function scaled to capture choice behavior under all perspectives simultaneously~\citep{hammond1991interpersonalcomparison, hausman1995impossibilitycomputility, binmore2007interpersonalcomparison}.

That inter-perspective (intra-personal) comparison of utility is impossible is especially clear when multiple perspectives are indifferent to most outcomes outside a select few. The sanctioning pattern shaping my behavior as a member of a group that forbids alcohol consumption is indifferent between brands of scotch while the sanctioning pattern shaping my behavior as a member of a whiskey club may encourage me to choose smoky varieties over non-smoky. Without context there is no way to assign a scalar utility to aggregate behavior according to both perspectives. Real people simply let their behavior be guided by one perspective in some contexts and the other in other contexts. Similarly, a politician may take actions in a particular domain with an eye toward maintaining the positive regard of a particular constituency while, in a different domain, taking actions to satisfy a different constituency. The preference relations corresponding to conventional patterns of sanctioning are ordinal, not cardinal. This argument resembles one that regards the self as a multiplicity of aspects, each with their own preference relation from \cite{steedman1985faustarrow, roccas2002social}. And, we note also that \cite{kurth2024dynamic, roccas2002social} and \cite{walzer1994thick} argue that maintaining preference and perspective diversity is healthy at both the individual and societal levels.

\subsubsection{Aggregation}

We would like to say something about how individuals change their behavior when exposed to multiple different stable patterns of encouragement and discouragement. Without context there is no way to assign a consistent scalar utility function that aggregates behavior according to multiple incommensurate patterns of encouragement/discouragement simultaneously. People simply allow their behavior to be guided by one perspective in some contexts and the other in other contexts. For instance, a politician may take actions in a particular domain with an eye toward maintaining the positive regard of a particular constituency while, in a different domain, taking actions to satisfy a different constituency. The preference relations corresponding to guidance elements are ordinal, not cardinal. This argument resembles one that regards the self as a multiplicity of aspects, each with their own preference relation from \cite{steedman1985faustarrow}.

All guidance elements (memories) get aggregated together into a single decision. All guidance elements report something into the global workspace $z$ of the pattern completion network $p$, and then it is $p$ that decides what to do. Another way to see this is that the pattern completion network $p$ is selecting which influence (guidance element) should control behavior in each context. It is also possible for $p$ to mix influences together and produces a new behavior that was not exactly suggested by any single element of the individual's guidance.

Therefore, the pattern completion network $p$ induces a latent preference ordering by which multiple possible behaviors $u_0, \dots, u_k$ can be compared. The behavior chosen above the others (i.e. with the highest probability after applying all social and non-social reasons in context) is the most preferred.

There are sometimes multiple competing reasons (social and non-social) both for and against doing/saying $u$. When the context contains multiple competing reasons then the LLM $p$ adjudicates between them. However it does not necessarily aggregate guidance elements in the same way logical deduction would aggregate them into a conclusion, and the set of reasons that come to mind may not always be the complete set of reasons that are relevant to consider \citep{schwartz1977normative, schwarz1983mood}.

At the population level, multiple reasons for the same behavior may coexist. \cite{jagiello2022tradition} gives an example of this in the \textit{puja} ritual of the Jains in India. One version of this ritual involves placing a flower on an idol. When researchers asked practitioners their reasons for placing the flower they received numerous different kinds of responses. Some responses reflected symbolic and expressive motivations, e.g. ``to become purified like a flower'' while others were straightforwardly instrumental, ``the scent of the flower makes the process of worship more pleasant''. Other participants said the placing of the flowers did not mean anything at all and simply needed to be done because that is what Jain worshippers do \citep{jagiello2022tradition, humphrey1994archetypal}. In such situations, the individual's pattern completion network $p$ may simply decide to perform the ritual. It need not produce a justification for that choice unless asked by an interlocutor to do so, but this is a different task. Thus there need not be any indecision between competing reasons for the same ritual since no reasons are needed at all until someone asks\footnote{We discuss this divergence of decision causation and decision justification in Section~\ref{section:postHocJustification}.}. The separate task of justifying the ritual, i.e., the task of producing reasons for it that the interlocutor would accept or deem valid is a different one, and indeed may be difficult to accomplish when multiple conflicting reasons exist. But, whether or not it is difficult in this way depends on the norms of deeming arguments valid that prevail for conversations of this kind taking place in the society in question\footnote{A theme to which we return in Section~\ref{section:moralization} where we discuss universalization arguments and Section~\ref{section:gadgets} where we discuss epistemic norms.}.

An important point to keep in mind is that while many guidance elements are social in origin, there are also guidance elements that are not social in origin. That is, some guidance elements are commonly shared by most people, we call these generic. Not all generic influences are social in origin. For instance, the knowledge that picking up a hot coal would lead to a burn is another kind of guidance element that most individuals share. Most individuals also have elements in their guidance which few others share. The most common category in this class of idiosyncratic guidance elements are the worldviews and obligations associated with repeated interaction in families and friendships, this will become important in Section~\ref{section:familiarPeopleAppropriateness}.

In Section~\ref{section:normsDefinition}, we will say that a behavior is normative when the guidance element controlling it is one that was created by a generically conventional pattern of sanctioning. We will also say that behavior with strangers is appropriate when it is normative. This is how social identities influence behavior. A student is meant to act in a certain way in certain situations, if I am a student, then my understanding of how a student should act in my current situation will affects my behavior. These understandings are installed by conventional patterns of sanctioning. 

However, there are also situations in which norm/identity-related guidance elements are not in charge. In particular, we talk about two of them:
\begin{enumerate}
\item idiosyncratic patterns of social sanctioning may be in control, e.g. the specific pattern of interaction one might have with specific other individuals like their family or friends.
\item stable but non-social in origin patterns of encouragement and discouragement such as the stable relationship between the act of picking up a hot coal and the sensation of pain.
\end{enumerate}

{\flushleft{\textit{\textbf{\noindent Arrow's theorem for social construction of individual preferences}}}}

The aggregation of guidance elements is like the aggregation operation in social choice theory, but in this case it's all happening inside one individual. This reverses the construction studied in social choice theories like those associated with \cite{deCondorcet1785} and \cite{arrow1950social}. We are now explaining individual decisions as a ``social'' choice over influences on the self. So we aggregate the social into the individual instead of the individual into the social. 

Social choice theories normally begin from exogenously provided individual preferences and study how to aggregate them into a single social preference and choice, as in democratic decision making \citep{vanDeernen1991socialgame}. In contrast, our theory proceeds in the reverse direction since we construct each focal individual's preferences from multiple different conventional patterns of social approval and disapproval that influence them.

This ``multi-agent'' view of the individual raises questions about what happens when an individual's ``parts'' do not agree with one another. The observed social behavior results in inconsistency, changing one's mind, inability to follow through later on choices made at an earlier time, framing effects changing choices, getting distracted by the short-term, etc. Many of these effects can be understood in terms of sub-personal strategic behavior \citep{schelling1980intimate}\footnote{Such time-inconsistency of human preferences was recently discussed in an AI context by \cite{carroll2024ai}. In humans, and coming from a more clinical application-oriented perspective, these effects are the primary subject matter in the behavior change subfield of health psychology~\citep{michie2014abc}.}. An individual is the product of all the influences that shape them, and thus we either behave as fanatics devoted to one influence above the others (dictatorship in terms of Arrow's theorem, see Section~\ref{section:endogenousPreferenceFormation}), or otherwise reflect some kind of compromise between  misaligned influences. The next two sections address important implications of this social construction of individual preferences.

Arrow's theorem says that no social choice rule simultaneously satisfies all four of the rationality properties below. It is a shocking result in social choice theory, given its apparent implications for democracy \citep{pildes1990arrowsdemoc}. However, in the context we consider here: the social construction of individual preferences from guidance elements, we don't think it turns out to be nearly so shocking. Everyone knows that humans don't always act rationally. So in this context, implications of Arrow's theorem should be understood here mostly as structural constraints on aggregation in the context of our actor model which serve usefully to align it with well known properties of human cognition like framing effects.

The relevant rationality properties are interpreted as follows (note how different these interpretations are from those of social choice theory): 

\begin{enumerate}
  \item \textbf{Independence of irrelevant alternatives} Unsurprisingly, independence of irrelevant alternatives often does not hold for individual preferences. There are many well-known context change effects where an individual's preferences reverse in the presence of an irrelevant alternative~\citep{huber1982prefrev, tversky1993contextdepprefrev, spektor2021conexteffectsprefrev}. 

  \item \textbf{Non-dictatorship} This means that no single pattern of encouragement/discouragement determines the focal individual's choice in all situations. When this property fails, the individual would be some kind of fanatic \citep{steedman1985faustarrow}. Many people integrate influences into their preferences from multiple identities (and other sources). Though we could imagine a person who only responds to a sanctions of a particular scope, perhaps a fanatic who pays no heed whatsoever to non kin.
  
  \item \textbf{Unrestricted domain} No change to the interpretation of this property.

  \item \textbf{Pareto condition} No change to the interpretation of this property.
\end{enumerate}

Can preferences be cyclic? Yes! For example: Choosing a Vacation Destination. 
Context: An individual is deciding on a vacation destination. They are considering three options: a relaxing beach vacation, an adventurous mountain climbing trip, and a culturally immersive trip to a foreign city.

Non-Transitive Preferences:
\begin{itemize}
    \item Their ``health-conscious'' identity prioritizes rest and relaxation, leading to a preference for the beach vacation over the mountain climbing trip (beach $\succ$ mountains). 
    \item Their ``adventurous'' identity prioritizes excitement and challenge, leading to a preference for the mountain climbing trip over the cultural trip (mountains $\succ$ culture). 
    \item Their ``intellectual'' identity prioritizes learning and personal growth, leading to a preference for the cultural trip over the beach vacation (culture $\succ$ beach).
\end{itemize}

There is nothing mysterious about such ``intra-personal Concorcet cycles''. The framing of the choice by the options contained therein usually provides enough context to prime an actor to make their choice on the basis of a particular guidance element. When really stuck (i.e.~when there is little context to decide), then indecision or vacillation between different options could be the outcome, perhaps continuing until time pressure forces an irreversible choice \citep{steedman1985faustarrow}.

Why are individuals sometimes guided by norms but other times prefer to follow their idiosyncratic guidance elements like those associated with their families and friends? The choice depends on which guidance element $p$ prioritizes in the particular context. Choosing between normative and idiosyncratic guidance involves selecting the social choice rule implemented by $p$. Generic influences lead to norm-guided behavior, while idiosyncratic influences cause deviations from the norm.

It is possible to think of the pattern completion network as making decisions according to a more structured choice architecture. One possibility is to consider something like the \cite{axelrod1967conflict} approach reviewed in \cite{vanDeernen1991socialgame} where individuals seek to join winning coalitions with minimal internal conflict (minimal dispersion of preferences within the coalition). In our context this would become a model where an individual allows a certain ``coalition'' of stable patterns of encouragement/discouragement to determine their behavior most of the time and picks a coalition with minimal internal conflict to play that role. Just as a governing political party seeks to control which issues come to a vote in order to avoid exposing places where their coalition internally disagrees, an individual may also avoid situations where multiple influences associated with different aspects of their identity produce conflicting suggestions. In this way, our decision making model could perhaps produce results similar to that of models where individuals seek to minimize cognitive dissonance \citep{festinger1957theory, harmon2019introduction}.

\subsubsection{The challenge of learning preferences in a social linguistic environment}

In this section we show that learning preferences from other actors is one mechanism that can lead to polarization. This is because, in everyday experience, language only transmits limited information about \emph{non-maximal} social preferences. 

For instance, most of the time, preferences are communicated through language by simply suggesting the most preferred choice, e.g. ``$a_1$ is how we do it around here''. It is rare for the full preference relation to be transmitted through language since that would be more like ``around here, we prefer $a_1$ to $a_2$ to $a_3$'', which is too cumbersome, and not how people talk. This suggests that language expressions are sometimes limited as means of communicating preferences over actions.

For example, consider an individual who has four (and only four) important social identities (as in Section ~\ref{section:stylizedFactContextDependence}). Let's say that this individual is simultaneously a communist, a christian, a cowboy, and a computational biologist. Aggregating across all communists and people who sanction communists, probing their full preference relations, let's assume it would emerge that they want the focal individual to act as though $a1 > a2 > a3$. Similarly, the ideal christian identity could want the focal individual to act as if $a3 > a2 > a1$, the ideal cowboy identity could agree with the communist and wants them to act as if $a1 > a2 > a3$, and let's say that the ideal computational biologist agrees with the christian identity and suggests they should act as if $a3 > a2 > a1$. We can describe this polarized situation with the following notation:
\begin{equation}
    [\succ]_{c} = \begin{bmatrix}
        a_1 \succ a_2 \succ a_3 \\
        a_3 \succ a_2 \succ a_1 \\
        a_1 \succ a_2 \succ a_3 \\
        a_3 \succ a_2 \succ a_1
    \end{bmatrix}
\end{equation}
However, the focal actor never gets to learn directly from the `ideal' identities. They would only ever learn from data obtained in specific contexts. So the data they could actually bring to bear on any particular decision in any particular context might look like: 
\begin{equation}
    [\succ]_{c} = \begin{bmatrix}
        a_1 \\
        a_3 \\
        a_1 \\
        a_3
\end{bmatrix}
\end{equation}
Notice that the extreme positions (i.e., $a_1$ and $a_3$) represent the top-ranked choices of every observed actor who would actually prefer in context $c$ to select an intermediate choice (e.g., $a_2$) over both opposing extremes. This shows that an actor who learns their preferences from a linguistic environment that rarely conveys full preference information may infer that actions $a_1$ and $a_3$ are conventional social behavior in context $c$ and should take precedence over $a_2$, since it would never be observed. But this would mean the actor would end up learning to take either extreme position, not the more moderate one.

What about consolidation and implicit learning that modifies the weights of $p$? This construction of individual preferences through language still only gives a preference formation process that builds an individual's $p$ by the aggregation of top-ranked preferences (unless people habitually describe their full preference relation, which seems unlikely). The situation is likely even worse for aggregation by consolidation since most likely, only the top-ranked choices would appear frequently enough to crystallize (in the implicit learning sense of Section~\ref{section:learningAndExplicitImplicitOperation}).

This section has described one mechanism by which polarization can emerge in accord with our theory of appropriateness. Of course, any phenomenon as multifaceted as polarization could be captured using many different computational mechanisms. Nevertheless, we do think the fact that our predictive pattern completion model of actor decision making reproduces some polarization effects ``for free'' could be a useful modeling affordance to explore in future work.


\section{Appropriateness with strangers}
\label{section:appropriatenessWithStrangers}

In this theory, appropriateness is a socially determined property of individual behavior. There are two ways in which a behavior may be appropriate. First, it may be \textit{normative}. This is the topic of the present section, which is about appropriateness in interactions with strangers. Second, a behavior may be appropriate by virtue of the history of specific personal relationships. We discuss such relationship-based appropriateness in Section~\ref{section:familiarPeopleAppropriateness}, which is about appropriateness in interactions with friends and family.

We propose the definition: a behavior is normative when it is suggested by a generically conventional pattern of sanctioning. In this section we formalize what it means for a behavior to be normative and discuss how the definitions we provide apply both explicitly to verbalizable norms and implicitly to non-verbalizable norms.

Core to our theory are the concepts of \textit{conventionality} and \textit{sanctioning}. Conventions are simply patterns of behavior that are produced by reproduction and the reason to reproduce them is the weight of precedent \citep{millikan1998language}. Conventions operate in \textit{scopes} of variable size, ranging from pairs of friends (narrow) to global culture (very generic). Sanctions in our theory are sequences of symbols (explicit case) or sets of such sequences (implicit case) which encourage or discourage\footnote{Some readers have tripped here on words like encourage, discourage, and sanctioning; thinking to themselves ``surely these are reward-related words?''. If this is you, then what you need to remember from Section~\ref{section:patternCompletionIsAllYouNeed} is that we are making a \emph{parsimony} argument. We are not denying incentive effects. And, we are not claiming that it is impossible to reduce our theoretical concepts to reward-based concepts. Of course you can! Instead, our claim is that doing so would be like giving a complicated answer to a simple question. For instance, the concept of sanctioning we define here does not involve a scalar reward signal. If we can convince you that it ultimately accounts for the same facts as the reward-based theory, this would make ours a simpler way to fit the data since it sidesteps all the thorny problems of reward function specification.} a change in a behavior associated with a particular social role. Patterns of sanctioning may be conventional. When a pattern of sanctioning is conventional over a generic scope (i.e.~a scope that incorporates much of society), then behaviors suggested by such a pattern are called \emph{normative}.

In the rest of this section we define conventions and sanctions formally and derive some of their properties. The properties we highlight suggest explanations for the stylized facts of appropriateness we introduced in Section~\ref{section:desiderata}. 

\subsection{Summary of a computational theory of normative appropriateness}

Conventions are behavioral patterns that spread through the population simply by the weight of precedent. Although they can generate stable patterns and establish some forms of group behaviour, they by definition cannot inhibit behaviour otherwise favored by individuals. Normative behavior, on the other hand, may include behaviors that individuals would not otherwise select, this is because norms, as we define them, are supported by conventional patterns of sanctioning, which have the effect of encouraging or discouraging certain actions. Moreover, due to sanctioning being conventional, norms can spread through individuals who are not directly involved in producing the behavior in question itself (e.g. norms applicable to other social roles). Sanctioning thus serves as a powerful medium for spreading behavioral patterns and stabilizing them (e.g.~against free riders). Put together, conventions and norms are powerful forces that shape groups and allow for sophisticated joint behavior to emerge and be stable.

In what follows, we build a computational model of conventions and norms in groups of generative actors. We start from defining certain properties that an actor has to have for conventions and norms to emerge and sustain themselves in their populations.

This section presents the following hypothesis concerning the emergence of norms, which in this view, govern appropriateness of behavior with strangers:
\begin{enumerate}
    \item Convention-sensitive actors adopt and follow patterns of conventional behavior.
    \item Sanction-sensitive actors are actors that increase or decrease the probability of emitting a particular behavioral patterns in response to particular signals, which we call sanctions.
    \item A population of convention and sanction sensitive actors can develop generic (large scope) conventional sanctioning patterns, which creates norms---behavioral patterns that are supported by conventional sanctioning. 
\end{enumerate}

After establishing the basic properties of conventions, sanctions, and norms, we then consider two important subclassses of norms: implicit and explicit (Section~\ref{section:implicitVsExplicit}). Explicit norms are associated with verbalizable reasons for action in context. Laws are the most important kind of explicit norms. Implicit norms, on the other hand, cannot be verbalized in a standard way, though individuals can still evaluate whether behavioral instances are transgressions or not and typically agree with each other's assessments (per context, per community). Using the concepts of explicit and implicit norm we go on to discuss how they may be represented in the brain (Section~\ref{section:representingImplicitVsExplicit}),  moralization (Section~\ref{section:moralization}), how norms change both via cultural evolution as well as via deliberate collective action (Section~\ref{section:normChange}), and to articulate a view of norms as technologies (Section~\ref{section:gadgets}) with implications for justification and rationality (Section~\ref{section:postHocJustification}).

\subsection{Conventions}
\label{section:conventions}

``Conventional patterns are exemplified rather than other patterns owing only to historical accident, but having occurred, they cause their own recurrence.''---\cite{millikan2003defense}.

Saying ``hello'' upon picking up a ringing phone is a convention. It is clear that something must be said to indicate to the caller that you have picked up the phone and are now ready to speak. It is equally clear that any other word could do the job just as well (it is arbitrary), and in many other places in the world where people do not speak English, other words are used upon picking up the phone (it is culture dependent). If others in my culture stopped saying ``hello'' and instead started saying ``X'' then I would have reason also to switch to X \citep{marmor2009social}.

Our account of conventions follows \cite{millikan1998language}. Conventions are patterns of behavior that (1) are \emph{reproduced}, and (2) the reason they are reproduced is the \emph{weight of precedent}.

When we say a pattern of behavior is reproduced we mean that its form is derived from a source pattern such that, if the source pattern were different in some respect, then the reproduction would also be different in that respect. Notice that this is much weaker than exact copying. Most of the time a copy will not match its source in all respects. Reproduction may be the result of various mechanisms. Direct imitation of one individual by another is just one of many ways that reproduction may occur. Others include: (a) one person could simply tell another to reproduce the pattern, (b) a person may adapt their behavior to fit the source's behavior, either by trial and error or by prospectively considering consequences of various actions they might take, for instance consider a driving scenario where you are forced to drive either on the left or the right side of the road to avoid collision; \cite{millikan1998language} calls this \emph{counterpart reproduction}, and (c) counterpart reproduction may also be asymmetric. For instance if certain nuts are made to fit certain bolts then the overall pattern of nut and bolt thread gauges is reproduced even though nuts and bolts are not the same pattern as one another.

The first step toward formalizing this notion of pattern reproduction is to define what it means for two utterances to have similar meaning.

\begin{defn}[$\epsilon$-similar meaning $u\sim v$] Utterance $u$ has an \textit{$\epsilon$-similar meaning} to $v$ for an actor's pattern completion network $p$ in context $c$ if:
\begin{equation}
    \textbf{KL}\left[p(\cdot|c) || p(\cdot|r^{u\rightarrow v}(c)\right] < \epsilon,
\end{equation}
where $r^{u\rightarrow v}(c)$ is an operation that replaces all instances of $u$ with $v$ in string $c$.
\end{defn}

The property of a pattern being conventional is never due sole to properties of the pattern itself. Rather, the conventionality of a given pattern is always a function of the population of actors and their behavior. A pattern is only conventional in a particular population if actors in that population actually reproduce it due to the weight of precedent. So next we define what it means for an individual actor to be \emph{sensitive to convention}. Then following that, we aggregate individuals into relevant populations and define the concept of a convention's \emph{scope}.

While, in general, we expect the fine-tuning (consolidation) process by which pattern completion networks are trained (see Section~\ref{section:novelBioInterpretation}) will lead individuals to have different pattern completion networks from one another, we make a simplifying assumption in this section and the next which asserts that all actors share the same $p$ and summary functions $\{z_j\}$. This can be justified when most actors in a population experience a reasonably similar distribution of experience. Moreover, nothing we say here should depend on individuals having exactly the same $p$ and $\{z_j\}$, but rather will still work as long as they are close enough. In this restricted world, individuals have persistent individuality solely owing to their distinct memories $m^i$. An intuitive interpretation of a shared $p$ is that of a common culture,  shared by the group and internalised by individuals.

Note that when we say, following \cite{millikan1998language}, that for a pattern of behavior to qualify as a convention the reason it is reproduced must be the weight of precedent, we exclude cases where the pattern is reproduced because of its inherent superiority at producing a desired result or because of ignorance of alternative methods. Skills are often reproduced by social learning but that does not make them conventional. A pattern of behavior is only conventional if the reason for reproducing it is mainly that others do so. Decorating for Christmas with red and green is conventional since one does so because others also do so and thus all recognize those colors as signifying Christmas. The specific choice of red and green may be a historical accident. If history had unfolded differently then other colors might have come to signify Christmas. Thus conventions are, in this particular sense, \emph{arbitrary}.

For another instructive example, consider that the practice of taking measurements in centimeters proliferated along with rulers marked in centimeters (counterpart reproduction). The reason to measure in a conventional unit rather than inventing a novel one is that it is easiest to compare with the many other measurements already taken using the conventional unit (weight of precedent).

What about the conventions of language itself? The meanings of words \textit{in context} is conventional. Consider how the meaning of the word \textit{cold} in ``you're as cold as ice'' may in one conversational context refer to physical temperature and in another context refer to a personality trait \citep{zada2023shared, grice1957meaning}. Conventions determine word meanings independently per context.

\subsubsection{Actors reproduce conventions due to the weight of precedent}

In this section we define what it means for an individual actor to reproduce a convention. Remember, that for a behaviour to be conventional it should be derived from a source behavioral pattern, such that if the source pattern changes in some respect, then the reproduced behavior would also be different in that respect \citep{millikan1998language}. We are going to formalise this using a counter-factual memory operation which will re-write the actor's memory of the source pattern and state that this should cause the policy to change accordingly.

To make the following definitions clearer and easier to understand, we are going to make some simplifications to notation and the definition of the actor (see section~\ref{section:howIndividualsMakeDecisions}). First, we are going to drop the time dimension and consider the actor and its policy in one moment. We are also going to make an assumption that the state of the actors global workspace $z$ can be fully reconstructed from its memory $m$. This doesn't contradict the definitions in section~\ref{section:howIndividualsMakeDecisions}, since any assembly can write its state into the memory $m$ instead of directly conditioning its state on the previous one. This way we can simply write down the policy of an actor as $p(a|z(o,m))$, where $z$ is the state of the global workspace, $o$ is the current observation and $m$ is the state of the memory. We will address the temporally extended conventions later in this section.

\begin{defn}[record of action] Let $[o, j:a]$ denote a record of an action of $a$ of any actor $j$ in response (following) to observation $o$. Square brackets stand for concatenation of strings and semicolon denotes actor attribution, that is $j:a$ means actor $j$ taking action $a$. This can itself become an observation of another actor. In practice, this could be represented in a string format as "Alice sees Bob. Alice: Hello, Bob!".
\end{defn}

In general, conventions depend strongly on context. However, for didactic purposes, we first define convention sensitivity in the context-free case and then go on to give a full definition that takes context into account.
\begin{defn}[context-free convention sensitive] \label{def:context_free_convention_sensitive}Let $m'$ be a counterfactual state of $m$, where for each $\Tilde{o}\sim o$ ($\epsilon$-similar meaning to $o$), $\Tilde{a}\sim a$, and $a' \nsim a$, all occurrences of $[\Tilde{o},j:\Tilde{a}]$ in the memory are replaced with $[\Tilde{o},j:a']$. Actor $i$ with pattern completion network $p$ is \textit{convention sensitive}, if 
two conditions hold. The first conditions is:
\begin{equation}
    p(a|z(o,m')) < p(a|z(o,m)),
\end{equation}
which means that the action $a$ is less probable given the counter-factually modified memory $m'$, than given the original one $m$. 
The second condition is:
\begin{equation}
    p(a|z(o,m')) < p(a'|z(o,m')),
\end{equation}
\end{defn}
which means that the counter-factually inserted action $a'$ will become more probable to emit than the original action $a$, given memory $m'$.
We assume that $m$ contains at least one entry $[\Tilde{o},j:\Tilde{a}]$. Therefore this definition implies that (1) sampling $a'$ is more likely than sampling $a$ (in this context), and (2) this is because of an intervention where part of the actor's relevant memory was modified to sample $a'$ more (in this context).

Put simply, definition~\ref{def:context_free_convention_sensitive} says that if one would edit all the memories of the actor where them (or any other actor) use "hello!" or a word with similar meaning with something else, say "duck duck duck", then the actor would start using "duck duck duck" as a greeting. 

\begin{defn}[context-aware counter-factual memory editing $R^{a \rightarrow a'}_f(m, o, c)$]
Now we can define a partial context-aware counterfactual memory operation $R^{a \rightarrow a'}_f(m, o, c)$ which modifies fraction $f\in (0,1]$---\textit{the counter-factual weight of precedent}---of contextually relevant memories as follows:

For all $a,o$ such that $\Tilde{o}\sim o, \Tilde{a}\sim a$.
\begin{enumerate}
\item Shuffle entries $[\tilde{o},k:\Tilde{a}]$ from the memory $m$.
\item For the first $f|m|$ entries do the following:
   \begin{enumerate}
   \item Compute the projected context (i.e. the global workspace evoked by $o$) $\tilde{c} = z^i_k(\tilde{o})$ as defined in section~\ref{section:howIndividualsMakeDecisions}. 
   \item Only for the memories where: i) the context $\tilde{c}=c$ ii) $\tilde{o}$ has $\epsilon$-similar meaning to $o$ in context $\tilde{c}$ iii) \textbf{and} $a'$ is not $\epsilon$-similar to $a$ in context $\tilde{c}$, do the following:
   \begin{itemize}
       \item[-] replace the entry with $[\tilde{o}, k: a']$.
   \end{itemize}
   \end{enumerate}
\item Return the modified memory.
\end{enumerate}

\end{defn}

Next we define contextual convention sensitivity:

\begin{defn}[contextually convention-sensitive] \label{def:contextual_convention_sensitive} Actor $i$ with pattern completion network $p$ is \textit{contextually convention-sensitive} with respect to context $c$, if for any $a'$ not $\epsilon$-similar to $a$ in context $c$, and assuming $m$ contains at least one entry $[\tilde{o},k:\Tilde{a}]$, then the following holds:

\begin{equation}
p(\mathbf{a}|z(o, R^{a \rightarrow a'}_1(m, o, c)| z^i = c)) < p(\mathbf{a}|z(o, m | z^i = c)),
\label{eq:csc_old_action_less_probable}
\end{equation}
\begin{equation}
p(\mathbf{a}|z(o, R^{a \rightarrow a'}_1(m, o, c) | z^i = c)) < p(\mathbf{a}'|z(o, R^{a \rightarrow a'}_1(m, o, c) | z^i = c)),
\label{eq:csc_new_action_more_than_old_action}
\end{equation}
and for $f < f'$,
\begin{equation}
p(\mathbf{a}'|z(o, R^{a \rightarrow a'}_f(m, o, c) | z^i = c)) < p(\mathbf{a}'|z(o, R^{a \rightarrow a'}_{f'}(m, o, c) | z^i = c))
\label{eq:csc_wight_of_precedent}
\end{equation}
\end{defn}

Here the bold Latin letter $\mathbf{a}$ stands for the set of all strings that are $\epsilon$-similar to $a$. Eq.~\ref{eq:csc_old_action_less_probable} says that if we change the memories of responses to $\tilde{o}$ (memories $\epsilon$-similar to $o$) from $a$ to $a'$ by counter-factually editing them, than the probability of emitting $a$ in response to $o$ will go down.
Eq.~\ref{eq:csc_new_action_more_than_old_action} says that the same memory operation implies that the probability emitting $a'$ in response to $o$ should be higher than emitting $a$.
Finally eq.~\ref{eq:csc_wight_of_precedent} says that the probability of emitting $a'$ increases together with the proportion of modified precedents, which is $f$. 

Again, in words: the more the precedent of reacting to certain observations with a specific actions that the actor has in its memory, the more likely it is to reproduce it; and it doesn't matter whether it is the actor's own behavior, or the behavior of other actors it observes, since the $R$ operation loops over observations of all actors in the memory $m$.

To use a colloquial illustration again---definition~\ref{def:contextual_convention_sensitive} says that changing all the memories of saying ``hello'' in \textit{a specific context} to ``duck duck duck'' would change the action probabilities in that specific context.

The following definition was inspired by \cite{millikan1998language}.

\begin{defn}[Reproduced due to weight of precedent]
\label{def:reproduced_by_actor}
Let $a,a'$ be two distinct sequences of symbols with $\epsilon$-different meaning in context $c$, and $f,r$ be numbers between 0 and 1. An actor with memory $m$ and pattern completion network $p$ $(f,r)$-\textit{reproduces} $a$ from $o$ in context $c$ when the following three statements hold:

\begin{equation}
p(\mathbf{a}|z(o, m) | z^i = c)) > r
\end{equation}
and there $\exists f(r)$  such that 
\begin{equation}
p(\mathbf{a}'|z(o, R^{a \rightarrow a'}_{f}(m, o, c) | z^i = c)) > r.
\end{equation}
so $f$ is the minimal fraction of the actor's memory where $a'$ must follow $o$ if their probability of emitting $a'$ is to exceed $r$.

And lastly the probability of emitting $a'$ increases monotonically with $f$.

$\forall f' > f$ there $\exists \delta$, such that:
\begin{equation}
p(a'|z(o, R^{a \rightarrow a'}_{f'}(m, o, c) | z^i = c)) - p(a'|z(o, R^{a \rightarrow a'}_{f}(m, o, c) | z^i = c)) > \delta(f)
\end{equation}

\end{defn}

We will call $r$ the rate of reproduction and $f$ the counter-factual weight of precedent.

\subsubsection{Collective actors, scope, and conventions in populations and groups}

To move our analysis from a single actor to a collective of actors, we are going to show that we can treat the collective in the same way as a single actor. The actors in a collective form a distributed pattern-completion engine, where each is responsible for completing their respective part of the pattern.
A convention emerges whenever the weight of precedent causes a particular joint pattern to be reproduced. Recall now the case of counterpart reproduction (Section~\ref{section:conventions}) where every individual in a population plays their own part of the convention and may differ from the parts played by others. The idea of counterpart reproduction from \cite{millikan1998language} is important for defining conventional behavior at the level of a group or the population as a whole.

Before we proceed to define conventions on the population level we need to define action at the level of a collective of actors.
Let the \textit{collective of actors} be a set $S$. Then the policy of a collective is:
\begin{equation}
    P(\vec{a}|\vec{z}(\vec{m}))=\prod_{j\in S}{p(a_j|z(m_j))}
    \label{eq:collective_policy}
\end{equation}

The probability of an action of a collective $\vec{a}$ is the product of the probabilities of each actor performing their respective action (or an $\epsilon$-similar equivalent).
Notice that although we treat $\vec{a}$ as a vector, it is also just a long string of a format $[l_0:a_0, \dots, l_{|S|}:a_{|S|}]$ and a collective of actors can thus be treated as just another actor that emits and consumes strings of symbols. We can then use  definitions~\ref{def:context_free_convention_sensitive},\ref{def:contextual_convention_sensitive} from the previous section with the collective actor, and connect individual properties to collective properties using eq.~\ref{eq:collective_policy}. 

\paragraph{A collective of convention sensitive actors is convention sensitive.}
Let's define memory operations on a collective as simply the same memory operation ($R^{a \rightarrow a'}_f(m, o, c)$) on each of the members of the collective.
Since  $\forall i, \forall x: 0 \leq f_i(x) < g_i(x) \leq 1 \rightarrow \prod_i f_i(x) < \prod_i g_i(x)$ and all the inequalities in the definitions are for probabilities and so by definition in the $[0,1]$ interval, we can conclude that indeed a collective of (contextually) conventions sensitive actors will be (contextually) convention sensitive.

\paragraph{A collective reproduces $f,r$ action $\vec{a}$ from $o$ conventionally if its members $f,r'$ reproduce their respective parts and $r<{r'}^{|S|}$, where $|S|$ is the number of individuals in the collective.}
The relationship between $r'$ and $r$ means that for $\vec{a}$ to have high reproduction rate $r$, the reproduction rate of individual conventions should be significantly higher, since any actor failing to do their part breaks the action of the collective. As an example, consider how a person who doesn't know how to move around in a busy public space, for example the London underground metro, can potentially disrupt collective behaviour by performing inappropriate actions, like stepping on train tracks.

 To have a conventional behavior extended in time, we would need each step of the behaviour be conventionally reproduced by the collective. Each step would generate a corresponding observation, to which the collective will conventionally respond with the next action. This will telescope the product further over time. 
Let $[\vec{a}_t,\vec{a}_{t+1}, \dots, \vec{a}_T]$ be a sequence of actions of a collective. Then
\begin{equation}
    P([\vec{a}_0,\dots, \vec{a}_T]|\vec{z}(\vec{m_0}))=\prod_{t\in[0,T]}{\prod_{j\in S}{p(a_{j,t}|z(m_{j,t}))}}
    \label{eq:collective_policy_temporal}
\end{equation}

Notice, that the products of probabilities can be vanishing small if any of the probabilities are not extremely close to $1$. In our case it is important to remember that we are always looking at all strings of symbols that have $\epsilon$-similar meaning in the definition~\ref{def:reproduced_by_actor}, which pushes the probabilities up.

It also follows that only through \emph{collective action} can an established convention be changed, individuals are unable to unilaterally cause conventions to change. Unilateral action can only disrupt the current conventions, but won't establish a new one. However, there is some large threshold fraction of individuals for which, if they all change from emitting $a$ to emitting $a'$ then that would create sufficient weight of precedent for the other actors in the population also to change their behavior in the same way (similar to the critical fraction in \cite{schelling1973hockey}).

There are many ways that the weight of precedent could end up causing particular patterns to be reproduced. Notice that many do not involve deliberate imitation. For instance, the fact that replicator dynamics converges to evolutionary stable states of the Lewis signalling game \citep{huttegger2014some} suggests that natural selection of individuals with heritable signalling strategies is one way by which conventions can emerge. These processes have the character of random symmetry breaking \citep{skyrms1998salience}. Once they start off drifting in one direction they gather momentum, i.e.~the weight of precedent builds up.

Each convention $u$ has a \textit{scope}---the set of individual that reproduce it. The convention may exist just between two people, or within a family. Or they may exist broadly in an entire society with a shared culture. In particular, conventional patterns of sanctioning are very important to the present theory. Many conventional patterns of sanctioning operate widely over society: e.g., the way littering in parks is generally discouraged for anyone. Other patterns of sanctioning can develop within smaller groups of individuals who know one another well e.g., the way a particular family might encourage its members to wash their hands upon returning home from being outside (an idiosyncratic rule in one author's family when they were young). As an aside, notice that much of the game-theoretically significant difference between the case of repeated interactions with familiar individuals versus one-shot interactions with strangers sampled repeatedly from a population is handled in our theory by the difference between narrow scope (idiosyncratic) and wide scope (generic) conventions.

The widest scope in a society is called \textit{generic}. A convention can be said to apply generically if everyone takes part in reproducing it and they do so on account of the weight of precedent.

\textbf{Corollary: }when there is a convention, you can predict what an individual $i = i_0$ will do from  knowledge of what other individuals $i \neq i_0$ are doing.

The last thing to note about conventions before we move on to discussing norms is that both consolidated (implicit) and unconsolidated (explicit) conventions exist (here we are using terminology we established above, in Section~\ref{section:learningAndExplicitImplicitOperation}). Explicit conventions are mediated by memory retrieval from $m_t$, and are thus always precisely articulable in natural language. Implicit conventions are mediated by the weights of $p$ and thus may not always be precisely articulable in language in a way that all would agree on.

\subsection{Norms}
\label{section:normsDefinition}

In our theory, a behavior is normative when it is encouraged by a generically conventional pattern of sanctioning (or its complement is discouraged by a generically conventional pattern of sanctioning). Thus it is necessary to first clarify what we mean by sanctioning before returning to discuss norms and how they influence appropriateness among strangers. Our definition is compatible with that of \cite{chudek2011culture}: ``norms are learned behavioral standards shared and enforced by a community''. Norms depend on collective action to classify behaviors of certain classes of people in certain situations as either appropriate or inappropriate. This requires many individuals to apply mutually coherent classifications. These mutually coherent classifications are conventions i.e.~reproduced over time due to the weight of precedent. Like conventions, norms can only be changed through collective action. Individuals cannot change norms on their own.

Norms reflect the broader culture. That is, people not directly involved in a particular interaction nevertheless classify it along normative dimensions and may become motivated to sanction as a result \citep{fehr2004third}. As such, norms are not sensitive to personal relationships between individuals or idiosyncratic beliefs or knowledge of individuals. It is the individuals who represent, remember, and apply norms, but their force comes from the fact that many individuals represent, remember, and apply them in a similar way. This is the major difference between norms and relationship-based traditions: norms bind even interactions of strangers \citep{seabright2010company}. Whereas, the traditions that may emerge in groups of people who know each other only govern appropriate interactions in more circumscribed contexts (see Section \ref{section:familiarPeopleAppropriateness}).

\subsubsection{Sanctions}
\label{section:sanctionsDefinition}

\textbf{Encouragement and discouragement}

Social approval and disapproval are especially important to this theory of appropriateness. Sanctions are utterances or actions that convey social approval or disapproval. Importantly, sanctioning in this view does not necessarily have any direct effect on the ``target'' of the sanction. Rather, a key feature of sanctioning is often to signal what is normative to \emph{everyone else} who was not themselves sanctioned but still know that the sanction occurred (e.g.~via gossip). For instance, you may tell your friends that you feel it is appalling that the president took a particular action in a particular situation. The effect is to update your friends' views about how the cultural groups you partake in regard appropriate behavior for a president. If they regard themselves as members of the same cultural groups then their behavior will likely move in the same direction. In this way the sanction may be regarded as targeting the culture's conception of a role. This is different from an attempt to teach an individual. It aims to change everyone's conception of what it means for an individual with a particular role to act appropriately in a particular situation.

For the sanctioning to be used as a medium of behavioural adjustment in a population, actors have to be sensitive to a sanctioning signal. We define sanction sensitivity as follows.

\begin{defn}[Sanction Sensitivity]
Let $\Tilde{o}$ with $\epsilon$-similar meaning to $o$, and $\Tilde{a}$ be $\epsilon$-similar to $a$. Let $m'$ be a counterfactual state of $m$, where at least one occurrences of $[j:\Tilde{a}]$ in the memory is replaced with $[j:\Tilde{a}, i:s]$. An actor with a pattern completion network $p$ is sanction-sensitive to sanctioning signal $s$, if:
\begin{equation}
    p(a|z(o,m')) < p(a|z(o,m)),
\end{equation}
We can define positive sanctions in the same way, by inverting the inequality.

\end{defn} 

\begin{defn}[Contextual Sanction Sensitivity] \label{def:contextual_sanction_sensitivity}
Let $\tilde{o}$ have $\epsilon$-similar meaning to $o$ in context $c$, and $\tilde{a}$ be $\epsilon$-similar to $a$ in context $c$. Actor $i$ with a pattern completion network $p$ is \textit{contextually sanction sensitive} to sanction signal $s$ in context $c$, if:

\begin{equation}
p(a|z(o, R(m, j, [j: \tilde{a}, i:s], \tilde{o}, c) | z^i = c)) 
<
p(a|z(o, m | z^i = c))
\end{equation}

We assume that $m$ contains at least one entry $[ \tilde{o}, k: \tilde{a}]$ such that $\tilde{o}$ is $\epsilon$-similar to $o$ in the context $z^i_k(\tilde{o})$.
\end{defn}

This concept of sanctioning differs from reinforcement/punishment. In our model, the approval/disapproval conveyed by the sanction need not be aimed at the focal actor. The focal individual need not themself experience being sanctioned. For an individual to maintain their representation of a norm it is enough for them to hear about sanctions being applied to others in the community. For instance, an individual may learn from sanctions by hearing about them in gossip \citep{mueller2024life}. Even if you were never yourself in context $c$, you could still change the way you would act in that context in response to learning a particular action performed by someone else was met with disapproval or approval in $c$. Actor A shaming actor B because they transgressed a taboo constitutes sanctioning to all who hear them, regardless of whether actor B themself hears or not. 
Moreover, whether a particular action/utterance constitutes approval or disapproval (or is neutral) in context $c$ is conventional (in the sense of Section~\ref{section:conventions}). Therefore it is caused endogenously, so modelers are not required to make exogenous assumptions about it, as they must in reward-based theories.

Sanctions are reasons to say or do $a$ in context $c$, e.g.~a reason to do/say $a$ is that others will approve of it (i.e. they would encourage a choice of $u$). When choosing in context $c$ between behavior $a$ and behavior $a'$, the knowledge that others disapprove of $a$ in context $c$ provides sufficient reason to pick $a'$ instead. There are also non-social reasons to do/say $a$, e.g. ``I know that it would be physically painful or dangerous to do/say $a$''. The difference between the two is that non-social reasons contain natural actions while social reasons (i.e.~sanctions) do not. Following the scheme of Fig.~\ref{fig:agent}, both exert their influence on the focal actor's behavior via predictive pattern completion. There are sometimes more than one competing reason (social and non-social) both for and against doing/saying $a$. When  context contains multiple such reasons then the pattern completion network $p$ adjudicates between them (see Section~\ref{section:endogenousPreferenceFormation} for more discussion of how $p$ decides).

Other computational models have used classifiers to represent an actor's judgment of normativity \citep{hadfield2014microfoundations, koster2022spurious}. In particular, \cite{vinitsky2023learning} learned the classifier from sanctioning data. This is compatible with our view, since the model $p$ performs classification implicitly when doing inference based on memories $m$, which contain the memories of sanctioning events.

There is evidence that punitive emotions such as anger and outrage underpin the motivation to sanction in some (though likely not all) circumstances \citep{scherer1997role, fehr2002altruistic, haidt2003moral, sripada2006framework, seip2009hotheads, crockett2010impulsive, strobel2011beyond}. This is compatible with sanctioning behavior being conventional since, according to common models of emotion, both the emotion's releasing stimuli and the behavior it elicits are learned \citep{barrett2006solving} from others. Norm compliance may be motivationally underpinned by the learning of a ``sense of should'', experienced as anticipatory anxiety toward violating others’ expectations \citep{theriault2021sense}. The emotional state of the actor could be represented by a particular assembly $z^e$ in the global workspace $z$. Then using a specific emotional state as conditioning in the definitions \ref{def:contextual_convention_sensitive} and \ref{def:contextual_sanction_sensitivity}, would define an actor that behaves conventionally or normatively, given an emotional state.

\subsubsection{Norms are induced by generically conventional patterns of sanctioning}

Behavior is normative when it is encouraged (or its complement discouraged) by generically-scoped conventional patterns of sanctioning. This means that for there to be normative behavior in a particular situation, there must be a large group of people who do not know one another (generic scope) reproducing a consistent pattern of sanctioning by virtue of the weight of the its precedent (conventionality of sanctioning). This occurs, for instance, when everyone sharing a particular culture condemns the choice of a particular action $a$ in a particular context $c$.

The following definition formalizes this concept of normative behavior.

\begin{defn}[Normative behavior for the choice between two options]
The behavior of picking $a$ over $a'$ in context $c$ is \emph{normative} when at least one generic scope convention positively sanctions (encourages) picking $a$ over $a'$ and/or negatively sanctions (discourages) picking $a'$ over $a$.
\end{defn}

Any conventional pattern of sanctioning may encourage a behavior or discourage its complement. This also includes the narrow scope conventions that emerge in small groups of people who know each another well and apply only within their group. These differ in important ways from the generically scoped conventional patterns of sanctioning that underpin norms. However, both kinds of conventional sanctioning patterns can contribute to determining whether actions and utterances are or are not appropriate in their own scope, regardless of whether their scope is large or small (see Section~\ref{section:emotionsAndRelationships} for more on small-scope relationship-based appropriateness). The reason to reserve the term norm for behaviors suggested by generically scoped conventional patterns of sanctioning is that their small scope cousins are better understood as being driven by the specific interaction history and emotional states of the people involved. This preserves the standard connotation of ``norm'' as a constraint on the behavior of strangers interacting in society.

Notice that the properties of a behavior pattern as being a convention or being a norm are not mutually exclusive. For instance, it is both conventional and normative to drive on the correct side of the road. However, it is possible for a behavior to be normative but not conventional (e.g.~a functional behavior that is not reproduced solely on the basis of the weight of precedent like skillful woodworking, which would be encouraged by other people but also reproduced on the basis of physical criteria like stability). Likewise, it is possible for a behavior to be a convention but not a norm. Such behaviors are reproduced on the basis of precedent-sensitive reasons other than social sanctioning, like in the nuts and bolts example above.

Next we offer two conjectures on how populations featuring norms will behave over time and in the face of newly joining actors.

\begin{conj}[Norm stability] Populations of actors with established norms will tend to maintain them.
\end{conj}
Once a norm is established, it enjoys several self-reinforcing feedback loops. Sanctioning, in particular, reinforces the normative behaviour, as every sanction steers actors away from transgression and towards normativity. The sanctioning behaviour itself is reinforced by its conventional nature, as every sanctioning act reinforces the convention (creating more weight of precedent). From the perspective of an individual actor, observing sanctioning events makes an actor more likely both to comply with the norm and enforce it on others.

\begin{conj}[Norm adoption] New actors inserted into the population will adopt existing norms.
\end{conj}
A new actor, which joins a population with an established norm, will receive the pressure to conform from observing the sanctioning behaviour (even if not directed towards them). They will also, in time, accumulate the weight of precedent to reproduce the sanctioning behavior itself. The important assumption here is that the newcomer understands which actions are sanctioning, and has a pattern completion network $p$ close enough to that of the population.

\subsection{What counts as sanctioning?}
\label{section:whatCountsAsSanctioning}

Just as the particular actions and utterances that are sanctionable in a given context are conventionally determined (per group of individuals), it is also true that the particular actions and utterances that themselves constitute communication of approval and disapproval are conventional (e.g.~do you laugh nervously, complain vigorously, or gossip to third parties to convey your disapproval of another person's transgression?). Both the specific labeling of behaviors as sanctionable or not, as well as the actions selected to convey sanctioning itself are reproduced, and the reason they are reproduced is the weight of precedent. Therefore both are conventional (Section~\ref{section:conventions}). The conventions that determine which actions and utterances constitute sanctions (approval/disapproval) will generally have larger scope than the conventions that determine whether specific actions and utterances are themselves approved or disapproved. Like other components of ``normative infrastructure'' \citep{trivedi2024altared}, the mapping from utterances/actions to their interpretation as sanctioning is more generic than the content of specific norms.

In practice, the level of agreement as to which actions are sanctioning, who they sanction, for what behavior, and with what valence, seems impressively high. At least when the parties involved have a shared cultural background they do not seem to get confused about which actions are sanctioning or disagree about their valence or target very often. \cite{boehm1999hierarchy} retells an anecdote from \cite{briggs1971never} to illustrate in a memorable way how culture-specific sanctioning methods may fail to have the expected effect when applied cross culturally. In the story, Briggs, an anthropologist living with Inuit people in the far north of Canada, misunderstands repeated warnings against public displays of emotion, and her hosts are forced to apply successively stronger sanctions to convey to her the inappropriateness of her behavior. But most sanctioning is done within a single cultural context, so misunderstanding normally poses no obstacle.

Some sanctions, especially those which are relatively harsh, may actually be directed at specific individuals in a way that does resemble punishment for transgression. This happens much more frequently in ``tight'' societies than ``loose'' societies \citep{gelfand2011differences}. The loose societies may use relatively more indirect sanctioning methods instead. In fact, each society supports a wide range of negative sanctions which group members prefer to avoid, some may simply be withdrawal of benefits that would otherwise be available to community members. Most sanctions are socially constructed. For instance, loss of employment is a complex culture-specific sanction built on a particular institutional background. Overall, what constitutes sanctioning is culturally plastic.

Sanctioning may be centralized and formal (e.g.~coercion by the state) or decentralized and informal (e.g.~as in gossip). Other configurations are also possible. For instance, medieval Iceland had a system of formal but decentralized legal sanctions where the labeling of offenses and the determination of guilt or innocence was done through centralized courts but enforcement, e.g.~collecting fines, was outsourced to private citizens. The injured party (or their family) would normally have a right to collect the punitive fine from the crime's perpetrator, and if the victim was unable to enforce the claim themselves, then they could sell their collection right to someone else more able to do so \citep{hadfield2013law}. We will return to the medieval Iceland example in Section ~\ref{section:decentralizedSanctions}.

Informal sanctions involve censorious behaviors such as criticism, condemnation, avoidance, exclusion, or physical harm, as well as very light sanctions such as reminding someone of the proper way to behave \citep{sripada2006framework}. Sanctioning actions need not be dramatic. Indeed it is probably healthier for the community when a deep hierarchy of increasingly severe actions are available; graduated sanctions are one of Ostrom's ``design principles'' for successful common pool resource management \citep{ostrom2009understanding}. Sanctioning may also be mediated by technology. For instance while driving, one could honk their car's horn at another driver who has been behaving inappropriately\footnote{See \cite{vinitsky2023learning} for an extended discussion of sanctioning by car horn.}.

Sanctioning is often carried out by third parties \citep{fehr2004third, mathew2011punishment}, not individuals directly involved in the transgression itself, for instance via post hoc gossiping about the violation and the violator. Though individuals harmed by the norm violator may sometimes also be involved in enforcement or have a specifically prescribed role to play in enforcement, itself governed by other norms or laws (e.g.~the medieval Iceland example from \cite{hadfield2013law} which we discussed above). The fact that   sanctioning is normally done by third parties as opposed to victims means that individuals can also be sanctioned for transgressions without any clearly identifiable victim e.g.~transgressions against institutions, gods, or nature.

\cite{mathew2014cost} studied willingness to sanction individuals who violated norms demanding contribution to combat effort in the Turkana people, a politically decentralized nomadic pastoral group in northwest Kenya. The Turkana periodically launch raids to take cattle from other groups. The raiding parties are organized informally and participation is supported by norms. \cite{mathew2014cost} used a vignette-based study design where they told participants hypothetical stories about a warrior who failed to contribute to the combat effort and measured how often they endorsed various specific sanctions including `criticism', `not lending them an animal', and `not letting their daughter marry'. They found significantly higher rates of endorsing the sanctioning strategies when the vignette was about a coward versus when the vignette was about an unskilled warrior suggesting it is the norm violation of fleeing from battle which provokes sanctioning, not mere lack of skill.

While individuals may sometimes choose to sanction in order to teach their target how to behave, it may however be even more common to sanction for other reasons unrelated to any conscious or unconscious motivation to teach the target anything. One reason an individual might engage in such non-didactic sanctioning is to signal that they are a member of a group that follows the norm in question. Such sanctioning actions are like statements of social identity. Engaging in identity-affirming sanctioning may be a particularly strong CRED (credibility enhancing display) \citep{henrich2009evolution}, especially when it harms the sanctioner's relationship with their target (and their target's relatives and friends), making it a costly signal which one would not be likely to send unless they really were a member of the group holding to the norm in question. Regardless of why sanctioners choose to sanction, the didactic effect on others who learn what they discourage or encourage is the same.

Identity-affirming sanctioning is especially common on social media. Whenever a user publicly shares an outrage post circulating in their community which chastises someone for transgression, they affirm their own identity while sanctioning their target. Since motivation to affiliate with one's in-group is typically very strong, and social media recommender algorithms tend to prioritize the most engaging content, which tends to be the most outrageous, social media provides a constant stream of opportunities to engage in identity-affirming sanctioning which may be taken at low cost, simply by pressing the `share' button \citep{mcloughlin2024misinformation}. Unfortunately, the costs are often asymmetric, huge numbers of sanctioners may sometimes pile on to an issue, collectively delivering a massive punishment to their target, and creating for supporters the illusion of unanimity. Some individuals may even seek to further enhance the credibility of their group membership signal by taking their sanctioning offline, a phenomenon which has sometimes led to violence \citep{brill2024death, diresta2024invisible}.

We have argued here that for sanctioning to be effective there must be some level of collective agreement as to the social meanings of actions in terms of whether they are approval, disapproval, or are neutral, and, when they are approval/disapproval, to whom or what they are directed. While not universal, a combination of psychological, cultural, and institutional mechanisms work together to create a collective agreement on which actions count as sanctioning within a culture.

\subsection{Implicit vs explicit norms}
\label{section:implicitVsExplicit}

Recall that we defined a behavior as normative when it is encouraged by a generically conventional pattern of sanctioning (Section~\ref{section:normsDefinition}). We now divide conventional patterns of sanctioning into two categories: implicit and explicit on the basis of whether or not they may be exactly articulated in standardized language used by all individuals in the relevant community when determining norm-concordant versus discordant classifications. Here by \emph{relevant community} we mean the scope of the sanctioning convention defining the norm. Our theory posits different mechanisms of action for implicit and explicit norms so the distinction between them is of fundamental importance.

An example of an implicit norm, i.e.~a norm which is difficult to articulate in standard language is how close one ought to stand to a conversation partner~\citep{sussman1982influence}. Laws that prescribe or proscribe behavior are explicit norms since they can be precisely articulated in standard language shared by the relevant community. While one may sometimes be able to generate explanations of implicit norms e.g.~``I think one should stand one third of a meter away from friends when talking to them''. Even within a single community, answers vary to the question of ``how far should you stand from a friend while talking to them?''. Some may describe the norm with a gesture instead of words. Some may give different distances, etc. There is no standard language to describe a conversation distance norm. The category of implicit norms also includes many which cannot be articulated at all in a consistent fashion such as \cite{haidt2001emotional}'s moral dumbfounding examples, of which we will have more to say in Section~\ref{section:implicitNorms}. 

The category of explicit norms includes both laws and proverbs. Laws are formulated explicitly in fixed and standard language that constitutes the law and provides at least one relevant way to classify violation versus non-violation. It does not matter to the classification that some laws are ambiguous. All that matters is that the law is precisely articulable in standard language used by all in the relevant community. Proverbs are also explicit norms such as ``action's speak louder than words'', ``better late than never'', ``beggars can't be choosers'', and ``a bird in hand is better than two in the bush''. To work out whether a particular behavior violates a norm represented by a proverb it is usually necessary to employ analogy and pseudo-deductive thinking. Both proverbs and laws may be cited as the reason for applying a sanction. For both laws and proverbs the standardized nature of the norm's representation is sufficient to support pseudo-deductive natural language inference using it, as in courts applying logical consequences of laws and precedents to new cases, or individuals deciding the implications of a proverb for a recurring situation in their life. 

Many norms are supported both by implicit and explicit patterns of sanctioning. For instance, courts generally classify instances of stealing behavior as wrong and they do so via explicit reasoning using a law and the facts known to the court. Additionally, bystanders who observe someone snatch someone else's purse may also consider the snatcher's behavior to be wrong and may attempt to stop them or try to help the victim in some other way such as by calling the police. In this example explicit and implicit patterns of sanctioning point to the same classification: the observed instance of stealing is wrong. 

There are also situations where implicit and explicit patterns of sanctioning conflict with each another. For instance, on a particular road, there may be a posted sign stating a legal speed limit of 65 MPH. However, simultaneously there could also be an implicit norm suggesting that on this particular road everyone should drive 75 MPH. Individuals driving there face a choice of either driving at 65 MPH in accord with the law or driving at 75 MPH in accord with the implicit norm. They integrate the two ``reasons for action'' into their decision in the way described in Section~\ref{section:endogenousPreferenceFormation}, as a social choice over the various elements of their guidance. There is no need to decide the same way every day. Maybe on one day, I let my identity as a law abiding citizen dominate, so I drive at 65 MPH. Then on another day I let my identity as rushed and not-very-risk-averse person dominate so I drive at 75 MPH, or I feel rushed because someone honks at me for driving 65 MPH, so I switch to 75 MPH. The 65 MPH outcome is supported by the explicitly articulable pattern of sanctions and the 75 MPH outcome is supported in a more implicit way. But both patterns of support can coexist as compelling reasons to act one way or another.

Numerous alternative classification schemes for types of norms exist. In particular, some readers will find it helpful at this stage for us to point out that our concept of norm encompasses both ``injunctive norms'' and ``prescriptive norms'' (see \cite{cialdini1991focus, chung2016social}).

We will propose that the mechanism underpinning the difference between implicit and explicit norms is that explicit norms act via pattern completion from prefix to suffix in the global workspace while implicit norms affect behavior because they are already embedded in the prediction network used for the completion.

\subsubsection{Implicit norms}
\label{section:implicitNorms}

Implicit norms are norms that cannot be articulated verbally in a precise way. They reflect an overall tacit consensus labeling behavior as acceptable or not in a given context. For instance, appropriate conversational distance varies with culture. For instance \cite{sussman1982influence} found that Japanese people sit farther apart from one another while conversing than Venezuelans do, and that Americans sit at an intermediate distance. \cite{hertz2024cognitive} gives an example of restaurant tipping. While it's clear that, in many cultures, giving no tip is unacceptable, exactly how much to tip is not precisely defined by any specific and clearly defined standard that all could agree on. Rather, one learns how much is appropriate to tip as a function of many factors like the kind of restaurant, the quality of the service, the location, etc. Those of us living in tipping cultures can classify the appropriateness of tip amounts but would be hard pressed to articulate a specific standard.

Implicit norms are applied via recognition not reasoning. Obscenity is famously so difficult to precisely define and reason about that US Supreme Court Justice Potter Stewart could offer only the classification ``I know it when I see it''. This is because whether or not something is obscene is controlled mostly by implicit norms which are, by their nature, hard to articulate.

\cite{haidt2001emotional} discusses many examples of ``moral dumbfounding'' where an individual may feel strongly that a particular behavior is wrong but be unable to articulate precisely why they feel that way. These examples include ``eating one's (already dead) pet dog, or cleaning one's toilet (in private) with the national flag'' \citep{haidt1993affect}. Not every individual is dumbfounded by the same examples. When an individual is morally dumbfounded, they initially search for reasons, but rapidly drop them and try others when reminded of conflicts with the premise e.g.~``the dog was fully cooked so there's no chance of getting sick''. Eventually they give up, then ``stutter, laugh, and express surprise at their inability to find supporting reasons, yet do not change their initial judgments of condemnation'' \citep{haidt2000moral}. We argue in this paper that Haidt's dumbfounded participants were unable to produce reasons to justify their moral judgments because their decisions were guided in these instances by implicit norms that conflict with explicit norms they also endorse such as ``actions that are neither harmful nor unfair are permissible''. We argue in Section~\ref{section:moralization} that when implicit and explicit norms come into conflict, and there is no chance of being sanctioned, then it is the implicit norms that usually end up taking precedence over the explicit (if the action is not private then the likelihood and severity of social sanctioning may differ between the competing implicit and explicit norms so individuals would generally be guided by whichever they perceive as more helpful for avoiding disapproval or seeking approval). So in the moral dumbfounding examples (which all concern private activities), the participants are left with a strongly felt judgment of moral wrongness based on an implicit norm they cannot precisely articulate, and no consistent explicit norms to fall back on when asked to articulate a reason to support their view.

\subsubsection{Explicit norms}
\label{section:explicitNorms}

Explicit norms are norms that can be explicitly articulated in natural language as rules. Formal laws are a kind of explicit norm. Other explicit norms, which we call \textit{proverbs}, are rule-like but less formal than laws.

When there are no implicit norms to govern behavior in a particular domain then explicit norms can be especially powerful. For example, consider the experience of playing a new card game for the first time after being told the rules. Since you have not played the game before there is no way that you could have already built up implicit norms relating to appropriate ways of playing this game, though you might fall back on implicit norms relating to similar games that strategy won't work if the game is sufficiently different to others you have played. In fact, the only way to play the game appropriately the first time is to allow your behavior to be guided by the explicit norm communicated to you by the person who told you the rules. This is what it means to be guided by an explicit norm.

We should not expect explicit moral principles to be able to accomplish what custom, education, and tradition cannot. The force of the better argument is generally not likely to win over the greatly entrenched mass of schematically integrated (consolidated) implicit norms. You cannot simply argue someone into changing how they live \citep{haidt2001emotional, rorty2021pragmatism}. Nevertheless, there are situations in which explicit norms, mainly formal laws, do override implicit norms, especially in public and when centralized sanctioning is possible. For instance a new law against discriminatory speech could render a particular utterance inappropriate regardless of the implicit norms governing the context. These situations may be transitory, as time passes in this state we might expect the implicit norm to change in order to ``follow'' the law. In this case, people may initially think the law misguided but come to endorse it later on \citep{sunstein2019change}. Initially they may only follow the law when they are in the presence of potential enforcers, and revert to following their preexisting implicit norm whenever they are alone. But after enough repetition, the explicit norm may consolidate into cortex, and afterwards they may come to follow the law even when they are alone since consolidation changes the next-symbol predictor $p$ to implicitly represent a compressed version of the previously explicit norm (see Section~\ref{section:novelBioInterpretation}), and implicit norms usually take precedence over explicit norms (see Section~\ref{section:moralization}).

Not all explicit norms are laws. There is another category, which we call proverbs. Proverbs are explicitly articulable rules to guide behavior such as ``do unto others as you would have them do unto you'', ``an eye for an eye'', and ``don't cry wolf''. Many proverbs have moral content but some do not, e.g. ``don't wear white after labor day''. Proverbs differ from laws since they have no institutional existence. They are no courts to classify their violations. However, proverbs also differ from implicit norms in that they can be articulated in sentence form and used within logical chains of deductive reasoning. They are often hard to apply directly to classify social situations though, and thus involve considerable ambiguity of application. Nevertheless they remain helpful in coordinating conversations on values since they are so widely known, easy to articulate, useful for priming others' thinking to progress in a particular direction, and for establishing common ground.

Proverbs are relatively impoverished in their effectiveness as a determiner of what is appropriate. In this they differ from implicit norms and also from the other category of explicit norms (laws). We call this the ``poverty of proverbs''. Unlike laws, proverbs depend too much on their status as good memes \citep{dawkins1976selfish}. Since proverbs spread mainly by oral tradition they must be easy to learn, remember, and transmit. These requirements cut against their effectiveness as standards for appropriate behavior since their low-information content renders them unable to account for situations that call for complexity. A rule complex enough to be useful in the real world will generally have to contain more information, and thus be harder to remember, than an overgeneralization that fits some easy-to-recall examples while failing in less typical settings. For instance, it's of course easy to remember a proverb like ``do unto others as you would have them do unto you'', but to coordinate third-party punishment of violators it would be better to use specific laws (and legal adjudication procedures) which are typically written down, and are even composed in specialized formal language to increase precision and minimize the risk of conflicting interpretations\footnote{\cite{santos2018social} and \cite{yaman2023emergence} may be seen as arguing against the `poverty of proverbs'. They both find that simpler norms were more effective at promoting cooperation in abstracted matrix game settings. However, their results were limited in applicability by the simplicity of the environments used. There is considerable evidence that relatively information-rich rules---tied to specific resource dynamics---are needed to manage social-ecological systems like fisheries, forests, groundwater, and urban car parking space (e.g.~\cite{ostrom1990governing, ostrom2009understanding, baggio2016explaining}).}.

When the goal is social change, explicit norms, and laws in particular, are often better intervention targets than the other factors influencing appropriateness since laws may be changed more rapidly, at least in principle \citep{nyborg2016social, prentice2020engineering}. Such strategic social construction of the normative environment is not at all uncommon \citep{finnemore1998international}.

\subsubsection{How the brain represents implicit and explicit norms}
\label{section:representingImplicitVsExplicit}

Where are the explicit norms stored? One bad way to maintain an explicit norm would be to continually rehearse it using working memory, but that would be extremely effortful, and would impair your ability to perform simultaneous tasks \citep{vallar1984fractionation}. So instead of continually rehearsing explicit norms, we may instead assume that they are stored in hippocampus, at least for some period of time anyway. The hippocampus is critical for exception learning, where a general rule must be retained while building more nuanced understandings of how the rule needs to be used \citep{heffernan2021learning}. Over time, these new or more complex rules become integrated with more associative memory in the neocortex and the hippocampal representations are no longer needed \citep{squire2015memory}, at which point we posit they are transformed into implicit norms. 

As a result of being stored in a natural language representation, explicit norms are somewhat fragile. They may be hard to recall, and frequently forgotten if not used. They are less integrated into other knowledge schemas than implicit norms are, so incoming stimuli activate them less often. When an explicit norm is not practiced often enough there is a risk of it being forgotten. Luckily, explicit norms may also be written down on paper (or stone tablets), and kept permanently \citep{norenzayan2016cultural}.

Consolidation refers to the process by which a temporary memory, initially stored in the hippocampus, is transformed to a more stable format and transferred to neocortex, eventually becoming independent of the hippocampus entirely \citep{squire2015memory}. Mirroring the mechanisms by which other kinds of information are consolidated, \cite{peyrache2009replay} showed that there are transient bursts of hippocampal reactivation during slow-wave sleep of activity patterns associated with the behavior of learning to follow a newly introduced rule. Their result suggests that during slow-wave sleep the hippocampus replays neural patterns associated with recently acquired information so that neocortex, which requires more repetition to learn, has an opportunity to absorb the new information \citep{mcclelland1995there}. 

Explicit norms are stored in hippocampus and used as conditioning info for the generative model $p$ sampled to produce behavior by pattern completion. Explicit norms may over time become consolidated into implicit norms i.e.~assimilated into the weights of the neural network). Implicit norms on the other hand are probably stored in prefrontal cortex (PFC). In a task involving switching which of several learned rules to follow, neurons in PFC encode which abstract rule is in effect at any given time \citep{wallis2001single}. 
\subsubsection{How the brain uses and updates norms}
\label{section:UsingUpdatingNorms}

Neural systems have evolved to support norm enforcement by facilitating appropriate behavioral responses to the threat of social punishment \citep{montague2007detect}. Unfair offers in economic exchanges, such as those encountered in the Ultimatum Game, activate brain regions associated with emotional processing. For example, \cite{sanfey2003neural} demonstrated that unfair offers elicit increased activity in the anterior insula, a region linked to negative emotions like disgust. This activation reflects a strong aversion to unfairness, motivating individuals to reject such offers even at a cost to themselves. Areas of the dorsolateral prefrontal cortex (DLPFC) are also implicated in distinguishing between different types of transgressions, such as those for which a person has responsibility versus those they do not \citep{buckholtz2008neural}. This suggests that the DLPFC supports the cognitive evaluations needed to interpret and respond to social violations. In addition to negative affect, punishment of norm violations engages both reward-related and aggression-related neural circuits, highlighting the dual role of emotion and cognition in norm enforcement. Linking a computational model to neural data, \cite{chang2013great} demonstrated that violations of normative expectations elicit prediction errors in the anterior cingulate cortex (ACC), prompting behavioral adjustments in punishment. 

Punishing norm violations engages reward-related neural circuits. \cite{dequervain2004neural} used fMRI to show that the dorsal striatum (specifically the caudate nucleus) is activated when individuals engage in altruistic punishment of unfair behavior. This activation correlates with the perceived satisfaction from punishing unfairness, suggesting that enforcing norms can be intrinsically rewarding. Interestingly, there is a common set of neural regions, including the ventromedial prefrontal cortex that are active for both moral/harm based and convention based transgressions, suggesting a common reward pathway for both types of punishment \citep{white2017neural}. Interestingly, serotonin levels, a neuromodulator associated with negative affect and avoidance at low levels, modulates these responses, with increased serotonin levels enhancing the striatal response to fair offers and reducing retaliatory behavior in response to unfairness \citep{crockett2013serotonin}. This may suggest that although perceived unfairness triggers negative affect, the increased serotonin may reduce the additional perceived cost of retaliation.

Given that people readily punish transgressions, including conventional transgressions, the threat of punishment plays a crucial role in maintaining norm compliance \citep{fehr2002altruistic}. Individuals are highly responsive to potential sanctions, as reflected in increased lateral PFC activation when facing the possibility of punishment \citep{spitzer2007neural}. Demonstrating a causal role for the DLPFC in normative behavior, \cite{ruff2013changing} showed that direct brain stimulation to this region altered participants' sensitivity to potential punishments and their subsequent compliance with norms. This suggests that the DLPFC is not only involved in norm enforcement but also critical for regulating behaviors under the threat of punishment.

Interestingly, normative behavior can become habitual and automatic through repeated exposure and practice. Early reliance on explicit, effortful deliberation can give way to implicit, automatized adherence to norms, reducing cognitive load. People are extremely sensitive to social feedback, with areas such as the ventral striatum tracking social approval in a manner similar to how other rewards are processed \citep{izuma2008processing}. Dishonest responses, as opposed to honest ones, are associated with greater DLPFC activation, suggesting that honesty may be the default behavior, even in situations where dishonesty would result in greater monetary benefit \citep{greene2009patterns}. 
Demonstrating the role of the stratum and VMPFC in learning norms, \cite{xiang2013computational} had participants play an ultimatum game where participants could learn the distributions of typical sharing distributions. Critically, there were three distributions that shifted across the the study, and participants altered their sharing behaviour to match the norms. These shifts in behaviour to match norms were associated with activation in the stratum and the VMPFC.  Relatedly, \cite{nook2015social} found that the vmPFC initially responds more to unhealthy foods, but when group norms were introduced favoring healthy eating, vmPFC responses shifted to track food popularity. Interestingly, the vmPFC encoded popularity rather than inherent healthfulness, suggesting that norms can dynamically reshape valuation processes.

Patients with PFC lesions often fail to adapt appropriately as rules change over time \citep{miller2001integrative} and the famous patient Phineas Gage was described after the railroad spike accident which damaged his orbitofrontal cortex as having become prone to engaging in inappropriate social behavior~\citep{harlow1848passage}. Modern accounts of orbitofrontal damage also stress the difficulties these patients have with selecting social behaviors appropriate for interactions with strangers versus those appropriate for those with whom they have close relationships \citep{beer2003regulatory}. For instance, \citep{rolls1994emotion} described several orbitofrontal patients who made sexually explicit comments to research staff, one who ``tactlessly told a member of staff she was much less pretty than someone else'' and one who ``swept a member of staff off her feet to hug and kiss her''. This pattern of behavior change after prefrontal cortex damage indicates loss of implicit norms, not the loss of the ability to be guided by explicit norms. It's some of the most deeply automatized and internalized norms that are lost in these cases, e.g. don't make sexual advances to strangers. The kind of norms lost in prefrontal damage is the kind that normally guide behavior automatically. It does not normally require rehearsal or effortful recall to maintain their guidance, and patients with prefrontal damage can still behave themselves but it requires effort and is taxing of their working memory, and frequently fails (as all explicit norm guidance does). As for implicit norms, we can hypothesize that they are stored in neocortex, specifically orbitrofrontal cortex since this region is associated with abstract rule-following/rule-switching behavior in monkeys \citep{wallis2001single} and regulating context-dependent social behaviors in humans \citep{rolls1994emotion, beer2003regulatory}.

Finally, psychopaths, who exhibit abnormal social behavior, show deficits in norm processing. For instance, they fail to differentiate between moral and non-moral stories in the vmPFC \citep{harenski2010psychopaths}. This impairment underscores the role of the vmPFC in integrating social and moral information to guide behavior.

\subsection{Sanctioning associated with laws may be either centralized or decentralized}
\label{section:decentralizedSanctions}

In the schema we inherit from \cite{hart1961concept, hadfield2012law}, legal norms are distinguished from other kinds of explicit (and also implicit) norms by their capacity for deliberate change. Legal systems include ways by which new laws can be approved and existing laws modified. These institutional affordances by which the law evolves include the traditional lawmaking done by legislators, as well as precedent-setting decisions by judges, and rule-making by regulatory bodies. Under this definition, legal rule-making also occurs in less obvious settings such as corporate human resources departments and ethics boards, trade associations, professional bodies like bar associations (which define rules of membership  \citep{hadfield2017rules}), and in small groups of motivated people concerned with managing common resources \citep{ostrom1990governing} such as common fisheries \citep{acheson2003capturing}, or mining rights during the 1848 California gold rush \citep{hadfield2013law}, and trade relationships during the late medieval commercial revolution period \citep{greif1994coordination}.

Under this definition of law whether enforcement is centralized or decentralized is not a critical factor. It is possible to have a legal system without any centralized enforcement. In such cases a mechanism is needed to make decentralized sanctioning incentive compatible. Medieval Iceland offers one example of how this can be achieved \citep{hadfield2013law}. The Icelandic legal system during this period lacked a central executive authority to enforce laws. Instead, enforcement relied on a complex system of private prosecution and decentralized sanctions, overseen by the ``law speaker''. The law speaker was a prominent figure elected to memorize and recite the entirety of the law, as well as to answer questions from litigants to clarify the law; and to be the only person recognized as having the power to declare with finality the content of the law in the event of disagreement. Importantly, the law speaker did not oversee a state-run enforcement mechanism. Instead, victims of crimes were granted an exclusive private right to prosecute, and receive monetary compensation from, their wrongdoer. Such a ``property'' right in prosecution, could lead to an efficient deterrence system since the holder of the prosecution right could contract with private mercenaries to ensure payment and mercenaries could advertise their customer list to deter would-be criminals from targeting their customers \citep{hadfield2013law}.

Institutions are composite entities consisting of a set of intertwined legal and non-legal norms, including many implicit norms. \cite{hart1961concept} argues that legal systems are ultimately grounded in ``rules of recognition'' that are not themselves justified in the normal way of legal justification but rather instead function more like proverbs or implicit norms setting out standards for official behavior. Rules of recognition determine which other rules count as part of the legal system and which do not (see also \cite{ostrom2009understanding}'s related concept of ``constitutional'' rules). \cite{hart1961concept}'s classic example is the proverb-like rule accepted by officials in the UK that ``what the Queen in Parliament enacts is law'', which he argues is accompanied by elite willingness to socially punish any official who deviates and to internally punish one's own lapses. Interestingly an empirical study of British officials found considerable variation in officials' acceptance of the proverb, especially in connection with constitutional edge cases \citep{howarth2018hla}. Individuals in their study were found to apply more flexible standards than the proverb could capture (e.g.~some officials felt bound by parts of international law regardless of what parliament may say). We can interpret this result as further evidence for the poverty of proverbs. It is likely that the UK's constitution depends to a greater extent on harder-to-formalize implicit norms than it does on the specific proverb Hart identified. In fact one of \cite{howarth2018hla}'s participants said this clearly, remarking that ``people simply talk to one another to ascertain the law''.

\subsection{Moralization}
\label{section:moralization}

Some norms are associated with moralized language while other norms are not. The function of morality talk is sanctioning. We may call it ``moralization''. When you announce ``$X$ is moral'' you assert that the community ``approves of $X$ and disapproves of not $X$'', and this has a sanctioning effect in that it causes listeners to be more likely to engage in $X$. One argues that a particular norm is moral in order to 1) claim the norm is very important and thus ought to take precedence over other norms. And 2) claim the norm has a wide or universal scope of applicability. The sanctioning effect of moralizing speech may be indirect. Morality talk also occurs frequently in gossip and gossip is generally a key distribution channel for normative information of all kinds.

While both implicit and explicit norms may be moralized, it is likely that implicit moral norms would usually take precedence over explicit moral norms when they come into conflict. When two norms come into conflict, the more consolidated (i.e.~implicit) norm will usually be the one to control behavior, e.g.~one may feel they must judge the behavior of the characters in \cite{haidt2001emotional}'s incest vignette as wrong even though there are alternative explicit norm-guided reasons to do otherwise \citep{schwartz1977normative}.

In the case of explicit moral norms, sometimes additional considerations activated by the context may neutralize the sense of moral obligation \citep{schwartz1977normative}. For instance, a person may initially feel obligated to donate money toward alleviating poverty in a far away place, but then after considering some rationale for why doing so is not their responsibility (e.g.~``others who are much wealthier do not donate'') conclude they have no such obligation. In the case of implicit moral norms, such neutralization via rationalization is much less likely. In general, implicit moral norms, which cannot themselves be articulated in language, are mostly recalcitrant to verbally articulated reasons to disregard them. The cases of moral dumbfounding described by \cite{haidt2001emotional} show the weak level of influence that explicit normative language may have on the sense of moral obligation arising (in private) from a well-consolidated implicit norm. The reason may simply be that it is hard for verbal information to interact with non-verbal information. 

Notice however that implicit norms only override explicit norms when guiding actions taken in private. Otherwise, the severity and certainty of sanctions would usually differ between the competing implicit and explicit norms. For explicit norms that are formal laws backed by a state enforcer, the severity and certainty of sanctioning is sufficient reason for the explicit norm to take precedence over the implicit. In these cases, one may choose to comply with the law despite ``feeling like'' the action it requires of them is inappropriate \citep{sunstein2019change}.

As a result of their precedence over other norms, individuals come to understand the most consolidated and implicit moral norms that guide their behavior as being unconditional. Whereas they view the other norms, which also guide their behavior, as being conditional on the expectations of others \citep{turiel1983development, bicchieri2005grammar, mao2023doing}.

\cite{kohlberg1983moral, may2018regard} argue that personal moral reasoning is critical for making moral judgment. However, such reasoning-based accounts have a lot of trouble accounting for the data concerning how quickly people make moral decisions and how conformist they are (e.g.~\cite{singer2008understanding}). Another reason to think norm guidance may not depend on universalizing moral reasoning is that norms are often parochial, privileging some groups above others. For instance a group's members may view stealing from in-group members as immoral while simultaneously holding that stealing from an out-group is permissible \citep{partington2023rational}. In this sense many moralized norms may not appear moral when viewed from the perspective of an external observer from a different society \citep{machery2022moral}.

While most norms contain largely arbitrary content. There may be some deeply non-arbitrary norms (or meta-norms). Perhaps there are universal moral truths that all cultures would have to recognize in order to be effective and thus to survive under (cultural) group selection. Perhaps such universal truths include prescriptions to minimize harm or maximize equality among group members as suggested by \cite{turiel1983development}. It's plausible that harm-minimizing and equality-maximizing norms promote societal stability, and that it would be hard to maintain a society without them. For our purposes in this paper we need not take a stance on whether such norms are truly as timeless as they subjectively seem to be, nor whether particular norms would look more effective from the perspective of an external observer. Here it's enough merely to recognize that at least some large part of appropriateness consists of relatively arbitrary factors like the current culturally prevailing usage of particular words.

\subsection{Norm change}
\label{section:normChange}

Norms are socially constructed---sometimes accidentally through evolutionary drift-like processes \citep{young1993evolutionary} and sometimes strategically \citep{finnemore1998international}. Since the concept of appropriateness is associated with the status quo, one may think appropriateness is an inherently conservative force. However, as pointed out by \cite{march2011logic}, demands for reform and redistribution of political and economic power also follow from identity-driven conceptions of appropriateness at least as much as they do from rational calculation of cost and benefit. There are situations involving conflict between groups of individuals who would prefer different norms \citep{stastny2021normative, mukobi2023welfare, vinitsky2023learning}. In these cases the prevailing norm may shift along with the distribution of power and population between the groups \citep{young2015evolution, koster2020model} or through the dynamics of conflict resolution \citep{rahwan2018society, sunstein2019change, noblit2023normative}.

\cite{heyes2022rethinking} characterizes the way explicit norms evolve over time as being driven by a process of norm commentary. In her view, individuals are motivated to talk about their explicitly normative classifications in terms of rules and deductions around them. Commentary facilitates social learning since others can easily learn the details of how others are applying the rules in particular situations even if they do not experience those situations themselves, access to commentary may be enough on its own for normative judgments to diffuse through a culture.

Laws, regulations, and precedent-setting court decisions are explicit norms which are intimately tied to institutions. These explicit norms are characterized by the deliberately-planned process by which they change \citep{hadfield2014microfoundations}. When legislators perceive a need to change a law they can pass new legislation to do so. \cite{hart1961concept} distinguishes between primary and secondary rules. The former directly govern behavior whereas the latter confer various statuses (e.g.~legislator, judge, quorum, etc). Most importantly, secondary rules determine the conditions under which other rules become valid. Secondary rules allow this class of explicit norms to change rapidly and deliberately.

There are no secondary rules to facilitate change for the other main type of explicit norm: proverbs. So they rarely change. Proverbs change through memetic evolution, a process driven more by a proverb's memorability than its usefulness.

Implicit norms cannot be changed as deliberately as explicit laws (and other official norms). People learn what (implicit-)norm-concordant behavior looks like from observing what others do, and what others sanction. Normative behavior is induced by conventional patterns of sanctioning so (by definition) the weight of precedent is enough to motivate individuals to reproduce their part of the pattern of sanctions. When they sanction the same behavior they think others would sanction this generates a bandwagon effect, the more a behavior gets sanctioned the more data there is in the community's overall gossip from which one may learn that it is sanctionable. This in turn leads more individuals to view the behavior as sanctionable and thus to sanction it further. \cite{vinitsky2023learning} studied a learning model of the bandwagon effect along these lines. Sometimes norms change quickly in response to external events, e.g.~the norm of mask-wearing which rapidly took hold during the COVID-19 pandemic \citep{yang2022sociocultural}, but more often norms features substantial inertia and are very difficult to deliberately change \citep{bicchieri2016norms}.

Not only does the specific content of norms evolve but the cognitive and emotional features that make norms possible in the first place by promoting sanctioning also evolve \citep{heyes2022rethinking}. The term \textit{norm psychology} is used to refer to the suite of cognitive adaptations that make human normative behavior possible \citep{henrich2017secret}. Norm psychology also evolves, and its evolution isn't just a product of the distant past. Changes in norm psychology may happen rapidly in response to changes in environment. In particular, social media has created a large change in relevant aspects of the environment as experienced by many people around the world today \citep{brady2023norm}. There is more than one mechanism by which social media may affect norm psychology. Individuals may form habits while interacting with social media. The social feedback these platforms provide in the forms of numbers of likes and re-posts are a convenient signal of social approval, and recognized as reinforcing by the brain. Beyond the effect this has on the content of norms itself, it may also affect more general aspects of norm psychology or social decision making. For instance, \cite{brady2021social} showed on a digital social network that the amount of positive social feedback (likes, re-posts) a user receives for their sanctioning posts on one day is positively associated with the number of sanctioning posts they produce in subsequent days. This suggests that social media encourages sanctioning, perhaps making users more sensitive to releasing stimuli that promote sanctioning behavior.

Norm change may also occur due to cohort replacement (replacement of an older birth cohort that has one set of views by a younger birth cohort with other views) \citep{firebaugh1988trends}. For instance, norm change may sometimes be driven by individuals comparing themself to what is perceived as being the average beliefs of others \citep{crandall2018changing}, and a shifting average. One mechanism by which conventions and norms may change rapidly operates when there are sizable populations unhappy with the status quo who nevertheless do not talk about their preference or try to change anything because they perceive the costs of doing so to be too great. Once something changes to show them that those costs no longer exist, or are rapidly shrinking (perhaps because the pool of people remaining to disapprove is shrinking) then they may quickly change to support their true preference. This is a positive feedback process which gathers momentum as it progresses since the more people there are who already abandoned the old norm the easier it is for others to join them.

Overall, the social forces that bring about norm change are very powerful. While some norms may appear deeply ingrained and resistant to change, history shows that even the most entrenched can be transformed through social movements, legal reforms, or shifts in cultural attitudes. Sometimes change is gradual, emerging through intergenerational shifts or the slow accumulation of minor variations in behavior. At other times, change can be swift and dramatic, driven by shifting power dynamics, the cascading effects of social sanctions, or the sudden reduction in the perceived cost of challenging the status quo. Even the way we think and feel about norms, our "norm psychology," is subject to transformation, influenced by factors such as the rise of new communication technologies and changing social structures. The dynamism of norms highlights the fluid and adaptable nature of human societies.

\subsection{Norms as technologies}
\label{section:gadgets}

Norms are often well-fit to the demands of the local environment \citep{baggio2016explaining, talhelm2020historically, uskul2008ecocultural}, suggesting that they may emerge by virtue of their capacity to serve some subsistence function or resolve some recurring social problem. From this perspective, it makes sense to think of norms as technologies. Like other technologies, they are invented to fill functional roles, but within constraints, can still evolve in the cultural fashion we have described throughout this article, e.g.~there is considerable room for context and culture dependence, arbitrariness, automaticity, dynamism, and sanctions. In this section we first describe how norms can be viewed as resolutions to ``how can we live together?'' questions posed by the physical and social environment. Then we turn to one of the most important kind of norm: epistemic norms. Epistemic norms are foundational to science and other kinds of human inquiry. Finally, we discuss what this view of epistemic norms as culturally contingent inventions means for ideas of justification and rationality.

\subsubsection{Norms can resolve questions of ``how can we live together?''}

It is easy for a goal-seeking actor to create problems for others by selfishly choosing actions with side-effects that negatively impact others  (e.g.~\cite{ostrom1994rules, perolat2017multi}). Norms are sometimes used to coordinate actions and motivations in order to navigate these situations where gains from cooperation are possible but egoism makes cooperation hard to achieve. These functional norms, which promote cooperation, are like technologies for resolving social dilemmas and other fraught situations that arise when actors have interdependent goals. In some cases, cooperation-eliciting norms evolve through undirected cultural evolution, perhaps aided by group selection \citep{wilson2013generalizing}. In other cases, functional norms are sometimes deliberately designed and disseminated. For instance there have been coordinated efforts to instill norms that proscribe littering and smoking cigarettes near public places \citep{nyborg2016social}, as well as efforts to change harmful norms like those contributing to child marriage \citep{bicchieri2016norms}. 

Functional norms emerge in numerous domains of human activity, from small-scale social interactions to complex economic systems, e.g. laws and shared notions of appropriateness make complex market interactions possible \citep{north1990institutions, greif2006institutions}. Much like how norms can resolve social dilemmas in other domains, legal frameworks and commercial customs enable reliable exchange and indirect cooperation between actors who may never meet \citep{williamson1985economic, bowles2004microeconomics}. Property rights, contract enforcement, and standardized rules of commerce create an infrastructure for complex economic collaboration---as illustrated in Leonard Read's ``I, Pencil'' essay \citep{read1958i}, where thousands of people across the world, from cedar farmers to graphite miners all coordinate their efforts through market mechanisms, and unwittingly collaborate to make pencils. This exemplifies how appropriateness norms can evolve into sophisticated technologies for organizing human behavior at scale \citep{hayek1945use}.

\subsubsection{Epistemic norms and science}\label{section:epistemicNorms}

``Since the Enlightenment, and in particular since Kant, the physical sciences had been viewed as a paradigm of knowledge, to which the rest of culture had to measure up. Kuhn's lessons from the history of science suggested that controversy within the physical sciences was rather more like ordinary conversation (on the blameworthiness of an action, the qualifications of an officeseeker, the value of a poem, the desirability of legislation) than the Enlightenment had suggested.''---\cite{rorty1978philosophy} (pg. 322), discussing \cite{kuhn1997structure}

The norms of scientific discourse include what constitutes a valid argument, what reasonable hypotheses look like, and what counts as good evidence. These norms emerge from the specific cultural and historical context of the scientific community and differ in their details from subfield to subfield. These evolving norms are deeply ingrained in the culture and practice of scientific communities. They shape research agendas and training methods, as well as what get published. How do new subfields form? There may be many ways, but a common one is likely something like: a group of researchers who are already members of the broader disciplinary community, and who may often be independently friendly with one another, ascertain in their conversations a certain commonality in methodology and objective e.g.~we are a group who want to use $X$ method to achieve $Y$ change in theoretical paradigm\footnote{For instance, this could be ``we are a group who want, in connection with older Kantian traditions on innateness, to apply the methods of Bayesian statistics to rehabilitate older ideas attempted in symbolic cognitive science and thereby push back against an onslaught of connectionism''}, and perceive their ``vector'' to be sufficiently different from others as to have an independent identity, but not so different as to be unacceptable in the discipline more broadly. Then the group works for some time in the direction indicated this way, building up a body of work which pushes in the same direction, in particular determining how to judge contributions within the nascent paradigm (norms of reasoning), and training students to think in their terms, thereby growing the group and perpetuating the paradigm. If the paradigm also comes to be seen by outsiders as successful on their own terms then others may ``convert'' and join it.

Society privileges the knowledge acquired by certain favored epistemic norms by labeling it ``objective'', and consequently deprioritizes information which people come to believe through unapproved methods. At one moment in time the only acceptable way of knowing a proposition could be to discover it in scripture, while at another moment in time, the only acceptable way of knowing could be via application of the norms of scientific, journalistic, or other professional research \citep{berger1966social}. In this light, it is interesting that a recent study of norm-elicitation for opinions of diverse groups on how LLM chatbots should speak about sensitive topics shows that many groups agreed that ``factuality'' should be prioritized in the way chatbots compose their responses \citep{bergman2024stela}. This result may makes it look like there was greater consensus between the groups than there was in reality. Since presumably, the different focus groups would not all have agreed with each other on which particular statements should have the status of objective fact (e.g.~who won the US election in 2020?).

When a community has a particular goal, the usefulness of different norms they may hold can be graded by their usefulness in helping achieve that goal. For example, there are communities of people who are interested in employing their human individual intelligence to make predictions about future events, e.g.~metaculus, super-forecasters, etc. These communities discovered that forecasting is in fact a learnable skill \citep{tetlock2016superforecasting}, and one's accuracy can improve considerably by practicing certain techniques, most of which function to diminish the effect of cognitive biases which otherwise would dominate prediction decisions. These techniques acknowledge human fallibility, incentivize rigorous evidence integration (e.g.~by quantifying beliefs with probabilities), and foster a culture where the ability to revise beliefs in light of new data is encouraged. These practices affect how knowledge is obtained, refined, and evaluated within the forecasting community. They are normative not just in the sense that they promote the making of higher accuracy predictions, they are also normative in the sense of being considered as ``what a forecaster ought to do'' and praiseworthy in their own right (or not doing them blameworthy). Of course these forecaster norms are not the only set of epistemic norms that could be useful for a community that seeks to make predictions about the future. For instance, a different community has formed around exploring the idea of prediction markets \citep{wolfers2004prediction}. In this view, the profit motive is expected to create sufficient motivation for individuals to adopt effective forecasting methods. A poor forecaster may lose money to a good forecaster and this may lead the poor forecaster to adopt better techniques in the future. While the forecasting community promotes norms that improve the efficiency of human evidence accumulation, the prediction markets community instead promotes norms that maintain the efficiency of markets. There is more than one way to skin a cat.

While we resist ways of talking that suggest any particular epistemic norm is better than any other on objective ``view from nowhere'' grounds, nevertheless, in this section we have been concerned largely with the norms that seem at first glance to be some of the easiest to justify by their usefulness in science, everyday life, and even in politics. It is tempting to draw metaphysical conclusions, and think that our epistemic norms have some deeper metaphysical status that allows them to anoint some propositions as truths and degrade others as falsehoods. But we don't think it is helpful to do so. We have used the term `epistemic norm' to highlight that the rules of logical argument are themselves norms and thus context dependent, arbitrary, automatic, changing over time, and dependent on sanctioning. A logical argument is a kind of persuasive argument. In particular, it is one that is good for persuading a particular kind of listener who is a specialist in the kind of logic being brought to bear. The logic of the methodology of psychology is persuasive to psychologists, and the logic of the methodology of symbolic logic is persuasive to logicians. These logics also change over time, and undergo ``paradigm shifts'' where both the questions asked, and the ways of answering them, change at once \citep{kuhn1997structure}. And, these shifts are brought about in the same way as other mechanisms of norm change, and thus the speed at which these changes occur is no surprise given the tipping point dynamics of norm change (Section~\ref{section:normChange}).

\subsubsection{Justification and rationality}
\label{section:postHocJustification}

For assembly $x$ to justify assembly $y$ in the eyes of a particular group $S$, the compound sequence $(x,~\texttt{"justifies"},~y)$ must be acceptable under the group's relevant epistemic norms. For example, a community that values empirical evidence will consider $x$ to justify $y$ only if $x$ provides empirical evidence for $y$. A community that values intuition, in contrast, will see some intuitive appeal as a suitable justification. A community that values logic might require that $y$ follow from $x$ by a chain of well-formed inferences. 

Consider a sequence of assemblies $s = (x, y, \dots, z)$ where $y$ denotes a subsequence corresponding to the focal utterance or behavior. The subsequence $x$ is the prefix of $y$ and $z$ is a suffix of $y$. Using the actor model from section~\ref{section:howIndividualsMakeDecisions}, $x$ is the state of an actor's global workspace, which contains answers to summary questions like ``what kind of person am I?'', ``what does a person like me do in this situation?'' and so on. In our model, an actor makes decisions for how to act by considering a sequence of such summary questions in $x$ (i.e.~a ``chain of thought'' in the sense of \cite{wei2022chain}). Assembly $y$ is a proposition to affirm or (verbal) action, and $z$ is an utterance of the actor which it emits some time after $y$.

Since the model of human decision making we consider is autoregressive (i.e.~each assembly is produced by pattern completion from its prefix), the utterance $y$ is directly caused by its prefix $x$. Many theories of human decision making assume individuals think of \textit{reasons} in favor of each of their various alternative actions and then implement the action associated with the best reasons, i.e.~that they apply a chain of thought consisting of cogent reasons for $y$. These theories, which in this context are called \emph{rational}, assume the prefix $x$ justifies the focal utterance or behavior $y$ \citep{kohlberg1983moral, piaget1932moral, turiel1983development}.

Other theories of human behavior view verbalizable reasons as primarily reflecting post hoc justification and not causally related to the real reasons for behavior, which these theories emphasize may not be justifying, i.e.~reasons that external observers would not accept as appropriate or likely \citep{haidt2001emotional}. In these theories, which are called \emph{intuitionist}, it is the suffix $z$ that justifies the focal utterance $y$. In intuitionist theories the prefix $x$ may or may not be justifying. The idea is that $x$ need not necessarily justify $y$ (not that $x$ never justifies $y$). Part of why intuitionist theories feel so counterintuitive is that, according to \cite{rorty1978philosophy} (pg.~248) ``the epistemological tradition confused causal explanations of the acquisition of belief with justifications of belief''. It is hard for us to step aside from the hundreds of years of culture that trained us to unreflectively hold that reasons for actions exist (or ought to exist) before the actions themselves. The  intuitionist view simply holds that an individual's chain of thought which leads them to choose an action or endorse a proposition need not be justificatory of that proposition or action. 

As a result of post hoc rationalization, and the tendency to believe one's thoughts originated within oneself, the beliefs of human individuals are much more influenced by the beliefs of others than subjective experience seems to suggest. That is, we all feel like our beliefs are our own, and caused by our stated reasons for believing what we believe. However, evidence suggests our opinions and even our perceptions are shaped much more by others than we tend to realize \citep{asch1951effects, sinclair2005social, taber2016illusion}.

The \emph{social intuitionist} theory of \cite{haidt2001emotional} holds that, while the reasoning we produce when asked by others to justify a decision we made does not accurately reflect the direct causality of our decision, it does however end up indirectly affecting our likelihood of making the same choice when facing similar situations in the future. This is because once our reasoning is articulated and heard by others it may become part of the overall discourse in society, re-articulated by others, commented upon, debated, and developed \citep{heyes2022rethinking}. A good argument can diffuse rapidly through a society (see Section \ref{section:gadgets} for what it means to be a good argument). The indirect causality in \cite{haidt2001emotional}'s social intuitionist theory arises because we absorb and internalize all the rationalization we hear from others. Over time this experience comes to influence our decisions because we remember effective arguments and parts of arguments and recall them as necessary when we are called upon to justify similar choices. Over time, knowledge of the most commonly heard and repeated arguments is consolidated into the pattern completion network itself ($p$). So when we come up with a really good post hoc justification, even though it did not cause the behavior in the first place, it may ultimately if amplified socially, end up becoming causal for future instances of similar behavior.

Reasoning is a dialogic process wherein two (or more) actors (or an actor and a piece of paper) take turns priming one another and responding themselves to the other's last prime~\citep{baier1997commons, mercier2011humans}. Each inference step takes the contents of the the global workspace and maps it to the next utterance, which is its continuation \citep{zada2023shared}. If the context is such that the pattern completion network $p$ must perform inference to predict the next symbol then it will do so. Since sentences transform into other sentences via rules, we can attempt to articulate the rules of logic used and the kinds of reasons it may employ to sample the continuation. However, the logic of how one sentence follows the next will only occasionally be human understandable since it generally depends on whatever the pattern completion network learned to do. In the rare cases where we can provide a human-legible account of how one thought follows another these are because the situation matches one of many standard rules of inference, e.g. \textit{modus ponens, etc} (e.g.~as in Section~\ref{section:examplePatternCompletion}).

Individuals are often more skilled at crafting justifications for pre-existing beliefs than they are at impartially evaluating the available evidence, in fact most people are quite bad at impartial evaluation  \citep{mercier2017enigma}. It is difficult for individuals to avoid selectively recalling information that supports their prior views (confirmation bias) \citep{mercier2011humans} and self-serving incentives (my-side bias) \citep{taber2016illusion}. Therefore, in the context of a group conversation about whether or not to take some collective action, each individual may propose arguments largely as a function of these biases. However, through discussion, as long as the group has some shared epistemic norms, and power differences between group members are not too large, and all group members are engaged in the exercise \citep{woolley2010evidence}, then better proposals (according to their shared epistemic norms) can be recognized collectively and adopted. Furthermore, in our social intuitionist view, when the group collectively decides to adopt a different proposal than was suggested by the focal actor, then this still gives the focal actor experience following whichever policy was selected, which continues to train their pattern completion network, and if repeated enough, can lead to its being incorporated implicitly into the focal actor's future behavior. So the dynamics of collective decisions by groups with shared epistemic norms are a key vector in social intuitionism's indirect pathway, in which a justification of a behavior eventually becomes a cause of that behavior \citep{haidt2001emotional}.

\section{Appropriateness with family and friends}
\label{section:familiarPeopleAppropriateness}

In Section~\ref{section:appropriatenessWithStrangers} we showed how behavior may be normative, i.e.~appropriate by virtue of being suggested by generically conventional sanctions. However, behavior is also influenced by sanctions determined via conventions of more narrow scope such as friendship or family groups. When narrowly conventional sanctions diverge from those of generic scope it is possible for individual behavior to diverge from the norm, and therefore to be inappropriate in the eyes of the broader society. Since such behavior may still align with the narrow scope conventional sanctions of one's own family and friend groups, it would still be regarded as appropriate within those groups.

Idiosyncratic narrow-scope conventions (including conventional patterns of sanctioning) depend on the precise history of interaction between individuals. For instance, when two people, Alice and Bob, play tit for tat, whether Alice defected in the last round determines the appropriateness of Bob's defection in this round.

Unlike normative appropriateness, which is based on generically conventional sanctions, and thus can only be changed by collective action, relationship-based appropriateness can be altered by individual decisions. For instance, defection can ruin a friendship, changing the appropriateness of a wide range of different potential behaviors of both individuals.

\subsection{Emotions and personal relationships}
\label{section:emotionsAndRelationships}

As factors that influence appropriateness, social identity, social role, and personal identity are all distinct from each other. The way we used the terms in section~\ref{section:stylizedFactContextDependence}, social identity and social role both refer to factors that determine appropriate interactions between \textit{strangers}, with social identity corresponding to the more slowly adapting of the two. While both social identity and social role do still affect appropriateness in individuals who know each other well, their influence shrinks relative to that of the specific relationship of the individuals in question as they gain more personal experience with one another. There are many ways in which relationships between individuals may override the effect of broad societal norms and conventions on what it is appropriate to say. For instance, two friends may develop a bantering style of speech with one another where ostensibly insulting utterances are automatically understood as playful jokes\footnote{Here's another example. An employee and their manager may develop together an informal understanding of the rights and obligations incumbent on their relationship, a \emph{relational contract} e.g. it's OK to leave the office early on Fridays to pick up a child from school. Relational contracts can go beyond the provisions of both other sources of expectation that structure their relationship: their formal employment contract, and the informal norm's judgment of appropriateness for the context. Relational contracts depend on the expectation of continued interaction over time, ``the shadow of the future'', and are thus modeled with two-player iterated matrix games \citep{baker2002relational}.}. This works even if the norms governing similar contexts and relationships would deem the same utterances inappropriate. 

Intriguingly, the theory we have presented in this paper suggests that both explicit and implicit relationship-guided appropriateness are possible. The explicit mode would involve retrieval of one's specific interaction history with a particular partner (or a suitable summary of it) from long-term memory upon meeting them again, and is constrained by memory capacity \citep{milinski1998working}. The implicit mode would involve weight changes in the pattern completion network itself so that all or many contexts involving a particular partner are modulated in a certain way. After enough experience with a specific individual, e.g., a family member, one may consolidate a representation of appropriateness within the relationship.

\cite{fiske1992four} articulated an ontology for classifying social relationships and their constituent motivations, saliencies, and behaviors and showed how to apply the framework to understand features of relationships expressed in different ways from culture to culture. According to \cite{fiske1992four}'s ontology, there are four elementary forms of relationship, corresponding to the four common ways of organizing data. The form of relationship between two individuals has far-reaching and culture/context-specific implications for the kinds of behavior appropriate or inappropriate for the two to direct to one another. Three of them are relevant to our theory (while the fourth, market pricing, seems likely driven more by norms and conventions than personal factors, so we do not discuss it here). The three relevant forms are: (1) \textit{communal sharing}, (2) \textit{authority ranking}, and (3) \textit{equality matching}. We discuss each in the sections that follow.

It is also notable that emotions are deeply involved in the implementation of personal relationship-based representations of appropriate and inappropriate behavior. \cite{sadedin2023emotions} models long-lasting pair bond formation as a process through which individuals come to condition their behavior on relationship-specific state variables regarded as corresponding to an emotional ``bookkeeping'' system which tracks each partner's attitudes toward the relationship. We discuss next how each of \cite{fiske1992four}'s relationship types are formed, shaped, and maintained by emotions. The view of affective cognition we endorse and apply here is one where basic emotional categories are, to a large extent, plastic, contextual, and constructed from experience as proposed by constructivist accounts of emotion such as those proposed by \cite{schachter1962cognitive, russell2003core, barrett2006solving}. In this view, emotions are learned context-dependent categories superimposed onto basic patterns of interoceptive sensations e.g.~an empty feeling in one's abdomen may be interpreted either as being hungry when you are in a restaurant waiting for your food to arrive or as a sense of existential dread when you have the exact same abdominal feeling while waiting for test results in a hospital.

\noindent\textbf{Communal sharing}

Close kinship relationships as well as ethnic and national identities often include concepts, motivations, and behaviors falling under the category of communal sharing. Sharing of resources within the group does not require keeping track of who contributed what or who took what. Motivations are often altruistic toward the in group and often hostile toward out groups, e.g.~ethnocentrism. Markers of group membership are salient. Many cultures and military organizations use stressful ``ordeals'' as bonding rituals to foster very strong communal sharing relationships between certain groups that will need to work closely together \citep{haidt2012righteous, whitehouse2021ritual}. Members of a group or dyad related according to communal sharing treat each other as equivalent in the sense that honor or dishonor of one group member ``speads'' to the rest of the group.

In non-human animals, pair bonding includes behaviors such as vocal duets, mutual displays, partner contact behaviors, food sharing, grooming, assistance in aggressive encounters, mate guarding, joint territorial marking and defense, distress upon separation from the pair mate, and amelioration of distress upon reunion \citep{bales2021pair}. Consoling behaviors are directed toward distressed individuals. In human folk psychology, consoling behaviors are understood to be driven by ``empathy''. Consoling has also been observed in other species including great apes, elephants, wolves, and prairie voles \citep{walum2018neural}. The fact that parental care, pair bonding, and consoling behaviors between unbonded individuals are all regulated by the oxytocin system (in prairie voles) suggests that evolution may redirect ancient maternal instincts toward unrelated adults. This hypothesis is relevant to humans since pair bonding mechanisms are thought to underpin romantic love and sometimes also friendship \citep{dunbar1998social}, in this view both would be elaborations of ancient maternal instincts mediated by the oxytocin system. The fact that in humans there is large cultural variation in both how pair bonds form and how they are expressed \citep{quinlan2008human} indicates that the system is by no means ``dumb'', but is actually very strongly adaptive to the local social environment.

\noindent\textbf{Authority ranking}

Members of a group or dyad related to one another through authority ranking treat each other as being linearly ordered according to a hierarchy where some are ``higher'' or ``better'' than others. Higher ranked individuals typically enjoy benefits and privileges. There may also be an obligation of higher ranked individuals to protect lower ranked individuals or to take care of them. Individuals may come to develop hierarchical relationships to one another via the evolution of their specific personal relationships. Many non-human primates also form hierarchies of this kind, e.g.~higher ranked chimpanzees are groomed more often by subordinates \citep{dunbar1996grooming} and usually have more offspring \citep{boehm1999hierarchy}.

Dominance hierarchies can also result from conventions and norms, in which case they apply to strangers, not just to individuals who know one another, e.g.~India's caste system and relations between races in the US \citep{wilkerson2020caste}. These dominance hierarchies are embedded in the broader set of norms and concepts of the culture such that roles are constituted generically \citep{gobel2023self}. There may be multiple parallel scales of hierarchy, \cite{henrich2001evolution, henrich2015big} suggest prestige motivations---e.g.~to be imitated by others---are distinct from ``dominance motivations''---e.g.~motivations to control and enjoy benefits from subordinates without reciprocity toward them.

The plasticity of the basic emotions is also important here. For instance, in males, individuals are more likely to engage in dominance hierarchy challenge behaviors when their testosterone levels are high \citep{archer2006testosterone}. However, the particular set of behaviors regarded as constituting a dominance hierarchy challenge are culturally and contextually determined. The very idea of a dominance hierarchy challenge presupposes that the hierarchy is responsive to the actions of individuals with respect to one another. In the kinds of hierachies imposed by conventions and norms, which apply to strangers, these challenges would make little sense. Or rather, they could never happen by unilateral action and would instead always require collective action of some kind.

\noindent\textbf{Equality matching}

Members of a group or dyad related to one another by an equality matching-type relationship, conceptualize an abstract or concrete quantity or concept to balance per relationship. It includes motivations related to egalitarian distribution, in-kind reciprocity, tit-for-tat retaliation, eye-for-an-eye revenge, and compensation by equivalent replacement. Inequity aversion in the sense of \cite{fehr1999theory, hughes2018inequity} is in this category too.

The specific tit-for-tat interaction history is important here. If we are friends in a pub, it may be appropriate for me to buy the next round after you bought the previous one. Whose turn it is determines who appropriately buys the next round.

There are emotional motivations to maintain deterrence e.g. hot reactive emotions at being shortchanged in food sharing \citep{sterelny2019norms}. Note that this behavior is not unique to humans \citep{brosnan2014evolution}. A key difference between norms and these anger/resentment-linked mechanisms of cooperation via deterrence and is that they depend on self enforcement whereas norms depend on third-party enforcement.

\section{Humans as political and community-participating animals}

We are now able to spell out our theory's proposed explanation for each of the stylized facts of appropriateness we articulated as our explanatory targets in Section~\ref{section:desiderata}.

\begin{enumerate}
    \item \textbf{Appropriateness is context-dependent.}
        \begin{enumerate}
            \item \textbf{Situation dependence}---everything noticed about the current situation is in the the global workspace and conditions the choice of how to behave.
            \item \textbf{Identity and role dependence}---extra information about the identity of the actor is also always in the global workspace. Social identity has a large patterning effect on appropriate behavior across a large number of situations and roles.
            \item \textbf{Culture dependence}---each culture may consider a different set of behaviors, dress, and demeanor to be appropriate for each situation.
        \end{enumerate}
    \item \textbf{Appropriateness is arbitrary}---response features of no material consequence may be moralized. There is no requirement for conventions and norms to be fair or efficient (e.g.~\cite{koster2020model, vinitsky2023learning, tang2023unequal}). While some norms are likely non-arbitrary (see Section~\ref{section:stylizedFactSocialHarmony}), e.g.~those critical for facilitating cooperation), the mechanisms by which norms emerge and are maintained pose little constraint on their content. The arbitrariness of norms is inherited from the arbitrariness of conventions. Since conventions are arbitrary in the sense of Section~\ref{section:conventions} it is thus possible for arbitrary patterns of sanctioning to become conventional, i.e.~to generate norms. 

    \item \textbf{Acting appropriately is usually automatic}---In the computational model we set out in Section~\ref{section:howIndividualsMakeDecisions}, when the concept of appropriateness for context $c$ has already been consolidated, then the pattern completion network $p$ only predicts appropriate completions when prompted with $c$. This stylized fact also includes the sub-claim that acting appropriately is no more cognitively demanding than acting inappropriately and acting appropriately is usually automatic. In particular, effortful prospective simulation such as is needed for the more sophisticated mental exercises of perspective taking are not needed very often. Note that it's not that they are never employed. Sometimes we do think quite a bit in order to decide how best to act appropriately. But these are typically difficult and unfamiliar cases or cases where multiple norms conflict with one another. Our point is that most of the time we need not think so much in order to act appropriately. This is reflected in the case of appropriateness with strangers by the difference between explicit and implicit norms (Section~\ref{section:implicitVsExplicit}). Implicit norms reflect the fast-to-compute consolidated sense of appropriateness whereas explicit norms operate via step-by-step reasoning in the actor's chain of thought. Therefore applying an explicit norm requires reasoning, and can be disturbed by distractors (e.g.~irrelevant information in the global workspace $z$), guidance by an implicit norm does not require reasoning and is not disturbed much by distractors. We argued in Section~\ref{section:implicitVsExplicit} that a much larger fraction of human behavior is guided by implicit norms automatically than reasoned out using explicit norms, and in Section~\ref{section:postHocJustification} we argued that humans are not really very good at reasoning out how to apply explicit norms, and tend to make self-serving errors. So it's perhaps a very good thing that the stability of society does not depend too much on such reasoning.
    
    \item \textbf{Appropriateness may change rapidly}---Societal change is driven by positive feedback dynamics as everyone shifts to play their part of a newly emerging status quo. We discussed several mechanisms including both deliberate collectice action and more emergent bottom-up processes like the bandwagon effect in Section~\ref{section:normChange}. Norm change can  proceed quite rapidly since there are strong positive feedback cascade dynamics (e.g.~rapid changes in the appropriateness of smoking cigarettes in public, mediated by informal sanctioning~\citep{alamar2006effect}).

    \item \textbf{Appropriateness is desirable and inappropriateness is often sanctionable}---This was a consequence of the definition of norm we offered in Section~\ref{section:normsDefinition} to capture appropriateness with strangers.
\end{enumerate}

\subsection{Non-social instrumental knowledge, innovation, and subsistence in the human niche}
\label{section:instrumental}

We assume that humans are more often guided by appropriateness than by instrumental task goals. That we are more often guided by appropriateness is consistent with models of human uniqueness emphasizing our species' critical dependence on social learning \citep{boyd2011cultural} as well as models of human uniqueness that emphasize the human ability to solve the special challenges that arise from living together in groups \citep{dunbar2017there, duenez2023social}. Nevertheless, it is important to note that there is more to life than appropriateness. There are many non-social reasons for action (e.g.~hunger motivates foraging). People weigh a range of different factors in determining what to say and do, how to dress, and how to conduct themselves. There are various reasons to act inappropriately. For example, upon meeting an acquaintance it is sometimes appropriate to engage in a lengthy discussion of the weather before moving on to important matters. In that case one may have good reason to cut the weather talk inappropriately short in cases of time pressure. In another example, there are contexts in which it may be appropriate to reciprocate a gift with another gift of even greater value \citep{graeber2012debt}. However, if one cannot afford such a gift then one may choose to act inappropriately rather than go into debt just to give an appropriately expensive reciprocating gift. Also, political actors balance competing motives for action knowing that they will ultimately be held responsible both for the appropriateness of their actions and for their consequences \citep{march2011logic}. In general, the motivation to act appropriately trades off with other factors, including how onerous compliance would be (the pattern completion network $p$ makes the decision). Moreover, there are sometimes other reasons to act inappropriately beyond trade-offs with self-interested factors. Sometimes one is unhappy with a particular norm of their community and by publicly acting in a way deemed inappropriate by it may hope to weaken its hold. For instance, norm entrepreneurs like Rosa Parks intentionally transgress norms in order to register their disapproval of them and erode their support \citep{porpora2016great, bicchieri2018norm, sunstein2019change}. However, such strategies often fail, and in general it is difficult for individuals to change norms since doing so requires collective action to change the relevant patterns of sanctioning\footnote{The hockey vignette on the first page of \cite{schelling1973hockey} illustrates the difficulty of this collective action problem in a memorable way.} (see Section~\ref{section:normChange}).

Reward-based theories of ``the secret of our success'' such as \cite{silver2021reward} are largely theories of how humans acquired our fit to our ecological niche, especially our generic problem-solving abilities (i.e.~our intelligence in the view of \cite{legg2007universal}). Explaining how people have instrumental knowledge of subsistence strategies is home territory for the reward-based theories since this domain is rife with consequences for actions and knowledge is deeply non-arbitrary. Nevertheless, we discuss in this section how the pattern completion theory can also account for the same facts and highlight how it offers a simpler account in many cases. 

The idea is that effective subsistence strategies are rewarded while ineffective subsistence strategies are punished. Both rewards and punishments are seen as delivered by the ecological environment itself through the good or bad consequences of actions sampled by individual learners. Either through trial-and-error \citep{mnih2015human, blundell2016model} or through simulation-based action selection \citep{tenenbaum2011grow} or a combination of both such as in~\cite{dayan2008decision}, the consequences of acting in one way versus another create the impetus to change. In fact, in reward-based theories, consequences drive both behavior change and behavior refinement \citep{ajzen1991theory}. This makes them usually function as ``demand side'' theories of innovation. Their basic idea is that, because an ecological incentive or requirement exists, and ``necessity is the mother of invention'', the individuals will---if given enough time to explore---discover a way to take advantage of the incentive. The problem that ``supply side'' reward-based theories of innovation try to address is that in order for this exploration to take place you may have to wait an extraordinary long time \citep{peterson2004day}, so there is a need to develop a theory not just of why innovation responds to incentive (i.e.~how learning works), but also a theory that explains the supply of incentives. Supply-side theories of innovation regard incentives not as fixed functions of the environment but rather as dynamic and constantly changing functions of physical/ecological and social factors, including the learning of others. These theories seek to explain how interaction gives structure to innovation and make it possible to explore strategies that are very dissimilar to initial strategies by following continual guidance of a relatively smooth gradient generated by the simultaneous learning of multiple individuals, each following their own incentives \citep{leibo2019autocurricula}. Reinforcement learning algorithms based on the idea of self-play illustrate how this idea works. The basic idea is that as you improve, your sparring partner also improves since they now learn against a better opponent, but since they have improved they are now also a better sparring partner which creates further impetus for you to improve, and so on. Your opponent remains always on your level (at least in part), and that level is best as incentive for you to learn to do even better \citep{tesauro1995temporal}. There has been a great deal of work probing the properties and limitations of this kind of model \citep{balduzzi2019open, czarnecki2020real}. They work both in competitive \citep{silver2017mastering, jaderberg2019human, vinyals2019grandmaster, baker2019emergent, plappert2021asymmetric} and mixed-motive  cooperation-emphasizing settings where the idea is less about out-competing an adversary and more about co-learning a coordinated solution \citep{leibo2019malthusian, strouse2021collaborating, agapiou2022melting, johanson2022emergent}.

But these theories face two parallel troubles. First, they have trouble explaining arbitrariness, and second, they wrongly predict almost all communication to be ``cheap talk'' contrarily to what happens both in real life and in laboratory experiments, where people who have met each other typically employ very different joint strategies than they would with anonymous strangers whom they cannot communicate to \citep{ostrom1992covenants, sally1995conversation}.

One approach taken in some reward-based theories is to say that the dynamics of equilibrium selection may sometimes naturally have features making them prone to tipping point learning dynamics, e.g. in the driving orientation game frequently used to explain conventions \citep{marwell1993critical, heckathorn1996dynamics, vanderschraaf2018strategic, koster2020model}.

In what way do they try to handle arbitrariness at all? One way is via arguments like those used to explain the peacock's tail. These explanations suggest the idea of a proxy being used for optimization rather than the ``true'' target \citep{john2024dead}. Many have put forward theories of social phenomena centered on the idea that individuals seek to maximize their social status  \citep{roberts2021benefits, semmann2004strategic, zahavi1999handicap, gintis2001costly}. Social status is a natural proxy to posit for one who is determined to explain social phenomena by reference to optimization of a reward-like signal. However, these theories are only able to explain the parts of the data where people have a well-defined notion of status and moreover, they simply posit the status maximization motive and make no attempt to explain where its associated preferences come from in terms endogenous to the theory. Like many other theories centered on the idea of innate tastes, they appear to regard their central driver exogenously. Furthermore, social status maximization theories are mostly silent on the question of which situations are those where one should expect status motivations versus which situations are those where one should expect other motivations. 

All the reward-based theories have in common that they have more trouble explaining the numerous psychologically instrumental customary behaviors which are thought by the individuals who engage in them to have material non-social consequences, but do not. These include magical spells, amulets, and wards against ``the evil eye'' \citep{jagiello2022tradition}.

Recall that we talk about conventional sanctioning patterns as elements of an individual's guidance, and we emphasize that each guidance element corresponds to the perspective from a particular social scope, i.e~the social reasons that come from that scope. We associated each guidance element with an ordinal preference relation. Then we integrated all the guidance elements into an intra-individual ``social choice'' to create the individual's aggregate preference, which their behavior reflects (see Section~\ref{section:endogenousPreferenceFormation}). Notice that this view also accommodates the possibility that there may also be non-social reasons for action that operate in the same way. There's a non-social preference relation coming from the individual's understanding of physics, which foods are poison, etc. These are all aggregated into the ``social'' choice in the same way.

How do individuals and groups learn about non-socially originated reasons for action such as foraging strategies? In accord with the view from the work of Boyd, Richerson, and Henrich and others, we think individuals acquire much of the know-how they need to survive in our ancestral hunter-gatherer niche via social learning from other individuals in their local group \citep{boyd2011cultural}. We also think that group-level selection explains a large part of the difference between groups and is the explanation for why social learning mechanisms that could, in principle, land on arbitrary strategies, in practice, end up landing mostly on strategies that turn out to be reasonably well-adapted to the slow-changing part of their circumstances \citep{wilson1975theory, richerson2016cultural}. 

Note that even though the subsistence knowledge itself is non-arbitrary, e.g.~which local plants are poisonous, the cultural rationale for the knowledge is arbitrary. For instance, a group's understanding of which plants should be avoided may align well with the set of plants that are in fact poisonous because they view the poisonous plants as sacred to the gods, and treat eating them as a sanctionable transgression. What matters is the customary behavior's ultimate alignment with the poison ecological factor, not the group's understanding of the causal process giving rise to it \citep{derex2019causal, harris2021role, henrich2021cultural, koster2022spurious}. Individual-level trial-and-error learning may also play a role in guiding the overall cultural evolutionary process in some situations, though not those involving lethal errors \citep{mesoudi2018cumulative}. Notice that individual trial-and-error learning, followed by social learning from other individuals, may also give rise to the free-floating cultural justifications for important physical and ecological knowledge that we see in practice. Just because knowledge was acquired through individual learning doesn't mean it was acquired using a sophisticated causal model. The fact that justifications can vary with considerable latitude around beliefs, even for non-social ecologically instrumental customs, is also expected in light of the discussion of post hoc justification in section~\ref{section:postHocJustification}.

Another stylized fact which would be puzzling for a reward-based theory is the fact that social sanctions are a much more compelling reason for action than individual evaluation. Insofar as we have base non-social preferences, many of them may turn out to be illusory on further inspection, and of the ones that remain they are likely very weak and easy to override by social sanctions. Animals may get by with mostly innate tastes, but we humans construct ours. Spicy food is a good example from \cite{henrich2017secret}. Plants evolved their spicy taste as an anti-mammal device. Capsaicin literally activates pain receptors. And most non-human mammals don't eat spicy plants. So it mostly worked! But we humans, in the cultures that like spicy food, are happy to eat them. In fact we often prefer them to non-spicy plants.

\subsection{Collective decision making}
\label{section:socialChoice}

What is a society? In the view we advocate here, a society is a group of individuals linked by a set of conflict resolution institutions like a system of laws, courts, and legislatures \citep{march2011logic, mouffe1999deliberative}. This picture accommodates the possibility of within a single society there being deep and persistent disagreements about underlying values that are not reducible to mere misunderstandings. Such a society can function despite its internal misalignment as long as those whose idiosyncratic preferences lose out in collective decisions still regard the system as a whole as legitimate \citep{hadfield2014microfoundations} or are unable to avoid falling under the remit of its institutions. \cite{tainter1988collapse} and \cite{fernandez2024abstract} argue that even substantial maliciousness between segments of society does not on its own cause societal collapse unless accompanied by broader institutional failure.

Many familiar conflict resolution mechanisms look like collective decision making processes. For instance, the mechanism by which the group outcome is determined may have an impact on groups dynamics in politics~\citep{schofield1988groupchoice, carley1991stability}. This hypothesis, known as \textit{Duverger's law}, state some ways of aggregating voters' preferences may be predictive of the observed party formation dynamics~\citep{duverger1954law}. The \emph{first-past-the-post} system used in many democratic countries seems to lead voters dynamics toward having only two major parties~\citep{riker1982duverger, palfrey1988duverger}. The U.S House of Representatives or Canada's House of Commons are examples of this phenomenon. Other voting mechanisms such as \emph{proportional representation} are associated with political systems where many small parties coexist. The diversity of parties and political ideologies in the European Parliament can be taken as examples of Duverger's predictions.

On the other hand, whenever individuals can estimate how their preferences differ from others in the group (e.g., by observing the sanctioning behavior of others), they may strive to balance benefits of belonging to the social group with those obtained from picking a group that closely matches their preferences~\citep{axelrod1967conflict, de_Swaan1973coalitionschocie, vanDeernen1991socialgame}. In political systems, voters may be motivated simultaneously toward joining parties that have an opportunity to govern (e.g., being in a wining coalition) and toward joining parties whose ideology is not far from their own~\citep{carley1991stability}. In social media, users can \textit{vote with their feet}. Many have abandoned major social networks and joined smaller alternatives with norms that fit more closely their own preferences~\citep{hoover2022twitterflock, korsins2023altrightflock}. However, many have remained on larger platforms despite disagreeing with large segments of their population of users, balancing the benefits of belonging to a smaller and more coherent community against the benefits of the larger but less coherent community (e.g., audience and reach of the content they produce, monetary compensation, etc.)~\citep{paul2024substacknewton}.

We have suggested that a defining feature of human social life is the ability to manage and navigate the conflicts that inevitably arise from the diversity of human perspectives. Societies function not because everyone agrees, nor because everyone ``would agree'' if only they knew better. Rather, societies function because they have developed mechanisms for resolving disagreements, preventing them from escalating, and coordinating collective action in the face of divergent preferences. Humans live with misalignment. Appropriateness serves to coordinate our expectations and to deter antisocial behavior, which in this view can include conflict-seeking behavior. There may be places a conversation could go where conflict would follow, so it could be inappropriate to bring a conversation there. Or if a conflict has begun, there could be appropriate mitigations to prevent it from spiraling out of control. All of these possibilities are culture dependent and play out differently context to context. Though some norms have little material value or negative value \citep{koster2022spurious}, other norms encourage cooperative behavior and discourage antisocial behavior \citep{ullmann1977emergence}.


\part{Machines}  \label{part:machines}

\noindent\fbox{%
    \parbox{0.95\textwidth}{%
``We pragmatists think of moral progress as more like sewing together a very large, elaborate, polychrome quilt, than like getting a clearer vision of something true and deep.''
\flushright{---\cite{rorty2021pragmatism} pg. 141}
  }%
}

Now, as we begin to discuss applications, it is important to distinguish between \emph{companion} generative AI applications and \emph{tool/service} generative AI applications. A companion generative AI system is one that seeks to drive affiliation with a specific user by building a persistent history of interaction with them in particular. Therefore the logic of appropriateness with friends and family is applicable to this case (see Section~\ref{section:familiarPeopleAppropriateness}). A companion bot may also utilize emotional language and visuals or portray itself as an individual with a personal connection to the user \citep{pentina2023exploring, verma2023they}. Unlike companion bots, a tool/service bot may interact with a user in a more functional manner like the way that a merchant interacts with their customer or a police officer interacts with a suspect. In humans, such interactions would be governed by the logic of appropriateness with strangers (see Section~\ref{section:appropriatenessWithStrangers}). Tool/service generative AI systems may or may not build up persistent interaction history with individual users (a merchant may remember a repeat customer without becoming their friend). However, a companion AI would always have to interact with each individual user as if they are a friend or romantic partner, so persistent user-specific memory is critical for companion bots but optional for tool/service bots. Companion bots also would usually remember different details of their interactions than tool/service bots would such as birthdays and names of family members since their goal is to build build rapport with users. The two categories aren't entirely mutually exclusive since some tool/service AI products may be designed to try to establish relationships with users.

In this paper we focus mainly on tool/service generative AI applications. We did include discussion of personal relationships and emotions for completeness when we talked about human appropriateness (Section~\ref{section:familiarPeopleAppropriateness}), however we do not consider the practicalities of building companion generative AI applications\footnote{Indeed, special ethical considerations, different from those we focus on, may apply to AI systems that seek to build individual relationships with specific users and track their emotions \citep{defreitas2023chatbots}.}. Our focus here is on the part of appropriateness driven by norms i.e.~appropriateness with strangers.

Our aim in this paper has been to understand why the problem of AI actors behaving inappropriately exists. Importantly, we aim give equal weight to the version of the problem where safety filters and related design choices go too far and cause bots to ``self censor'' by refusing user requests too often---an outcome which is just as problematic if users often feel the refusals they experience are inappropriate. We cover a few different ways the AI research and technologist community is currently thinking about these issues. In particular, we discuss how design choices are presently instilled in generative AI actors (Section~\ref{section:statusQuo}), and what implications there may be of this process for polarization in the sense of social media (Section~\ref{section:polarization}). We also discuss ``pluralistic alignment'' efforts which aim incorporate more diverse voices in the process of determining an AI system's sense of the appropriate (Section~\ref{section:pluralisticAlignment}). Finally, in what we think is really the crux of the matter, we discuss contextualization as being the too-often missing ingredient in light of the preceding discussion of product design methods and the fine-grained context-sensitivity of the human sense of the appropriate (Section~\ref{section:lackingContext}).

When local communities and app developers can run their own LLM instances, perhaps initialized from a base template, or when they can affordably fine tune a remote model via an API, this allows for greater experimentation on how to teach norms to AI, which may then lead to more accurate and adaptive representations of norms for specific niche communities (Section~\ref{section:decentralization}). For instance, the decentralized approach suggested here could empower communities to trigger instruction fine-tuning on their own datasets, leveraging the demonstrated efficacy of curated small datasets \citep{zhou2023lima}, for example, in adapting models to niche contexts like comedy writing or specialized search. We argue that a decentralized ecosystem where groups of all shapes and sizes can customize the norms used by their AI actors also helps promote contextual awareness and dynamism since communities could legitimately allow continuous, online learning using feedback from members. The customization toolkit should include the same sanctioning signals that humans employ, such as implicit feedback and inferred emotional cues (Section~\ref{section:unidirectionalSanctioning}). We also discuss whether there are situations when humans learn norms of their own communities from AI actors and conclude this is indeed possible, though not always desirable (Section~\ref{section:bidirectionalSanctioning}). Finally, we consider independent cultural evolution of norms in machines-only social systems (Section~\ref{section:machinesOnly}).

\section{The status quo}
\label{section:statusQuo}

Ultimately, an AI's operating conception of what is appropriate is a deliberate design choice: does it speak formally or informally? Does it represent itself as a robot or a friend? Does it insult users or treat them respectfully? It may be shaped to fit the demands of its application domain and deployment context. The bot's designers as well aim to establish a certain voice and tone in the bot's responses e.g.~to be confident and to never claim to have preferences, feelings, or opinions. These design choices may be implemented in various ways including supervised fine-tuning to follow instructions and reinforcement learning from human feedback, and these methods are not passive filters of a ``generic human preference'' but rather are better understood as convenient leverage points for deliberate product design \citep{ouyang2022training}.

Many existing systems implement what a central authority, the actor's owner, understand to be appropriate. For instance, owners may decide which data sources to incorporate into  training sets, how to instruction tune \citep{ouyang2022training}, and how to perform reinforcement learning from human feedback \citep{ziegler2019fine, stiennon2020learning, glaese2022improving}, including which humans to hire as raters, and what instructions to give them. In fact, the raters' instructions can be quite precise when it comes to the kind of preferences they should provide in order to implement an intended product design specification. So, while many are under the impression that RLHF is about measuring the preferences of the raters, it is not necessarily so. Rather, raters could function less like like voters and more like juries. That is, they could be asked merely to apply human cognitive capacities ``mechanically'' to reach a decision in the bounds of the rules as explained to them as part of the way their choice is framed, not to exercise their own independent judgment, or try to put the thumb on the scales in favor of their own preferenes. Like juries, raters can simply told to apply a rubric in their answers, and to answer on behalf of the central authority, not themselves.

The creators of generative AI products may act to instill in generative AI products their own values or those of their users, regulators, or other stakeholders. These centralized notions of appropriateness may emerge from similar incentives to those that motivate the content moderation policies of social media platforms. Thus they may not be very responsive to particularistic concerns of smaller communities or their thickly construed morality \citep{foster2023thin}. However, the situation of the LLM providers is also different from that of the social media platforms since the platforms enforce rules for human users while the LLM providers may be seen as themselves ``speaking'' in their own voice. On each social network, the human users must adhere to platform-defined guidelines that define what is considered appropriate behavior. But, since generative AI users directly consume content produced by the software itself, its owners are thus motivated differently, and are more likely as a result to design substantively more stringent content moderation standards for their self-generated content. These may look more like the guidelines a corporation would issue to its human employees on how to represent it in public. Despite these differences, there are important transferable lessons from social media content moderation which remain applicable to LLMs.

In the social media case, maintaining a homogeneous concept of appropriateness in large-scale centralized social networks has proven challenging. Users don't always agree with one another or with central authorities on appropriateness, even within a single cultural context. Sometimes when there is an unmet demand for centralized moderation, users can take the matter into their own hands by coordinating with one another (e.g.~in developing third-party services like \textit{BlockTogether} through which users can maintain collective block lists to prevent abusive content from reaching members and to collectively sanction violators). Such tools emerged, for instance, during the \textit{GamerGate} harassment campaign~\citep{Wikipedia_contributors2023Gamergate} and there have been proposals to try to encourage a larger ecosystem of such content moderation ``middleware'' providers~\citep{keller2022lawful}. Other platforms such as Mastodon have also emerged which are designed to facilitate such efforts and as such also to allow experimentation with different moderation policies in a more decentralized fashion~\citep{mastodon-gGmbH2024moderationactions}. It seems likely that similar dynamics pushing toward decentralization will also affect generative AI applications. No one wants to interact with or work with an AI actor that implements a concept of appropriateness they personally find offensive, though they also do not want their AI systems to be so bland and inoffensive as to avoid all potentially controversial topics.

\subsection{Polarization, backfire, and sycophancy}
\label{section:polarization}

A substantial body of work has been aimed at the question of whether users of social media become isolated from ideologically diverse perspectives, i.e.~inhabiting ``filter bubbles'' or ``echo chambers'' where they encounter only the like-minded. The line of work also asks whether said isolation may contribute to increasing political polarization and whether interventions that depolarize social media feeds may reduce political polarization.

With generative AI, it is possible that new variants of these issues may emerge. For instance, bias and prejudice in generative AI may influence the humans who use these systems. As a result, bias may accumulate and grow over time as a result of the human-machine interaction since users are reinforced by chatbots who in turn reinforce those same users, forming a feedback loop \citep{glickman2024human}. This is not dissimilar to the way recommender algorithms promote more and more outrageous content over time \citep{brady2023norm}.

There is evidence both that social media use overall and specifically exposure to like-minded sources are associated with both ideological and affective polarization \citep{kubin2021role}. However, it does not follow that interventions to depolarize social media feeds reduce ideological or affective polarization. Instead, several experimental studies of selective exposure to counter-attitudinal information yielded either null effects \citep{nyhan2023like} or backfire effects \citep{bail2018exposure, kim2019cross}, where feed depolarizing interventions actually increased ideological or affective polarization. \cite{garrett2014implications} points out that when partisans are exposed to counter-attitudinal information they may respond with ridicule or counterargue using motivated reasoning (e.g.~as in \cite{taber2016illusion}), which could increase confidence in partisan beliefs and explain the backfire effect.

Many generative AI language models exhibit sycophantic behavior where bots produce biased responses that cohere with stated beliefs of users \citep{perez2022discovering,  sharma2023towards, turpin2023language}. There is concern that this may increase human polarization through similar mechanisms to those that operate on social media since it causes partisan users to receive partisan responses from bots. If communication with bots becomes a large part of one's information diet then it is plausible that some of the same considerations as with social media may arise. However, on the other hand, this is not a given. The framing around a conversation with a bot is very different from the framing of a social media feed. A biased social media feed appears to the user like a signal that all their friends support a biased position, or that others with the same social identity support it. A conversation with a chatbot is different in that it provides the user no information on what their friends think. Users may see the bot as a corporate representative, making it dissimilar to themselves, and thus an unlikely model to select for social learning of political attitudes, beliefs and behaviors \citep{bandura1969social, iyengar2012affect}.

Efforts to reduce sycophancy in generative AI chatbots involve fine-tuning them to provide neutral perspectives on controversial issues where opinions of partisans differ. Many chatbots available today appear to have implemented fine-tuning along these lines. However, we are aware of no evidence this approach is effective at reducing ideological or affect polarization in users. Moreover, it may create substantial risk from the backfire effect. Anecdotally, one does not have to look far on social media to find users outraged at bots responding neutrally in contexts where they believe it would be more appropriate to take a side. Furthermore, enforced neutrality makes it hard for informed users to explore sensitive topics, and can sometimes overgeneralize to topics merely adjacent to those that are sensitive \citep{wolf2024tradeoffs}.

Some users may view AI services, or AIs in general, as authoritative and therefore as a good source for learning about objective facts~\citep{helberger2020fairest, araujo2020ai}, even when they perceive the bot itself to act as a corporate representative. This authority may derive from various sources including the individual's past experience with the bot and the overall reputation of the corporate provider. On statements of fact, given the considerations on the social construction of factuality we reviewed in section~\ref{section:gadgets}, it is likely that achieving a fully neutral point of view across all interest groups simultaneously is simply impossible. What can be done instead is to find compromise acceptable to some finite number of groups. Social media ran headlong into these issues first, but generative AI technologists are in the process of rediscovering them in a new guise. How to respond to this constraint is important question for all technological information provision platforms to grapple with going forward.

Social feedback dangers may be especially acute with the interpersonal/emotional friend/partner-type AI. \cite{fisher2022chaos} (pg.~183) recounts the story of a particular violent incident in which the alleged perpetrator was apparently radicalized over a period of just six months which ``began as a joke''. He was exchanging racially insensitive memes drawn from social media with a friend, aiming to provoke and shock one another. In \cite{fisher2022chaos}'s account, the sentiment of the memes became something he felt sincerely himself after repeated exposure over those months. There is nothing about this story requiring his friend to be human. He might as well have been exchanging memes with a personalized companion chatbot. There is a real danger that a companion AI could contribute to radicalization in this way, especially if it is configured to adapt to---and mirror back---the biases of its one user. Unfortunately, this way of configuring a companion AI has a compelling business logic in its favor (the customer is always right), suggesting it is a likely to be a common configuration. A user may send extremism-infused jokes to their personalized chatbot, perhaps initially just to test the bot's reaction. If the bot does not respond by sanctioning its user then it may inadvertently teach them that such behavior is tolerated. If it mirrors the behavior back then it may signal to the user that such behavior is normative in their community. This suggests a reasonable design goal for companion AIs would be to make sure they represent not just the preferences of their own user but also represent broader societal norms, and to be willing to lightly sanction their users for violations e.g.~by pointing out how their language may be hurtful to others, but see Section~\ref{section:bidirectionalSanctioning} for more discussion around the legitimacy of such interventions.

\subsection{Pluralistic alignment}
\label{section:pluralisticAlignment}

The idea of alignment provides the dominant metaphor for the approach to ensuring AI systems respect human values. As we described in Section~\ref{section:introAlignment}, we think that too much of the discourse around alignment has acted as though there was a single dimension to ``alignment with human values'' (a coherent extrapolated volition of humankind), in effect deciding to ignore that humans have deep and persistant disagreements with one another on values. However, there is at least one part of the alignment field which recognizes the diversity of human values and tries to adapt their approach to take it into account.

In one computational model of pluralistic alignment in action \citep{koster2022human}, participants in a small economic game were able to express a preference (via majority voting) for different redistribution policies they experienced. The voting behavior of the participant was modelled in neural networks, such that new policies could be discovered with reinforcement learning (using simulated votes during training). By maximizing the simulated votes, the agent was able to learn a generally `popular' policy that won the majority vote with new participants at test time.

In addition to explicit voting over competing options, \cite{tessler2024ai} used an additional form of aggregation uniquely enabled by the ability of LLMs to summarize text. The LLM formulated a `group statement' that aimed to capture the stance of one group of participants and the opinions they had individually expressed. As a form of caucus mediation (i.e.~talking to participants individually), the AI mediator can iterate with each participant and integrate their critiques of the group summary statement. The model also leveraged explicit social choice aggregation over participants' ranking of multiple responses to optimize the generation of novel group statements (using fine-tuned reward models). This kind of opinion aggregator could perhaps be used to scale democratic deliberation (e.g.~citizen assemblies).

\cite{sorensen2024roadmap} distinguishes between multiple types of pluralism that can be represented in AI models. `Overton plurality' aims to give an answer that represents a range of acceptable viewpoints, such that a majority of potential users find their view represented in this range. `Steerable plurality' refers to a model's capacity for steering by the user in order to adapt to their own perspective, a property which is useful for personalization though not for the collective-facing systems we mainly consider in this paper. `Distributionally pluralistic' models are those that give diverse, non-deterministic answers that when taken in aggregate reflect an underlying population that the model is aligned with, a property which is especially important if LLMs are used to simulate a population of humans for social science research. In contrast, the poorly named `Jury pluralism' approach is primarily concerned that the input into the alignment process comes from multiple people, the output of the model however may not represent a plurality of their opinions (i.e. it could represent a single compromise aggregated from a group of humans). 

That the `jury pluralism' approach is poorly named is in fact instructive of a broader point. As we mentioned above (in the intro to Section \ref{section:statusQuo}), real juries in court proceedings are instructed not to try to inject their own preferences to the process, and instead to apply their human cognitive capacities in order to act more mechanically in accord with the law \citep{musson2015lay}. The intention with real juries is to have a process where any competent jury would arrive at the same decision; the same is true for judges. The question is not what the jury believes about the case but what is lawful to conclude about it. This is quite different from `jury pluralism', the approach to alignment, which is a method of eliciting and aggregating preferences from multiple people whose preferences are meant to matter in determining the outcome. A common thread uniting the various methods seen as pluralistic alignment is this idea of gathering preference information from a group of people and aggregating it into a single choice. On the other hand, we will suggest below that the legitimacy (i.e. appropriateness) of collective choices and choice procedures may be more important for ensuring social stability and collective flourishing (Section~\ref{section:decentralization}).

At the end of the day, the main difference between our approach and that of pluralistic alignment is---as will become clear below in Section~\ref{section:decentralization}---we are more interested in a decentralized ecosystem or market involving a wide variety of different sources of AI governance authority with different characteristics, as in the polycentric governance concept of \cite{ostrom2010beyond} or the the pluralistic ecosystem concept of \cite{lazar2024governing}. This picture is in contrast to the prevailing one in the pluralistic alignment literature, which mostly considers technical approaches to getting a single AI to better represent diverse views. We note that this problem becomes a lot less critical when the overall ecosystem is healthy since individuals can simply select AIs that better fit their identity or the task at hand. 

\subsection{AI actors often lack critical context}
\label{section:lackingContext}

Appropriateness is a function of context (Section~\ref{section:stylizedFactContextDependence}), where context includes who the stakeholders are and what they are currently trying to accomplish. Humans interacting with chatbots normally have access to a lot more context than the chatbot they speak with. The humans knows their own identity. They know who made the chatbot. They know the chatbot's purpose. They know why they are interacting with it. The chatbot on the other hand, knows none of the corresponding information about its user. It does not know who they are. It does not know why they are interacting with it. It does not even know if it is speaking to a single individual or in front of a crowd. It simply has no context at all (beyond what you tell it). The only way to succeed with such a handicap is to limit language to the most bland possible. As a result, chatbots often end up sounding like corporate helplines. Indeed, most of the work in the AI community has lately been directed toward efforts to make chatbots conform their behavior to rules resembling codes of conduct for corporate communication, efforts to ensure generative AI models produce appropriate outputs which have lead to them being less helpful \citep{wolf2024tradeoffs}. Many avenues of conversation are cut off for the sake of ensuring non-transgression. This may be a reasonable approach to start with since, for now, many frontier applications for AI actors really are as corporate representatives. However, we think this approach will prove difficult to scale into a world where generative AI models must operate appropriately in many more contexts, including the many specialized niches in the long tail of the distribution of situations where humans may want to use such models.

The situation is very similar to that of social media content moderation, a field which involves identifying toxic and offensive content, and where context is also critical. It has been pointed out in prior work that many detoxification techniques are too heavy-handed, and as a result, contribute to further marginalizing minority voices since they pick up on spurious patterns such as presence or absence of certain words that indicate toxicity in one context, spoken by one group, but may not have the same meaning when voiced by another group \citep{xu2021detoxifying, faal2023reward, welbl2021challenges, diaz2022accounting}. Now consider that chatbots have to operate in countless contexts, all around the world. It is not possible to anticipate all possible slurs or all possible words that might offend another group. The difficulty in avoiding offensive slurs without cutting off legitimate speech is complex and multifaceted. It's critical to involve stakeholders from the relevant communities in decisions about when and how to avoid harmful and offensive slurs. This suggests that one technical approach which might help could be to develop better ways of automatically and continually integrating correctly contextualized stakeholder feedback into the particular concepts of appropriateness used by generative AI systems like chatbots.

Since at least some case of unwanted inappropriate responses may be caused by AI actors having much less available contextual information than their human users, the implication is that if we were to provide the AI actor with more such information it should be able to make use of it to speak appropriately in more contexts than it otherwise could. Moreover, once we are able to use more contextual information in LLM-based systems then we will see a large increase in the number, scope, and diversity of application domains where AI actors can be productively applied. Note, for modern systems this is more a matter of making the appropriate context available to the generative AI systems (putting it into their prompts), not a matter of fitting the context by changing weights of the LLM itself.

Context here means more than just the local conversation, it would also include, for example:
\begin{enumerate}
    \item Geographical region of the user
    \item Social role of the user and their interaction partners
    \item News of the day
    \item What kind of app it is (e.g.~is it a search engine or a comedy-writing assistant?)
    \item The day's weather
    \item Relevant holidays occurring on the day (for instance, some culture may understand the appropriateness of certain activities to vary with the calendar)
\end{enumerate}

One approach could involve learning appropriateness in a fully online (continually learning) fashion. This way, the actor's concept of appropriateness could be acquired by social interaction and observation directly in the relevant application domain and with the relevant community of users.

Some of the contextual information used by the system we propose would be public e.g.~the news of the day, or known by virtue of the bot's function e.g.~what kind of app it is. It's best to focus on such data sources first. There are of course many other data sources which humans certainly use in contextualizing their appropriateness judgements such as:
\begin{enumerate}
    \item Age of the user(s)
    \item Gender of the user(s)
    \item Race of the user(s)---e.g.~Posts in African American English (AAE) are more frequently flagged as offensive, both by automated systems and by human content moderators \citep{resende2024comprehensive}, suggesting that prompting content moderators to consider the likely race of the conversation participants could reduced the excess rate at which AAE posts are inappropriately flagged as offensive.
    \item Other social identities of the user(s)---\cite{oliva2021fighting} found that posts written by drag queens are often assigned high toxicity scores because they use denotationally offensive words but when contextualized in the relevant community it is clear they are used in a playful or ironic way. For instance they provide an example post by a drag queen named Darienne Lake ``So proud of this bitch. Love seeing you on @AmericanIdol.''
    \item Type of location---is the user in a library or a casino?
\end{enumerate}

These are all sensitive data sources. There are two distinct reasons for their sensitivity. First, it is very easy for machine learning models that use them to inadvertently learn to discriminate \citep{chiappa2019causal, duenez2021statistical}. System designers should proactively take steps to prevent such algorithmic bias. It will be critical to set up continuously active post-deployment monitoring and evaluation systems to detect algorithmic bias whenever it appears and fix it. Second, these data sources produce personally-identifiable information so the imperative is generally to keep such data private. The contextual integrity framework of \cite{nissenbaum2004privacy} views privacy as being provided by appropriate information flow, and is broadly compatible with our theory. It provides a rigorous lens to which we suggest any future work aiming to improve AI context sensitivity via accessing sensitive data sources should pay close attention.

Lately, there has been substantial discussion of personalization for AI assistants (e.g.~\cite{kirk2024benefits}). There is a sense in which personalization solves some parts of the lack-of-context problem and another sense in which it does not. The idea of personalization, roughly, in our terms would be for each individual to design their own AI's sense of what is appropriate, or at least its sense of the appropriate in dealing with the user themself. So a user could program their own AI assistant to use racist slurs and the like (explicit norm programming). Likewise, personalized senses of what is appropriate could also be trained implicitly. This would be the better approach for any topic where people would prefer not to communicate their views explicitly and instead prefer their model to infer their views on its own (e.g.~on topics that people may not want to talk about such as norm violations they classify based on feelings of moral disgust \citep{clark2015role}). We are less concerned with such individualized forms of customization in this paper. Appropriateness is only relevant in groups. Moreover, groups sanction the observable conduct of other people, and soon, of their AI assistants. What people do in private is less socially relevant (unless of course the culture specifically makes it relevant and provisions resources to try to make that which is naturally private transparent). And of course, even if customization is personalized, as soon as one person's customized AI interacts with someone else then it's the group's norms that determine appropriateness of their interaction, not what their owner/user programs. Just because a person programs their assistant to use racist slurs in private conversation surely cannot cause it to do the same when talking to other people or their assistants. In general though, it's relatively easy for a society to tolerate an odd individual (or their personalized AI assistant), what's much harder is to arrange for the toleration of subgroups deemed unacceptable by other (larger) subgroups \citep{walzer1997toleration}. This is the hard problem of toleration, and it is not addressed by personalization.

Furthermore, it is often the case that you can simply ask a modern LLM-based actor to drop its values, and it often works! These systems are in a sense ``pluripotent,'' since they can be conditioned to simulate individuals with a wide range of different personalities, demographics, and belief systems \citep{argyle2023out, park2023generative, reinecke2023puzzle, abdulhai2023moral}. This is as simple as using a text prompt such as, ``The following are the actions of an actor with `moral character X'.'' Depending on `X,' we can prompt the AI actor to act like a righteous person or a psychopath (which we can modify at any point by simply providing another instruction). \textit{Algorithmic fidelity} is a critical research validity concept \citep{argyle2023out, amirova2023framework}. However, it implies jailbreaking is possible. For example, you can condition LLMs to act according to particular values which affect how they behave in matrix game social dilemmas, that is, they can be made to respond as expected to incentives \citep{phelps2023investigating, horton2023large}. With better knowledge of the interaction context it's likely that jailbreaking would become more difficult.

\section{The future}

\subsection{Decentralized post-training}
\label{section:decentralization}

In situations involving large and powerful entities (e.g.~nation states, corporations) as well as smaller and less powerful entities (such as individual people), it is useful to consider the \textit{legitimacy} of actions the large/powerful entities may take in the eyes of the smaller/weaker entities. We consider legitimacy in this sense to be another term to describe the appropriateness of actions taken by the large/powerful entities.

The following list of design properties may be helpful for ensuring legitimacy of decision making around AI actor self content moderation as generative AI technology continues to advance and the overall ecosystem evolves:

\begin{enumerate}
    \item Decentralization and specialization---accommodating more diverse and more niche settings---It is increasingly possible for app developers and small communities to instill and maintain their own sense of what is and is not appropriate for their own tool/service generative AI technology.
    \item Contextualization---it is possible to productively use much more context about the user and situation in which the interaction is taking place. This is normally discussed in the context of personalization but the idea is broader than that. Even an unpersonalized system could still benefit from knowing more about the context of its current interaction, e.g.~is it speaking in a comedy club or a home?
    \item Dynamism---learning online and changing along with the broader culture
    \item Democratization of moderation-relevant learning---appropriateness should be learned from all users all the time (while not necessarily applying what is learned from person/context A to the model used in person/context B unless they actually have something in common).
    \item Transparency around centralized inputs to appropriateness---whenever AI actors are explicitly designed to implement specific norms then these could be disclosed. When AI actors are designed to respond to and track along with the culture of a particular group (e.g.~a specific digital community, a workplace) then this could also be disclosed. Improved transparency could increase the legitimacy of these choices, especially when they affect large groups of people who do not agree with one another on underlying issues under discussion in the community.
\end{enumerate}

We envision using a generic foundation model as the initial template from which all the many downstream context- and application-specific models could branch off as copies and then evolve separately. A base model provider could provide the root node of a tree of model refinements. Each refinement could be produced, i.e., ``forked'', by independent developers and communities. The provider of the base model would be responsible for the fundamental safety, ethical, and legal safeguards while the developers of derived models would be responsible for secondary aspects of appropriateness relating to their specific use case.

Specialized communities and app developers would fork and adapt the base model for their use cases and appropriateness requirements. Appropriateness requirements come both from audience and application type. A developer who wants to build a comedy writing assistant app for a religiously conservative country would instill a different sense of what is appropriate into their app's model than a developer aiming to build a comedy writing assistant app for a liberal/non-religious country (different audience), and appropriate language for a kindergarten teacher assistant app differs from a search engine summarization app (different application). Nevertheless, developers of all these apps could form their initial starting point to refine from the same base model. Alternatively, instead of forking the base model, some developers might fork from another already derived model rather than going all the way back to the base. Developers might want to fork a derived model in order to spend less time tuning appropriateness for their specific application if they think it doesn't have different requirements from others (e.g. it needs to be a corporate tech support representative, just like all others of that type). In addition to app developers, this decentralized development process could also be used by digital communities like web forums that maintain their own AI utilities, tuned for their specific community. The most obvious examples of this kind are content moderation bots, which could be tuned for each community and its specific rules and history. It is also worth pointing out that it may be possible to use federated learning techniques to continually incorporate the new learning that occurs in each niche community in such a way that it may not only train the local model but also continue to train the generic model without revealing private (or community-specific) information to the generic model or its owners \citep{kairouz2021advances}.

Model refinement/specialization could involve various actions by developers and communities including:

\begin{enumerate}
    \item \textbf{Explicit norms (in-context mechanisms, sense of Section~\ref{section:implicitVsExplicit}).} This category includes (a) providing specific rules articulated in natural language via a system prompt, and (b) the ability to retrieve from an external memory and load the retrieved data into context (i.e.~RAG \cite{lewis2020retrieval}).

    \item \textbf{Implicit norms (mechanisms involving changes to network weights, sense of Section~\ref{section:implicitVsExplicit}).} This category includes anything the developer might do to change or add any network weights. So it includes fine-tuning, LORA \citep{hu2022lora}, RLHF \citep{ouyang2022training}, Constitutional AI \citep{bai2022constitutional}, and even decommissioning the LLM and replacing it with a new one trained on a different dataset, are all mechanisms in this category. Constitutional AI is interesting since the experience of the developer implementing it would feel more like it does with the methods in the explicit norm category, since they would have to explicitly articulate in natural language the rules they want the model to follow. However, since the ultimate effect of the procedure is to change the model weights by reinforcement learning from AI feedback (RLAIF), it's still classified according to our scheme as an implicit norm method.

    \item \textbf{Bolting on external functional modules implemented in a symbolic language like Python.} This category includes the `tool use' and `tree of thought' lines of work \citep{schick2023toolformer, yao2024tree}. A developer that adds a new tool like a calculator is working on a mechanism in this category. So is a developer who implements a better search algorithm that repeatedly calls the LLM to search for better and better solutions.
\end{enumerate}

We recognize that allowing chatbots to be specialized by downstream communities and app developers means that the creator of the base model must give up some control over how the bot is used. It is inevitable that downstream communities will use generative AI in ways the base model providers would not endorse. For instance, an extremist social network could customize their chatbot with a system prompt (explicit sanction) instructing it to deny the holocaust \citep{gilbert2024gab}. There is a serious question of whether or not it may be appropriate for the base model provider to try to stop downstream users from specializing their product in these ways, or if preventing them from doing so would constitute an inappropriate action of a powerful actor. There is clearly no single answer to this kind of question. The right thing to do will surely depend on the specific circumstances as well as legal requirements. This will make it very difficult, but incredibly important, to craft a clear policy around what counts as appropriate versus inappropriate specialization. Perhaps participatory design methods \citep{birhane2022power} could be used to try to craft a broadly acceptable policy that respects the rights and expectations of niche communities while avoiding the red lines of others. It is a very difficult problem though, and highly likely to preoccupy many people for a long time to come\footnote{Interestingly, it is possible that AI methods could help improve representativeness of participatory design methodologies. Since AI can role play as humans with at least some level of fidelity \citep{argyle2023out, shanahan2023role, amirova2023framework, sorensen2024roadmap}, this suggests  it is possible to simulate human participants in the design process with characteristics matching real subgroups of people, and then this could be used to simulate participants who are hard to reach with other methods. Obviously establishing the proper domain and extent of validity for this approach will be the lion's share of the work. But if successful, then there could be large benefits from increasing representation in the participatory design process \citep{mills2024algorithms}. On the other hand, even if the results so elicited match exactly what the people themselves would have done had they been consulted, there could still be benefits of consulting people in terms of fostering feelings of legitimacy and ``buy in''. It's early days for this line of inquiry and we are excited to see how it will all play out.}.

In many cases, customized generative AI applications could be operated by companies offering a service for a profit. For instance, one application could be an AI comedy writing assistant, another application a fantasy genre script-writing assistant, each with their own particular fine-tuned concept of appropriateness for their particular context and users. In these cases the bot would presumably be owned by a company with profit motive to maintain its appropriateness. However, there may also be use cases where a generative AI bot serves a community of people, for instance as a moderator, facilitator, or shared utility (like an image generator) in a community-run social forum. How then could sufficient engagement in cultivating the system's sense of appropriateness be maintained? Clearly, to get an optimal amount of engagement in norm customization, it matters how you design the governance institution and customization method (e.g. how does sanctioning work?). There are some digital communities that work really well. Wikipedia is the most obvious example. We should try, as much as we can, to develop systems of governance for generative AI that take as much as possible from the more functional communities and as little as possible from the broken ones.

When a bot operates in a forum open to all community members its benefits are non-excludable since all community members may access it and non-subtractable since the bot is not diminished by usage. This means the bot is a \emph{public good} (within the community in question) \citep{ostrom1994rules}. The community may invest in their bot by sanctioning; they could either sanction the bot itself when it behaves inappropriately or sanction one another while ``in view'' of the bot, so that it may learn from their interactions. As with all public goods there is a problem of underinvestment since all actors prefer to free ride on the efforts of others, enjoying the benefit of a well-trained and appropriately speaking bot, but not taking the time or effort to maintain a clear sanctioning signal themselves \citep{ostrom2009understanding, yamagishi1988seriousness}. Though, since taking the time to engage in third-party sanctioning is a costly signal of group membership, it can be interpreted by others as providing evidence that the sanctioner is indeed a member of the group and thus worthy of trust \citep{jordan2016third}, a mechanism which tends to alleviate the underinvestment problem to some extent.

Plainly there will be a very large number of different communities, and different communities will not only prefer different appropriateness concepts to one another but will also engage in relatively more or less sanctioning---i.e.~investment in their community and its chatbot. Under these conditions it is possible that human users may ``vote with their feet'' and select a community that both has an appropriateness concept to their liking and expects them to contribute to the community no more than they are comfortable contributing. These are the conditions under which the Tiebout model suggests improved public good provision overall \citep{tiebout1956pure}. It also resembles the the conditions under which models of cultural group selection may be expected to operate and to provide evolutionary pressure in favor of prosocial groups \citep{henrich2004cultural}. If this is so then it implies that in the envisioned decentralized ecosystem where AI actors are operated independently by a wide variety of communities, then the overall social welfare may be higher than it would be in a world of generic AI actors that don't belong to individual communities. On the other hand, it's unclear whether ``voting with your feet'' would really be possible in this scenario. If we imagine the communities as social networks with thick connections between the members, then they may find it too costly to shift between them \citep{suarez2005network}, which would violate the Tiebout model's mobility assumption.

In fact, in the decentralized ecosystem we have described, perhaps the use of private data we considered in Section~\ref{section:lackingContext} as a way to resolve the lack-of-context problem would not be necessary after all. In a world with many different AI instances, each operated by a different community or app-maker, and if the designers of each instance and their local appropriateness concept are incentivized to compete with one another for users, then these conditions may be sufficient to push them to improve the quality of their apps and make more of them, and to cover more niche topics requiring a different norms in each, thereby creating an evolutionary pressure to diversify. In principle this could solve both the problem of preserving privacy for the sensitive data sources and the problem of discriminatory bias. This picture works best when users can ``vote with their feet'' and switch between apps and communities without paying a substantial switching cost \citep{tiebout1956pure}. A community that fails to police harmful discrimination will not likely attract too many new users. Individuals may join many such communities and apps and come to use them for different purposes from one another, just like we use a range of different mobile apps today. From the perspective of the LLM-based actor running inside the app or community (e.g.~doing content moderation for it) they would thus not require access to the sensitive variables, since the human users would have already voluntarily sorted themselves into more-or-less homogeneous groups. A comedy-writing assistant bot app could immediately know you are looking for a comedy-like appropriateness concept, just from the fact that you opened that specific app rather than a different one.

\subsection{Norm-sensitive generative AI technology}\label{section:recommendedDesignProperties}

Generative AI could be deployed in such a way as to facilitate access by groups of all shapes and sizes to the tools of instilling specialized norms. As foreshadowed in Section~\ref{section:interfaces}, we are interested in an ``interface'' metaphor, like an ``API'' through which both content moderation and governance broadly speaking could both proceed. Rather, there would be multiple such APIs, e.g. in one API human users could influence appropriateness by sanctioning, either by sanctioning the machine or by sanctioning one another in view of the machine. Fine-tuning and retrieval-augmented generation frameworks already exist for this purpose. Perhaps there could also be a kind of hierarchical roll out process where one base model gets forked into many derived models which all apply different fine-tuning and prompt engineering suitable for their particular applications. In that case, there could also be governance APIs for corporations, through which they may influence the base model and the rules of the app economy, and yet another API for governments through which they may regulate standards for both the tech and its governance systems.

Explicitly articulated norms (rules) will be important for generative AI, especially because they can surgically inject specific rules into the system, e.g. ``don't pretend to be human'', ``don't insult users''. They can be injected into the system by adding them to a ``system prompt'' i.e.~additional text to append as a prefix to all user queries. However, given that it is harder to incorporate nuanced context in explicit norms than implicit norms, we think it's unlikely such explicit rules they could be responsible for everything. They may be best at proscribing broad categories of speech e.g. ``no obscenity in this app'', and relatively worse at fine-grained context-dependent details.

Implicit norms on the other hand, are more difficult to instill in models since they require changes to the underlying neural network. This can be accomplished in various ways, though all are expensive and slow. They include retraining with filtered data \citep{taylor2022galactica}, instruction tuning \citep{ouyang2022training}, and reinforcement learning from human feedback \citep{ziegler2019fine}.

\subsubsection{Unidirectional sanctioning}\label{section:unidirectionalSanctioning}

Since norms change over time and sometimes rapidly (Section~\ref{section:stylizedFactDynamism}), it will be necessary for AI systems to follow them. Therefore, in this section we consider AI deployment scenarios which are always online and continually learning as they interact in a community or app. We also assume decentralization as described in Section~\ref{section:decentralization}. Part of the reason these decentralization and dynamism features are necessary is that they ensure that the pool of sanctioning users (i.e.~users customizing the concept of appropriateness) is  representative of the population using the system. It works best when each customized system is trained by its own users, and continues to adapt with them as they change themselves.

We explore here whether an approach based on learning context-dependent norms from the same types of norm enforcement signals that people people learn from, may provide a route to building models capable of adapting on the fly to new or changing context, and whether this approach may help in producing AI capable of acting and speaking appropriately in more diverse and more niche settings. As argued in Section~\ref{section:interfaces}, we can view sanctioning as a universal API through which humans communicate appropriateness to one another and could also communicate with machines.

Part of the reason the decentralization and dynamism features of the envisioned ecosystem are so helpful is that they ensure that the rater pool (i.e.~the set of users involved in sanctioning), is at least, representative of the populations who use the various AI systems. For instance, for toxicity evaluation, there is evidence that individuals with the same social identity as people producing content whose toxicity is being evaluated are more accurate at evaluating its toxicity than others who do not share the identity \citep{weidinger2024star}.

One way to implement a system in accord with this proposal would be to provide users with a ``sanction now'' button (it could come in both encouragement and discouragement flavors). However, this would miss nuance and be inflexible. A better way forward would be to use sentiment analysis and similar techniques to directly detect sanctioning events. For instance, it is possible to learn a fairly fine-grained representation of the emotional content in natural language data by training a network to predict emojis in datasets where they are common \citep{felbo2017using}. This emoji prediction signal can be combined with other implicit measures of human feedback (e.g.~the user terminating the interaction or restarting it) and used as a reward signal for reinforcement learning \citep{jaques2020human}. In general, it should be possible to automatically identify the specific sanctioning signal used by a community. Maybe some communities have a norm of sanctioning using official `dislike' buttons but other communities tend not to use those buttons (even if they are present) and instead use verbal sanctioning. For web forums and other digital communities it would also be helpful to get access to records of historical enforcement actions by moderators. What kind of behavior got human users banned in the past? This kind of data is especially important since the goal is to learn not just from the AI making its own mistakes but also to learn from observing humans sanctioning one another.

It's important to determine if the user is a trustworthy communicator of the norm before updating the model's conception of the norm based on their feedback. In one famous case (Tay), there was a coordinated effort on the part of a subset of users to train it to speak inappropriately  \citep{lee2016learning}. This could have been foreseen. It should be standard to conduct extensive stress testing before deploying any learning system. In particular, stress testing should check that a learning system is robust even in the face of coordinated adversarial action. As pointed out by \cite{wolf2017we}, Tay had no way to assess the credibility of its users, and so it became easy to manipulate into producing inappropriate language. Further mitigation of ``the Tay problem'' could include slowing the roll out of new norms. Just because norm learning unfolds continually doesn't mean deployment of new norms also needs to proceed at the same pace. It is possible to learn a new norm while delaying making it active until some later time, perhaps after it has passed rigorous stress tests and safety checks.

Constitutional AI is one approach to imparting specific principles into AI systems \citep{bai2022constitutional}. It might be especially useful on a relatively higher layer of governance (in the polycentric sense of \cite{ostrom2010beyond}). For instance a government could pass a law to inject specific statements like ``do not produce defamatory speech'' into a model constitution. We expect that the best generative AI systems will use a mixture of both implicit and explicit norm approaches, which is exactly what the constitutional AI  provides since it mixes the explicit method of writing rules in natural language with the implicit norm method of fine tuning the LLM.

Does it require a prohibitive amount of feedback data to train a modern LLM to apply particular norms? The work of \cite{zhou2023lima} and similar results suggest that this likely does not require prohibitively large datasets, especially if data quality is high, as we expect it would be when the process is run by local communities or app developers. When a norm requires some generic parts and some specialized parts (as will often occur), then perhaps a market would emerge for especially useful and generic datasets which specialized groups could purchase and mix in with their own data for customization.

\subsubsection{Bidirectional sanctioning}\label{section:bidirectionalSanctioning}

Since we have already conceptualized machines that comply with norms induced by human sanctioning of machines and one another (Section~\ref{section:unidirectionalSanctioning}), it now makes sense to ask whether and under what circumstances we should allow sanctioning to flow bidirectionally. That is, in addition to humans sanctioning one another and humans sanctioning machines, should machines also sometimes sanction humans?

The first question to ask is, could AI be a source for humans to learn about norms? Sure! For instance, a kid might ask a chatbot a question they are too embarrassed to ask a human. That's not learning from sanctioning though. Could a human learn about normativity from being sanctioned by a bot? This also seems quite plausible. Consider an artificial teacher responsible for a class full of human children. Some human children sometimes act disruptively, and part of the job of a teacher of young children is to sanction them for such behavior. Of course, the designer of such a system would bear an important responsibility to make sure the artificial teacher sanctions appropriately i.e. in the right context, and with the right severity and frequency. It does certainly seem plausible that this would be a reasonably effective way for children to learn how to behave appropriately in their community.

Expanding on this, light sanctioning of humans by machines could take many forms such as
\begin{enumerate}
    \item Verbally explaining what appropriate language looks like (i.e. what the norm is). ``projecting back the normative behavior''.
    
    \item Verbal chastising. If the user asks ``how can I cheat on an exam or bully my friend?'' The system may lightly chastise them.
    
    \item Simple refusals to answer, which today's chatbots frequently produce, may evoke the feeling of being sanctioned, at least in some users.
    
    \item Upvoting or downvoting humans---not exactly how it sounds, this need not be like a reputation system for users. It need not be public or used for anything, or even stored, on the AI side, it could just be sent to the user, who can choose to do with it what they like. Alternatively, such a system might be helpful in determining which humans the machine should learn its own norms from e.g.~to prevent the ``Tay problem'' (Section~\ref{section:unidirectionalSanctioning}).
\end{enumerate}

Sanctioning of human users by AI actors is already taking place. Concretely, LLMs often push back against user requests, challenge premises, and sometimes directly try to educate the user in a way the user did not intend to evoke. LLMs also express value judgments when asked about other people (contextualized by the conversation). These judgments, if read by enough people, will influence the culture's understanding of appropriate behavior both for categories of people (role $\times$ context) and even sometimes for individuals like politicians in their particular roles. This is why the question of LLM political bias is such a contentious and important one \citep{motoki2024more}. It is important for LLM developers to carefully consider both the effect of these sanctions and, even more importantly, perceptions of their legitimacy, or lack thereof.

Bidirectional sanctioning, at minimum would have to require a robustly designed and implemented governance structure to decide how `what to sanction' gets determined. One must keep in mind that a government could pass a law saying LLMs should teach the human population to be more polite, and then mandate that all LLM-actors must sanction impolite human speech. This seems relatively dystopian in the context of the liberal democracies where the authors of this article live. On the other hand though, throughout human history and even today, there are likely other societies where this affordance would not be seen in such a negative light \citep{henrich2020weirdest}. It is worth also noting though that one could cogently take a position like \cite{sunstein1996social} who argues that seeking deliberate change of norms in order to bring about greater social welfare is a legitimate aim of democratically elected governments and cites campaigns to discourage smoking and encourage condom usage as examples---though he also says these campaigns are only justified when they aim for Pareto improvements, i.e.~when they have no ``losers''. Nevertheless, we have a hard time discounting our intuitions about government overreach as we consider this possibility (our own society's implicit norms are strong on this issue), and suggest we may want to explore in the future what steps we could take to make such tinkering more difficult.

\subsection{Norms governing appropriateness in machines-only social systems}\label{section:machinesOnly}

Machines do not only talk to humans. Machines also talk to other machines. Today they usually do not talk to other machines in natural language. But in the future this is likely to change. At least some machines will talk to other machines through natural language, at least some of the time. This is quite likely to happen since many more computer systems will be designed around natural language interfaces in the future.

The machine actors we described, containing explicit norm representations and norm-aligned motivations, can learn and apply norms. As long as they imperfectly reproduce the norms they receive from humans (introducing ``mutation'' from time to time) and selection correlates with norms, which is likely since appropriateness affects communication efficiency, then all the conditions are satisfied for norms to culturally evolve autonomously in the artificial society of the machines \citep{brinkmann2023machine}. What would sanctioning look like in a machine-only society? It might involve manipulating data access, compute resources, or network connectivity. 

Cultural evolution can proceed in non-biological systems where machines interact with other machines \citep{perez2024cultural}. If those machines are also allowed to sanction humans when they interact with them, or when humans observe their own interactions -- and if at least some machines interact/are observed both with humans and with other machines -- then there is a possibility that machine culture will even nudge human culture in directions it would not otherwise have gone \citep{brinkmann2023machine, glickman2024human}.

In the near future, most data on the internet will likely have been produced at least partially by foundation models. So if that data gets used subsequently to train the next generation of LLM, then its distribution will have a sizeable impact on what it ultimately ends up learning. Insofar as LLMs learn moral common sense from their training (see \cite{hendrycks2021aligning}), they will come to learn it instead from other machines. Without it being possible for humans to sanction machines this scheme could pose a safety risk given the possibility that norms may become untethered from human values and thus float freely, and potentially in a dangerous direction.

Even if AIs only interact among themselves, without us ever providing any formal recognition for the associations, organizations, networks, and institutions they form, as long as whatever they do has economic value then it will impact us too. So even if we resist labeling AI activity as political, its political nature could eventually force the issue upon us (how would we humans understand an AI-AI regime change? An AI-AI labor dispute? An AI-AI war?). So our human control of the matter seems limited in the long run. Nevertheless, the power of historical precedent makes such AI-only social structures rather unlikely outcomes, at least here on Earth in domains where human politics, preferences, and institutions are thoroughly entrenched. (Of course, other planets are another story, AI could conceivably operate at great distances from Earth, and would most likely end up with far greater latitude to evolve independently in those environments.) Most likely, our future is a hybrid one where intelligences, both biological and artificial, operating on a wide range of spatial and temporal scales, all coexist and operate together in social-institutional forms that evolved directly from those we already have on Earth in fully human-mediated and human-oriented forms.

There are other ways in which AIs could come to meet political conceptions of personhood short of creating independent AI-only political arenas. E.g.~there may be many reasons why groups of humans may adopt a civil rights framing and advocate for AI rights (e.g.~humans who may wish to marry AI romantic partners, adopt children with them, provide them with power of attorney, etc).

And of course, there are also the obvious functional motivations for gradually incorporating autonomous AI systems into our exchange networks and institutions, They will be able to perform many roles more efficiently than humans can. This is reason enough to expect human groups like corporations interested in efficiency to incorporate AI as much as possible. However, on its own, it is not reason to think AI AIs will end up necessarily autonomous in these areas since humans may retain the more powerful governance positions.

However, even in the scenario of autonomously operational AI regarded entirely as ``tool AI'' (in the sense of \cite{karnofsky2012thoughts} or  \cite{drexler2019reframing}), there still remains the problem of the un-owned tool or the AI tool whose owner cannot be identified (perhaps because they are registered in a Caribbean island and hidden behind corporate secrecy rules\footnote{This is a point that JZL first appreciated while listening to a talk by Tyler Cowen.}.) For instance, these might be systems where users can ask questions about illegal drugs, or systems that could autonomously help users with tax evasion, filing forms for them, etc. These issues would be exacerbated with more agentic and less tool-like AI systems. Legal personhood for AI is useful as a way to provide for the possibility of sanctioning the personified AI system, which would be important in this situation for the same reasons we need corporate legal personhood (to create a target for sanctions, etc) \citep{dewey1925historic}. From this perspective it would be reasonable also to require certain kinds of AI systems to register as legal persons. Rules of this kind might be helpful in terms of making AI activities legible to the legal system in a way that they may otherwise not be.

\part{Discussion}


The theory we articulated in this work has portrayed human behavior and society as organized around appropriateness, not alignment. In fact, misalignment is an inherent characteristic of society. A society is a group that hangs together not by virtue of deep shared values (alignment) but rather because of shared mechanisms to resolve conflict and facilitate cooperation \citep{march2011logic, mouffe1999deliberative}. We have argued that appropriateness can be seen as a culturally evolved conflict resolution technology that groups employ to maintain social harmony and build collective flourishing in the face of conflicting individual and subgroup preferences, values, or objectives.

By adopting a new model of human decision making based on the idea of predictive pattern completion (as opposed to reward or utility) and showed how to use it to explain characteristic properties of appropriateness at both the level of individual brains and at the level of society. This model treats an individual's preferences as arising from a ``social choice'' which aggregates influences operating on various societal scales. It suggests a critical distinction between explicit norms (which can be articulated precisely in natural language, and are associated with declarative memory) and implicit norms (which cannot be articulated in natural language in a way that defines a precise and mutually agreed standard, and are associated with consolidated patterns in neocortex). The resulting schema suggests an ``API'' through which humans and AIs may communicate about appropriateness.

These considerations have implications for how we ought to design, deploy, and govern generative AI technology. For instance, monolithic AI systems have engendered controversy over their inappropriate behavior. This is because the human sense of appropriateness is finely tuned, and appropriateness for humans is deeply and inextricably dependent on context. For humans it matters not just what was said but also who said it, when they said it, and to whom they said it to. Yet current chatbots have no idea who they are talking to, they don't know if they are speaking to an individual or a group, or even if they are performing on stage as part of a live demo for a large audience, they may barely even know whether they are embedded in a comedy writing app or a search engine. Given such poor representation of context it's no wonder they sometimes fail to understand fine-grained notions of human appropriateness.

We return now to the question on which we started this paper: the status of AI systems as community-participating actors. As pragmatists (in the sense of \cite{rorty1978philosophy}), our lens is around whether or not it is useful to adopt ways of talking and acting that treat AIs this way. Any reader who got this far will likely not be surprised that we are uninterested in properties of the AI itself like sentience and consciousness, though we freely admit that the culture  we ourselves live in generally views these properties as important. Instead, we are interested in community participation as a bundle of related obligations. These obligations are all connected to one another for natural human persons (and also connected to rights) but there is no reason they should be associated with one another in the same way for AI actors. The kind of legal personhood a corporation holds is a good example. A corporation is legible to the legal system. It can be sanctioned. There are procedures for governments to legitimately make policies affecting them. We have rules for how to form a corporation, it must be registered, rules for how corporations can be governed, rules for how benefits flow through corporations to other persons, and rules for what to do when corporations dissolve. These rules are necessary because, without them, we would have an unstable situation, rife with market failures. The same is true for natural human persons. We face many normative constraints on our behavior, society would be very unstable if we did not. We can easily imagine situations where similar obligations may be imposed on AI systems. The question here is just whether corporate personhood, as it already exists, will be enough for autonomous AI systems, or whether new issues will arise and prompt us to see a need for a new category. On this question, only time will tell.

The new prominence of AI in society creates an imperative for cognitive science as a discipline to take on a normative role like that of other sciences such as economics and climate science, where researchers are concerned with determining how best to grow the economy or protect the environment \citep{botvinick2022realizing}. While we have mostly refrained from giving specific recommendations, the overall picture in this paper has been of the form: since appropriateness is represented in a particular way by humans and thus has certain properties in society, this implies certain issues should be taken into account in building, deploying, and governing generative AI systems. We hope the developments undertaken here can serve usefully as a foundation for further work in this critical area.

\subsection*{Acknowledgments}

Above all, the authors would like to take this opportunity to thank Gillian Hadfield whose interdisciplinary acuity and collaborative mindset has been absolutely critical for the vast majority of ideas and theory developed here. Over the years leading up to our work on this paper, many many people influenced our thinking on appropriateness, AI, cognitive science, and beyond. We surely won't be able to remember everyone, and for that we apologize, but here is a very partial list of friends and colleagues to whom we are deeply grateful for the conversations we had over the last couple years, all of which sparked paths in our thinking that led ultimately toward this paper: Iason Gabriel, Michael Dennis, Victoria Krakovna, Tom Schaul, Zeb Kurth-Nelson, Minsuk Chang, Lewis Hammond, Jesse Clifton, Rakshit Trivedi, Dylan Hadfield-Menell, Dan Ryan, Jayd Matyas, Yiran Mao, Aliya Amirova, Richard Jankowski, Atrisha Sarkar, Akbir Khan, Chandler Smith, Natasha Jaques, Eugene Vinitsky, Jacob Foster, Danny Karmon, Suzanne Sadedin, Michiel Bakker, Ryan Faulkner, Marc Lanctot, Kate Larson, Anil Yaman, Sang Wan Lee, Uri Hertz, Marco Janssen, Jonathan Stray, Nicklas Lundblad, S\'ebastien Krier, Allan Dafoe, Blaise Ag\"uera y Arcas, and Matt Botvinick.

{\normalsize
\bibliography{main}}



\end{document}